\documentclass[10pt,twocolumn,letterpaper]{article}

\usepackage[pagenumbers]{wacv} 

\usepackage{graphicx}
\usepackage{amsmath}
\usepackage{amssymb}
\usepackage{booktabs}
\usepackage{lipsum}
\usepackage{kotex}
\usepackage{color}
\usepackage[table]{xcolor}

\usepackage{array}
\usepackage{tabularray}
\usepackage{enumitem}
\usepackage[accsupp]{axessibility}  

\usepackage[pagebackref,breaklinks,colorlinks]{hyperref}

\usepackage[capitalize]{cleveref}
\crefname{section}{Sec.}{Secs.}
\Crefname{section}{Section}{Sections}
\Crefname{table}{Table}{Tables}
\crefname{table}{Tab.}{Tabs.}

\newcommand{\yr}[1]{\textcolor{black}{#1}}

\def\wacvPaperID{1939} 
\def\confName{WACV}
\def\confYear{2024}

\begin{document}

\title{Fast Sun-aligned Outdoor Scene Relighting based on TensoRF} 

\author{Yeonjin Chang$^1$
\and Yearim Kim$^1$
\and Seunghyeon Seo$^1$
\and Jung Yi$^2$
\and Nojun Kwak$^1$
\vspace{1mm}
\and $^1$Seoul National University \\
\tt\small \{yjean8315,yerim1656,zzzlssh,nojunk\}@snu.ac.kr
\and $^2$Yonsei University \\ 
\tt\small master9906@yonsei.ac.kr
}

\maketitle

\begin{abstract}

In this work, we introduce our method of outdoor scene relighting for  Neural Radiance Fields (NeRF) named Sun-aligned Relighting TensoRF (SR-TensoRF). SR-TensoRF offers a lightweight and rapid pipeline aligned with the sun, thereby achieving a simplified workflow that eliminates the need for environment maps. Our sun-alignment strategy is motivated by the insight that shadows, unlike viewpoint-dependent albedo, are determined by light direction. We directly use the sun direction as an input during shadow generation, simplifying the requirements of the inference process significantly. Moreover, SR-TensoRF leverages the training efficiency of TensoRF by incorporating our proposed cubemap concept, resulting in notable acceleration in both training and rendering processes compared to existing methods.

\end{abstract}

\begin{figure}[t!]
  \centering
  \begin{subfigure}{0.75\linewidth}
    \centering
    \includegraphics[width=1.0\linewidth]{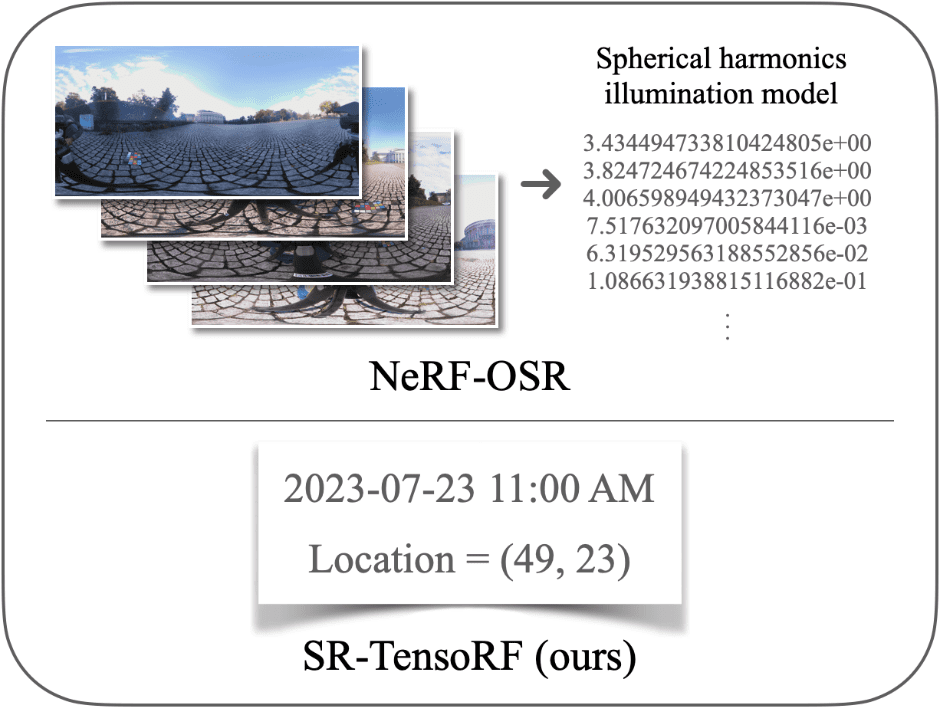}
    \caption{training}
    \label{fig:teaser-a}
  \end{subfigure}
  \begin{subfigure}{0.75\linewidth}
    \centering
    \includegraphics[width=1.0\linewidth]{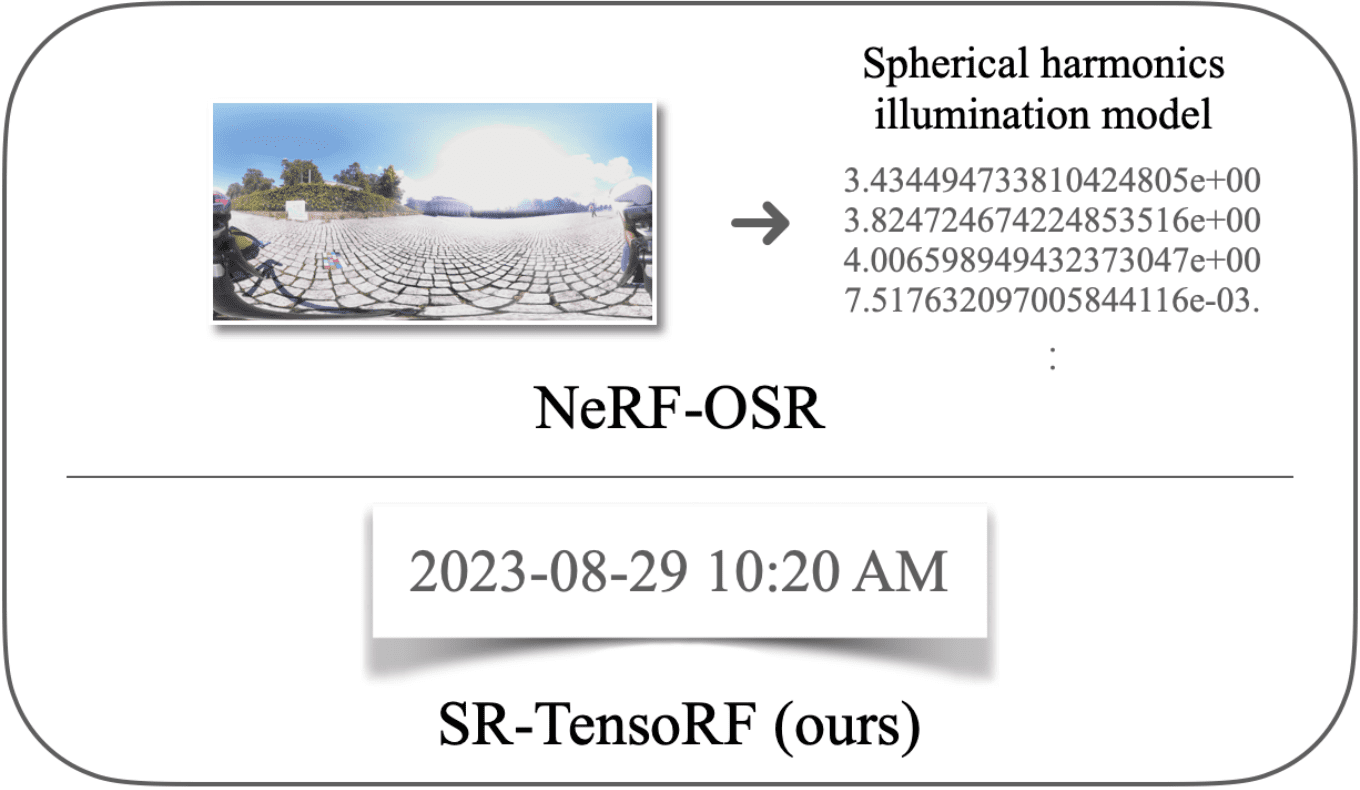}
    \caption{inference}
    \label{fig:teaser-b}
  \end{subfigure}
  \caption{\textbf{Our requirements.} Instead of using environment maps, we propose a simplified relighting pipeline that relies on time and location information, making the requirements much simpler. } 
  \vspace{-3mm}
   \label{fig:teaser}
\end{figure}

\section{Introduction}
\label{sec:intro}
Recently, Neural Radiance Fields (NeRF)~\cite{mildenhall2020nerf} has achieved a breakthrough in the novel view synthesis task, sparking significant interest within the academic community in implicit 3D reconstruction. 
While NeRF's impressive performance is undeniable, numerous challenges remain for further enhancement and research, such as reconstructing materials with various properties like transparency, translucency, or reflectiveness~\cite{fujitomi2022lb, tang2023able, verbin2022ref, yan2023nerf, guo2022nerfren}, reconstructing dynamic scenes~\cite{attal2021torf, tretschk2021non, li2021neural, pumarola2021d} or scenes with varying scales~\cite{tancik2022block, turki2022mega, barron2022mip}, efficient training and inference~\cite{muller2022instant, chen2023mobilenerf, fridovich2022plenoxels, yu2021plenoctrees, chen2022tensorf}, and more. 
In addition to the aforementioned challenges, another area of interest lies in the editing of reconstructed scenes~\cite{yuan2022nerf, haque2023instruct}.
One aspect of such editing involves manipulating lighting conditions~\cite{srinivasan2021nerv, rudnev2022nerf, toschi2023relight}, which is referred to as relighting.

Relighting is a classical task in computer vision that enhances the realism and versatility of visual content, benefiting applications like virtual environments, entertainment media, and architectural visualization.
Due to the intrinsic difference in lighting between indoor and outdoor scenes, researchers have treated them as separate domains and categorized the relighting problem into indoor and outdoor scene relighting.
In this study, we focus on the daytime outdoor setting which can be characterized as;
1) Sunlight stands as the most dominant light source.
2) Outdoor settings lack the ability to control light sources contrary to indoor settings utilizing controllable light conditions.
3) Outdoor scenes are inherently unbounded so that they are profoundly influenced by the surrounding objects and buildings while consistently dealing with infinite potential light sources. 
4) Training images might have varying white balances, even within the same timeframe.

Recently, with the purpose of achieving the dual objectives of novel view synthesis and relighting,  NeRF-OSR~\cite{rudnev2022nerf}, which decomposes the implicit scene using shadows, albedo, and an illumination model based on spherical harmonics, has been proposed.
While the approach generated plausible outputs with their relighting dataset, confirming its alignment with the aforementioned outdoor scene features is challenging.
Moreover, during the inference process, it requires the transformation of environment maps to spherical harmonics. 
To attain a specific lighting condition, it might be necessary to capture an environment map tailored exclusively to that particular condition. 
While NeRF is progressing towards becoming a more lightweight and faster model, employing environment maps hinders the efficiency of the combined task of NeRF and relighting.
Hence, particularly for daytime outdoor scenes, we present the feasibility of relighting without environment maps.

In this paper, we leverage the fact that sunlight is the most dominant light source in outdoor scenes.
Therefore, an effective approach to relighting in outdoor settings involves accurately generating shadows cast by the sun. 
In our pursuit of simultaneous novel view synthesis and relighting, we organize the synthesis process into three core components: \textit{albedo (color), shadow, and tone (brightness)}. 
The acquisition of albedo closely resembles the learning process for color in existing NeRF methods.
Shadows, governed by the sun direction and the surrounding geometry, are learned by taking the sun direction and shadow features as input, allowing the training of time-dependent shadows. 
While albedo and shadow already form the foundational elements of relighting, it is important to note that training images may exhibit varying tones even within the same time frame.
To disentangle the influence of tone on the learning of albedo and shadows, we introduce a latent vector that deals with tone throughout the training process.

Another key aspect of our model lies in its foundation architected atop TensoRF~\cite{chen2022tensorf}, which is known for its efficient training capabilities.
However, since TensoRF was not originally intended for relighting, its performance in relighting datasets falls short of satisfaction.
To address this limitation, we have devised an architecture based on TensoRF to the specific demands of the relighting task to achieve fast and effective relighting.
This tailored approach endows both our training and rendering procedures an appreciable acceleration, surpassing NeRF-OSR~\cite{rudnev2022nerf} by a substantial margin.

We summarize our contributions as follows:
\begin{itemize}[leftmargin=3mm] 
 \vspace{-3mm}
    \item \textbf{No need of environment map} \ Our model performs relighting based primarily on the sun direction, thereby eliminating the complexity of using an environment map in training and inference processes (see Fig.~\ref{fig:teaser}).
     \vspace{-3mm}
    \item \textbf{Handle unbounded scene} \ Our model employs cubemap based on TensoRF to successfully model unbounded scenes. 
     \vspace{-3mm}
    \item \textbf{Fast relighting and rendering} \  Our model achieves a speedup of around tenfold in training and inference processes compared to previous research.
\end{itemize}

\section{Related Works}
\label{sec:rel_works}

\subsection{Neural Radiance Field}
\label{subsec:nerf}
Neural Radiance Field (NeRF)~\cite{mildenhall2020nerf} has been mainstream for the novel view synthesis task thanks to its simple architecture and splendid quality of rendered images.
NeRF utilizes a Multi-Layer Perceptron (MLP) to depict a scene by associating 3D coordinates and viewing angles with colors and volume densities, and generates a novel view using the volume rendering technique.
Since its introduction, NeRF has been actively studied toward various directions, \eg, a multi-scale representation~\cite{barron2023zip, barron2021mip, isaac2023exact}, data efficiency~\cite{yu2021pixelnerf, seo2023mixnerf, niemeyer2022regnerf, seo2023flipnerf}, dynamic scene reconstruction~\cite{attal2021torf, tretschk2021non, li2021neural, pumarola2021d}, and so on.
Furthermore, there has been a line of research to address the challenging problems resulting from the fundamental limitations of NeRF's architecture, \eg, lengthy training time per scene~\cite{muller2022instant, chen2023mobilenerf, fridovich2022plenoxels, yu2021plenoctrees, chen2022tensorf}, incapacity to handle unbounded scenes~\cite{tancik2022block, turki2022mega, barron2022mip}, novel illumination conditions~\cite{srinivasan2021nerv, rudnev2022nerf, toschi2023relight, boss2021nerd, ling2023shadowneus}, and so on.
Among them, TensoRF~\cite{chen2022tensorf} represents a scene with a voxel feature grid decomposed into several low-rank tensor components, achieving a more compact scene representation.
In this paper, we address the representative bottlenecks altogether and propose SR-TensoRF, which performs relighting successfully for unbounded outdoor scenes with a significantly reduced training time compared to other baselines.

\begin{figure*}[t!]
  \centering
  \includegraphics[width=0.9\linewidth]{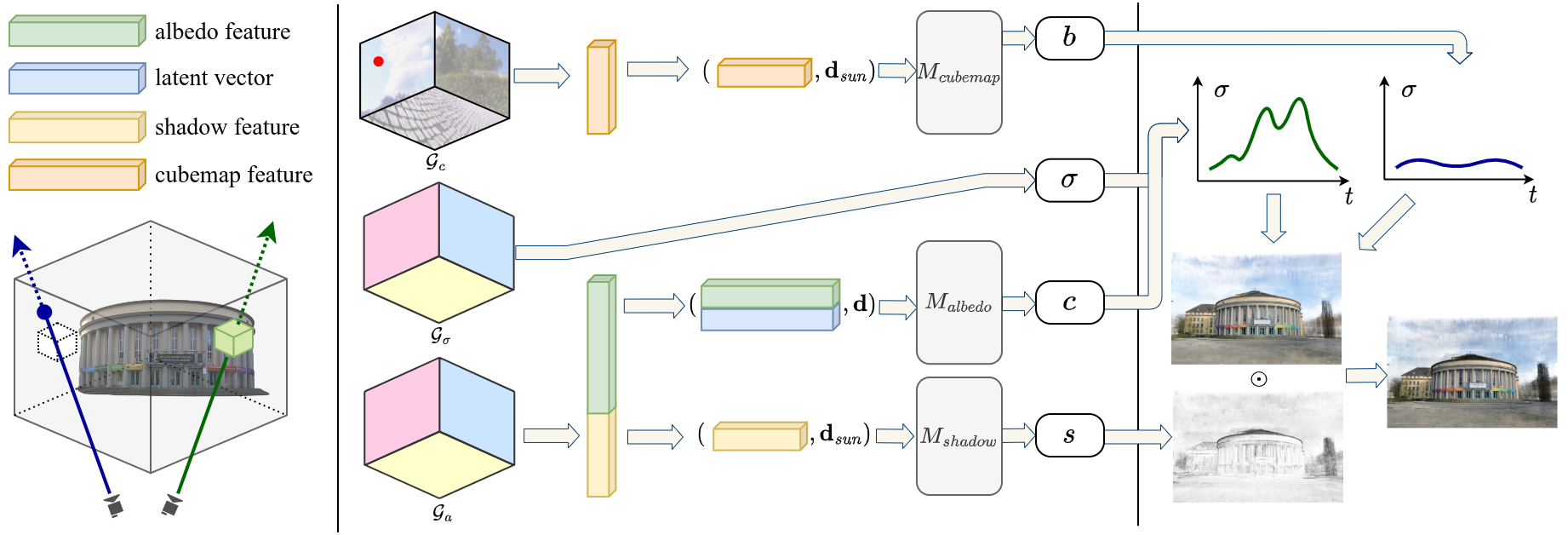}

  \caption{\textbf{Overview.} SR-TensoRF comprises three tensors: $\mathcal{G}_{\sigma}$, $\mathcal{G}_a$, and $\mathcal{G}_c$, among which $\mathcal{G}_{\sigma}$ learns features of density, $\mathcal{G}_a$ is for both albedo and shadow, and $\mathcal{G}_c$ is for cubemap, respectively.
  These components, along with the viewing direction $\mathbf{d}$, sun direction $\mathbf{d}_{sun}$, and latent vectors, are processed through shallow MLPs to generate albedo, shadow, and cubemap backgrounds.
  These generated components are then used in volumetric rendering to produce the final output, where the $\odot$ operator denotes pixel-wise multiplication. 
  The green ray depicts the standard volumetric rendering, while the blue ray represents a case where the accumulated transmittance of sampled points on the ray is low (small $\sigma$ over the entire ray-marching), necessitating the cubemap.
  }
  \label{fig:pipeline}
\end{figure*}

\subsection{Editing Neural Radiance Field}
\label{subsec:edit_nerf}
While NeRF achieves notable performance in generating photo-realistic novel views, editing them remains quite challenging since the 3D scene is represented by MLP parameters in its framework, which are hard to interpret and manipulate.
There exist various lines of research with regard to editing NeRF, \eg, scene stylization~\cite{haque2023instruct, huang2022stylizednerf, zhang2022arf, wu2022palettenerf}, object-level editing~\cite{yang2021learning, kholgade20143d, shetty2018adversarial}, and so on.
Among them, a series of studies~\cite{lyu2022neural, srinivasan2021nerv, jin2023tensoir, toschi2023relight} have addressed the task of controlling attributes of objects or scenes, \eg, illumination, material, and so on.
TensoIR~\cite{jin2023tensoir} expanded upon TensoRF~\cite{chen2022tensorf} to enable material editing and relighting with high-quality rendering.
However, it assumed a set of relatively simple scenes with controlled lighting or simple objects, which are far from real-world scenarios.
Our proposed SR-TensoRF empowers controlling on more complex unbounded outdoor scenes, all while leveraging the effective representation of TensoRF.

\subsection{Outdoor Scene Relighting}
\label{subsec:osr}
There have been various studies on outdoor scene relighting, emphasizing its importance in graphics and VR/AR.
Traditional researches \cite{troccoli2005relighting, haber2009relighting, duchene2015multi} optimized and computed variables without training.
Duchêne \etal\cite{duchene2015multi} used a 3D reconstruction with sun direction to segregate images into reflectance and shading, requiring manual shadow clicking to determine sun direction.
After the advent of deep learning, researches~\cite{philip2019multi, liu2020learning, carlson2019shadow, yu2020self} leveraged neural networks to achieve relighting. 
Yu \etal\cite{yu2020self} used a self-supervised method with a neural renderer and a dedicated network for shadow prediction, but it lacked accurate color representation due to the absence of color correction.

After NeRF's introduction, NeRF-OSR \cite{rudnev2022nerf} extended a novel view synthesis framework to handle outdoor scene relighting, which augmented NeRF's pipeline for learning shadow and normal decomposition.
However, it does not consider the characteristics of outdoor scenes and requires a lengthy training time. 
Unlike the previous methods, we enable a much more efficient pipeline by using solar information in the training process, skipping the pre-processing step of creating an environment map.

\section{Method}
\label{sec:method}

In this section, we present our approach, SR-TensoRF, to efficiently achieve simultaneous relighting and novel view synthesis in outdoor scenes, as shown in Figure~\ref{fig:pipeline}.
SR-TensoRF is designed based on TensoRF~\cite{chen2022tensorf}, which employs 4D tensor decomposition to model voxel grids (Sec.~\ref{subsec:preliminary}).
Instead of environment maps, we leverage sun direction directly, accounting for its variation across time and location, to facilitate relighting (Sec.~\ref{subsec:sun_direction}). 
For unbounded scenes not encompassed within TensoRF's architecture, we incorporate cubemaps (Sec.~\ref{subsec:unbounded}). 
Furthermore, we utilize latent vectors to disentangle the non-uniform color tones present in the training images (Sec.~\ref{subsec:latent}). 
We optimize SR-TensoRF using reconstruction loss, TensoRF's training loss, and shadow regularization loss (Sec.~\ref{subsec:loss}).

\subsection{Preliminary}
\label{subsec:preliminary}
\subsubsection{Neural Radiance Field}
NeRF~\cite{mildenhall2020nerf} employs a neural network to depict a 3D scene with the 3D coordinates of a point $\mathbf{x}$ and the viewing direction $\mathbf{d}$ as input, producing the color $\mathbf{c}$ and density $\sigma$ of the points as output.
The input 3D points are sampled along a ray $\mathbf{r}(t) = \mathbf{o} + t\mathbf{d}$, and each ray is cast toward a pixel in training images. 
The outputs, namely the color $\mathbf{c}$ and density $\sigma$, are then utilized to render a pixel color $\hat{\mathbf{C}}(\mathbf{r})$:
\begin{equation}
\begin{gathered}
    \hat{\mathbf{C}}(\mathbf{r}) = \sum_{i=1}^{N} T_{i}(1-\exp (-\sigma_{i} \delta_{i})) \mathbf{c}_{i}  = \sum_{i=1}^{N} w_i \mathbf{c}_{i}\\ 
    w_i = T_{i}(1-\exp (-\sigma_{i} \delta_{i})), \ 
    T_{i}=\exp(-\sum_{j=1}^{i-1} \sigma_{j} \delta_{j})
\end{gathered}
\label{equ:rendering}
\end{equation}
where $N$ is the number of sample points, and $\delta_i$ and $T_{i}$ are the step size of a ray and the accumulated transmittance at each sampled point $\mathbf{r}(t_i)$, respectively.

\subsubsection{Tensorial Radiance Fields (TensoRF)}
TensoRF~\cite{chen2022tensorf} represents a 3D scene with a 3D geometry tensor $\mathcal{G}_\sigma$ and a 4D appearance tensor $\mathcal{G}_a$, where an additional dimension corresponds to the 3D point's 1D appearance feature.
The element at each 3D position $\mathbf{x}$ of $\mathcal{G}_\sigma$ is a density $\sigma (\mathbf{x})$ while that of $\mathcal{G}_a$ is a 1D feature vector. This feature vector, along with the viewing direction $\mathbf{d}$, is fed into a shallow MLP or a spherical harmonics function $M$, ultimately generating the color $\mathbf{c}$:
\begin{equation}
    \sigma, \mathbf{c} = \mathcal{G}_\sigma(\mathbf{x}), M(\mathcal{G}_a(\mathbf{x}), \mathbf{d}).
\label{equ:tensor}
\end{equation}

The key advantage of TensoRF lies in its lightweight model architecture and rapid training speeds. 
This is achieved by employing an alpha mask for densities below a specific threshold, ensuring that only the minimum required number of remaining voxels is retained. 
The voxel-wise alpha value is computed as $\alpha = 1 - \exp(-\sigma\delta)$.
For more architectural details, kindly refer to the paper of TensoRF.

\begin{figure}[t!]
  \centering
   \includegraphics[width=0.8\linewidth]{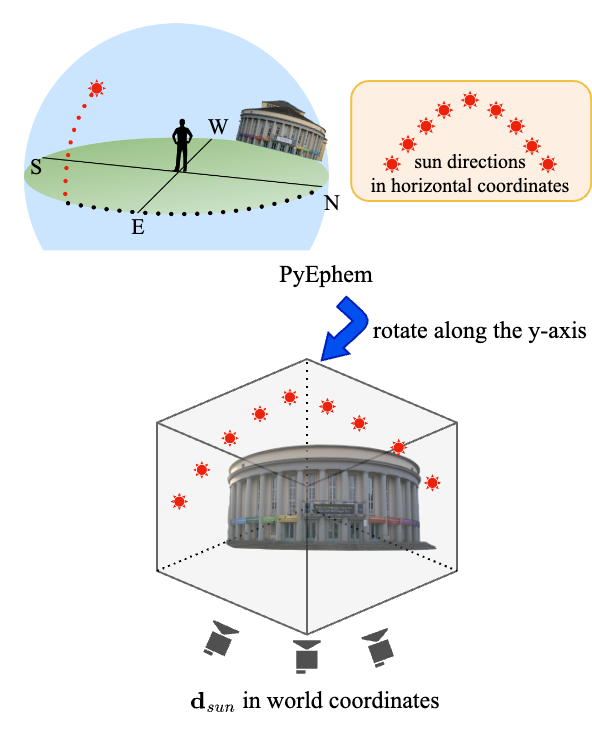}
   \caption{\textbf{Calculating the sun direction from time and location information.} Using PyEphem~\cite{ephem}, we get the sun directions from the local observer's coordinates and subsequently align them with the first camera's coordinates.} 
   \label{fig:sundirection}
\end{figure}

\subsection{Sun Light Direction as Input}
\label{subsec:sun_direction}
For successful daytime outdoor relighting, it is essential to generate shadows caused by sunlight accurately. 
These shadows are primarily determined by the sun direction and are influenced by the surrounding geometry while remaining independent of the viewing direction.
Taking these aspects into consideration, we organize the process of generating shadows in two steps: the first involves learning \textit{geometry} using a shadow feature from a 4D appearance tensor, and the second is using the \textit{sun direction as input} of the shading process.

We implement the first step by making the 4D appearance tensor $\mathcal{G}_a$ incorporate not only the albedo but also the shadow, \ie, $\mathcal{G}_a(\mathbf{x}) = \mathcal{G}_{albedo}(\mathbf{x}) \oplus \mathcal{G}_{shadow}(\mathbf{x})$.
The shadow tensor $\mathcal{G}_{shadow}$ is intended to depict intricate shadowing conditions influenced by the surrounding geometry. Later, in the second step, the shadow feature, $\mathcal{G}_{shadow}(\mathbf{x})$, is input into the shadow MLP $M_{shadow}$ alongside the sun direction $\mathbf{d}_{sun}$ as follows:
\begin{equation}
\begin{gathered}
    \sigma = \mathcal{G}_\sigma(\mathbf{x}) \\
    \mathbf{c} = M_{albedo}(\mathcal{G}_{albedo}(\mathbf{x}), \mathbf{d})\\ 
    s = M_{shadow}(\mathcal{G}_{shadow}(\mathbf{x}), \mathbf{d}_{sun}) 
\end{gathered}
\label{equ:suninput}
\end{equation}
where $s \geq 0$ indicates the strengh of shadow at 3D sampled point $\mathbf{x}$.
The $\sigma$, $\mathbf{c}$, and $s$ obtained from Eq. (\ref{equ:suninput}) are used to volume render albedo $\mathbf{A}(\mathbf{r})$ with $(\sigma, \mathbf{c})$ and similarly, to render shadow $S(\mathbf{r}) \in [0,1]$ with $(\sigma, s)$ as in Eq. (\ref{equ:rendering}). Subsequently, the two results are multiplied pixel-wise to generate the final output $\hat{C}(\mathbf{r})$.

The sun direction $\mathbf{d}_{sun}$ can be computed from the date, time, and location coordinates of the captured images using PyEphem~\cite{ephem} as in Fig.~\ref{fig:sundirection}. 
Once determined, this sun direction, being in the horizontal coordinates, which uses the observer's local horizon as the x-z plane, must be aligned with the world coordinates of the training images. 
Given the convention of the NeRF camera pose where the camera pose of the first training image serves as the world coordinate, the sun direction is rotated along the y-axis based on how much the first training image deviates from the south.

\subsection{Cubemap to Handle Unbounded Scenes}
\label{subsec:unbounded}
TensoRF encounters two types of challenges when reconstructing unbounded scenes due to its limited ability to effectively utilize alpha masks.
When alpha masks do not mask out densities sufficiently, it can lead to an elongated training time, the unintentional learning of extraneous densities, and, consequently, the emergence of artifacts.
On the other hand, when alpha masks function excessively, it can result in cracks in the necessary background regions of unbounded scenes. 
Hence, we incorporate the concept of cubemaps in computer graphics to handle backgrounds into TensoRF. 
Cubemaps are beneficial at capturing missing background information, resembling the role of background networks, especially as seen in prior NeRF research like NeRF++~\cite{zhang2020nerf++}. 
However, since TensoRF parametrizes scenes using cuboid shapes, cubemaps offer a more seamless and suitable background.

To integrate our proposed cubemap, we add background to the volume rendering equation (\ref{equ:rendering}):
\begin{equation}
\begin{gathered}
    \hat{\mathbf{C}}(\mathbf{r}) = \sum_{i=1}^{N} w_i \mathbf{c}_{i} + (1-w_i)\mathbf{b} \\ 
    \mathbf{b} = M_{cubemap}(\mathcal{G}_{c}(\mathbf{r}), \mathbf{d}_{sun}), \\
\end{gathered}
\end{equation}
where $\mathbf{b}$ is the cubemap background, $M_{cubemap}$ is a shallow MLP, and $\mathcal{G}_{c}(\mathbf{r})$ is the cubemap feature bilinearly interpolated from where the ray intersects with one of the six 3D cubemap tensors $\mathcal{G}_c$.
$M_{cubemap}$ is conditioned by the sun direction since background colors might vary based on it.
Furthermore, to address the inherent TensoRF's inability to model unbounded scenes, we have integrated the ray contraction technique from MipNeRF-360~\cite{barron2022mip}.

\subsection{Latent Vector for Tone Disentanglement}
\label{subsec:latent}
In outdoor scenes, it is infeasible to capture training images under controlled conditions, resulting in variations in tone, \ie, white balance and brightness, across shots.
The presence of such variations in the training images could lead to inconstancy in the ground truth during training, potentially leading to failed training and artifacts.
To mitigate this, we integrate these variations into a latent vector, thereby encouraging the training process to disentangle albedo and tone.
As in \cite{martin2021nerf}, we pretrain the latent vectors using the approach of Generative Latent Optimization (GLO)~\cite{bojanowski2018optimizing}. This ensures that each image is assigned to a latent vector $l$, which is used as an input to $M_{albedo}$ and $\mathbf{c}$ in Eq.(\ref{equ:suninput}) is replaced with Eq. (\ref{equ:suninput2}):
\begin{equation}
    \mathbf{c} = M_{albedo}(\mathcal{G}_{albedo}(\mathbf{x}), l, \mathbf{d}). 
    \label{equ:suninput2}
\end{equation}

\subsection{Loss Function}
\label{subsec:loss}
Our SR-TensoRF is optimized based on Mean Squared Error (MSE) loss, L1 loss for density tensor, and Total Variation (TV) loss for all tensors we use.
Additionally, a regularization loss is incorporated to constrain shadow values from approaching 1.
Please refer to the supplementary materials regarding the weights of the loss terms.

\section{Experiments}
\label{sec:experiments}

\subsection{Experimental Details}
\label{subsec:details}
\paragraph{Implementation details.}
We implement our model based on TensoRF~\cite{chen2022tensorf}. 
We utilize shallow MLPs for all three $M$ components ($M_{albedo}, M_{shadow}$ and $M_{cubemap}$). 
The numbers of voxels for the density and appearance tensors are the same as in the original TensoRF. 
The number of elements for the cubemap tensor is fixed at 128$\times$128 for a single facet.
For optimization, we use the Adam optimizer~\cite{kingma2014adam} with a learning rate of 0.01 and a decay of 0.2 for tensor parameters, while the MLPs employed an initial learning rate of 0.001.
To prevent overfitting of the cubemap during training, we withhold from integrating the cubemap for the initial 5,000 iterations.

\paragraph{Datasets and metrics.}
We evaluate our approach on NeRF-OSR dataset~\cite{rudnev2022nerf}, which contains eight outdoor scene images in total. We show our results for three of them, as in \cite{rudnev2022nerf}. 
For the evaluation metrics, we adopt PSNR, Structural Similarity index (SSIM)~\cite{wang2004image}, Mean Absolute Error (MAE), and Mean Squared Error (MSE) of entire test images.
For quantitative evaluation, since NeRF-OSR dataset does not include masks, we generate them using the segmentation model SegFormer~\cite{xie2021segformer} and exclude the sky and tree regions from them.
Also, we follow the metric computation as in NeRF-OSR.
Regarding the latent vectors, as in \cite{martin2021nerf}, we optimize them on the left half of each image in the test set and compute the metrics both on the right half (Tab.~\ref{tab:half}) and the whole image (Tab.~\ref{tab:quantitative}).

\begin{figure*}
  \centering
  \begin{subfigure}{1.0\linewidth}
    \centering
    \resizebox{0.9\linewidth}{!}{
    \begin{tabular}{{>{\centering\arraybackslash}m{0.14\linewidth}m{0.22\linewidth}m{0.22\linewidth}m{0.22\linewidth}m{0.22\linewidth}}}
    NeRF-OSR~\cite{rudnev2022nerf} &
    \includegraphics[width=1.0\linewidth]{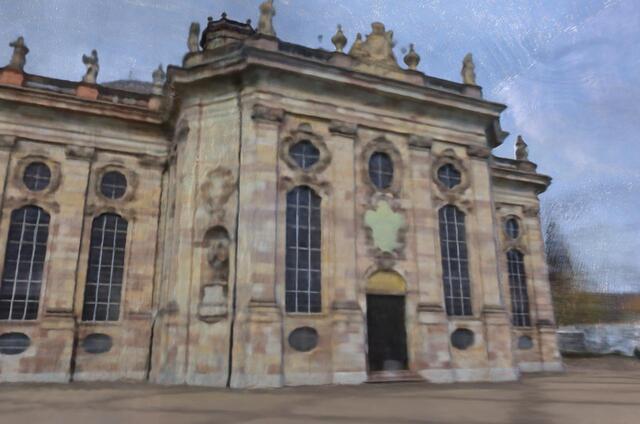} & 
    \includegraphics[width=1.0\linewidth]{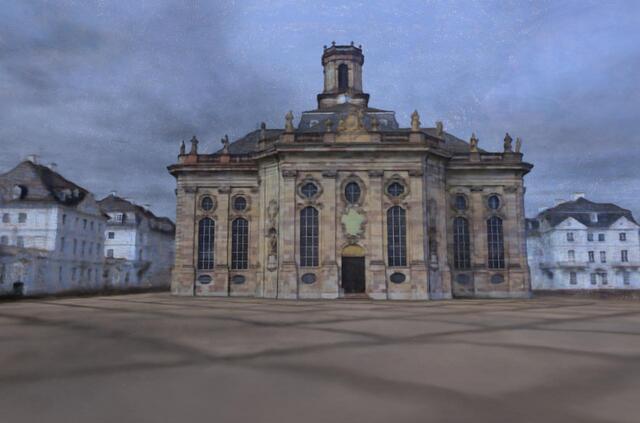} & 
    \includegraphics[width=1.0\linewidth]{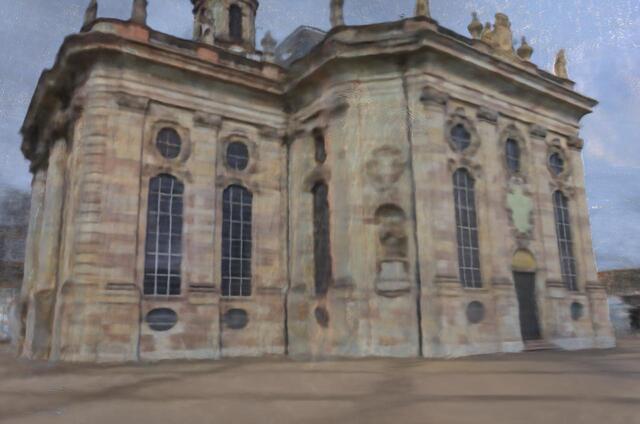} & 
    \includegraphics[width=1.0\linewidth]{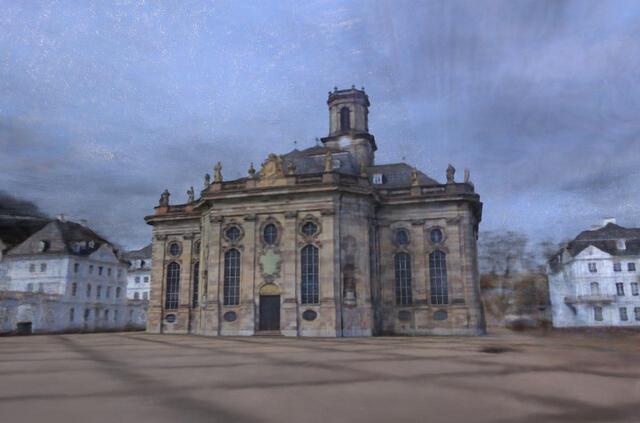} \\
    Ours &
    \includegraphics[width=1.0\linewidth]{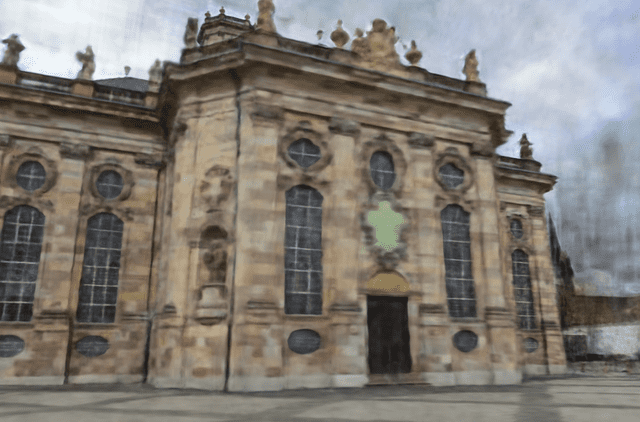} & 
    \includegraphics[width=1.0\linewidth]{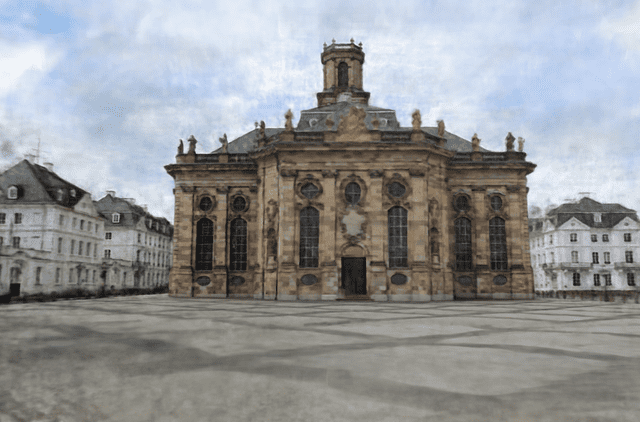} & 
    \includegraphics[width=1.0\linewidth]{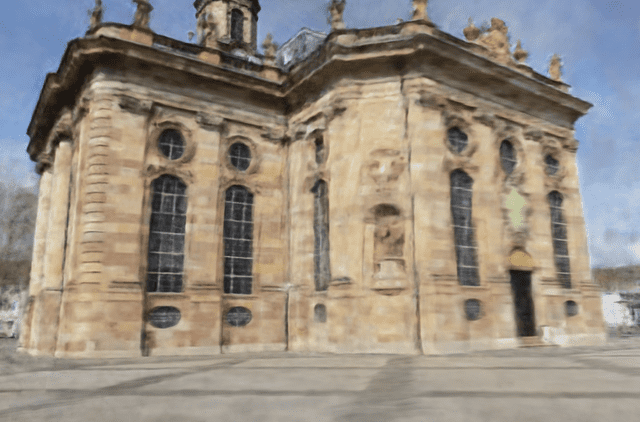} & 
    \includegraphics[width=1.0\linewidth]{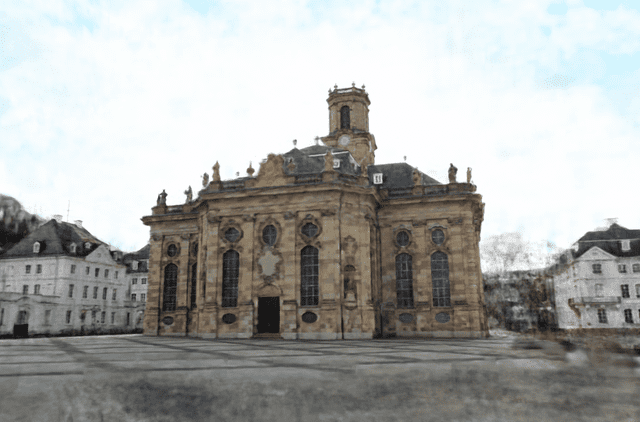} \\
    GT &
    \includegraphics[width=1.0\linewidth]{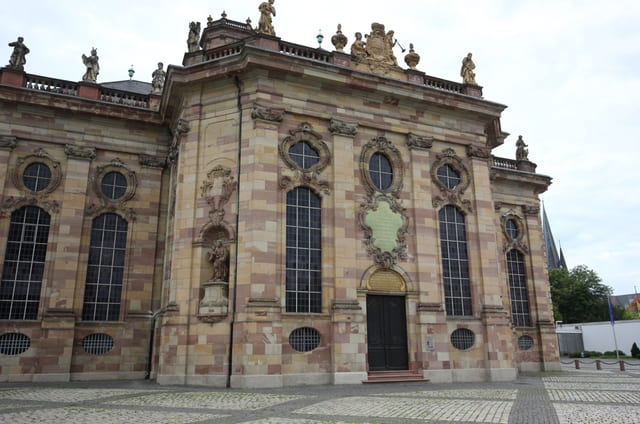} & 
    \includegraphics[width=1.0\linewidth]{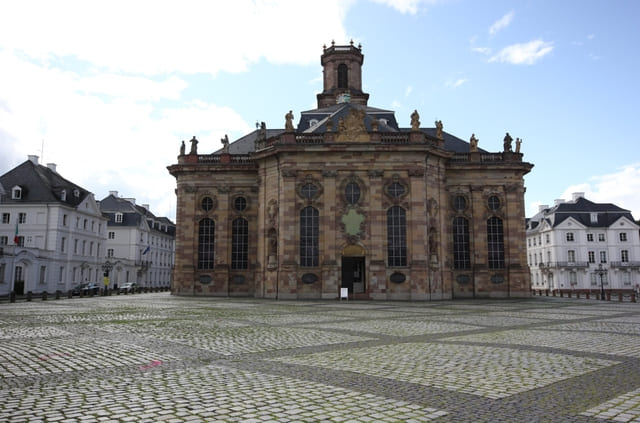} & 
    \includegraphics[width=1.0\linewidth]{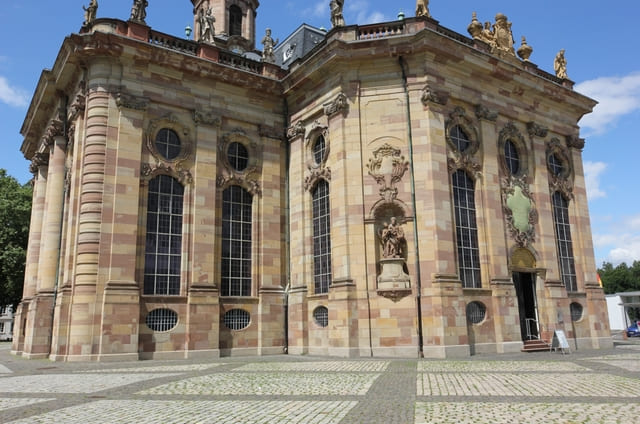} & 
    \includegraphics[width=1.0\linewidth]{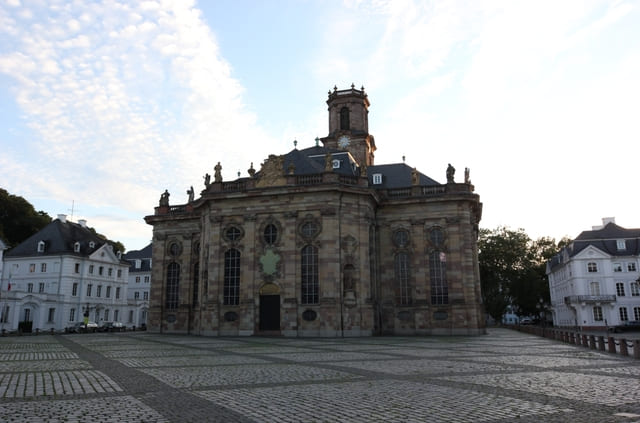} \\
    \end{tabular}
  }
    \caption{Site 1}
    \label{fig:qualitative-a}
  \end{subfigure}
  \begin{subfigure}{1.0\linewidth}
    \centering
    \resizebox{0.9\linewidth}{!}{
    \begin{tabular}{{>{\centering\arraybackslash}m{0.14\linewidth}m{0.22\linewidth}m{0.22\linewidth}m{0.22\linewidth}m{0.22\linewidth}}}
    NeRF-OSR~\cite{rudnev2022nerf} &
    \includegraphics[width=1.0\linewidth]{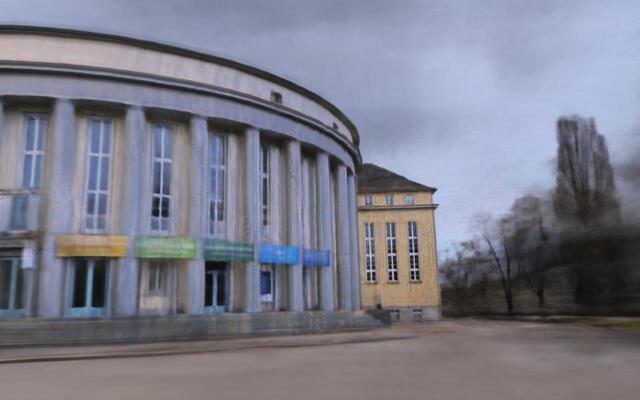} & 
    \includegraphics[width=1.0\linewidth]{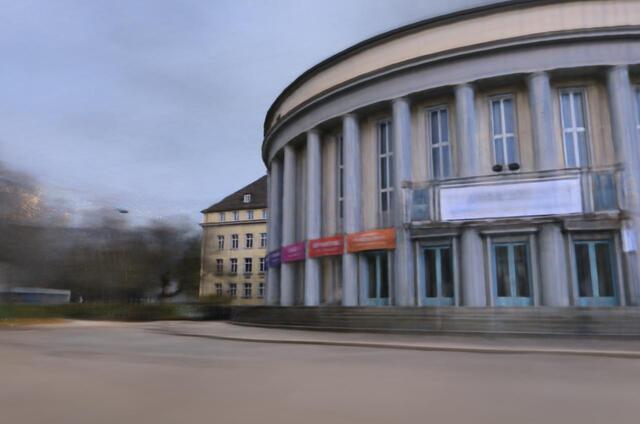} & 
    \includegraphics[width=1.0\linewidth]{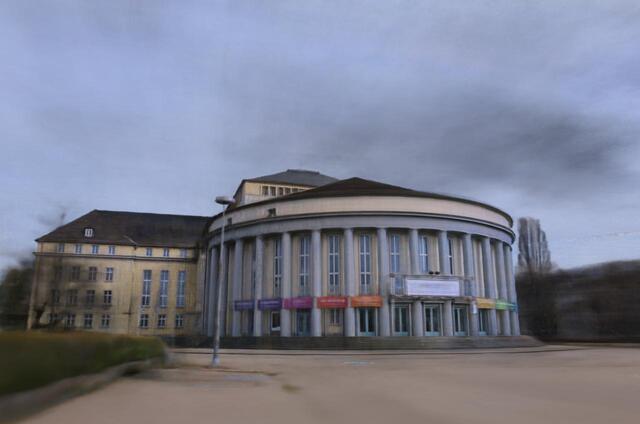} & 
    \includegraphics[width=1.0\linewidth]{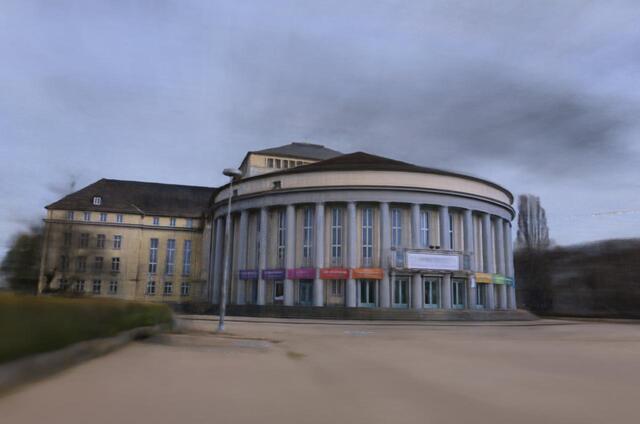} \\
    Ours &
    \includegraphics[width=1.0\linewidth]{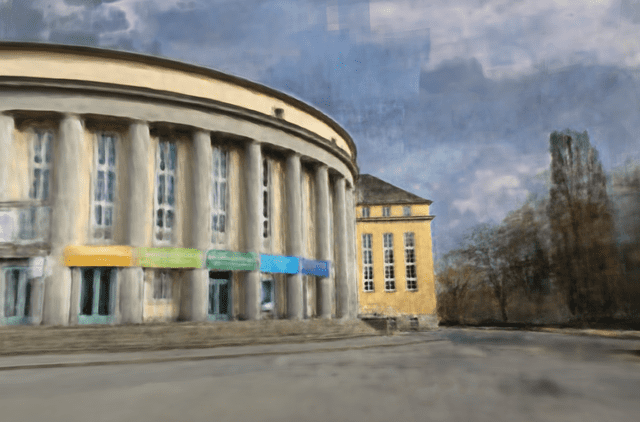} & 
    \includegraphics[width=1.0\linewidth]{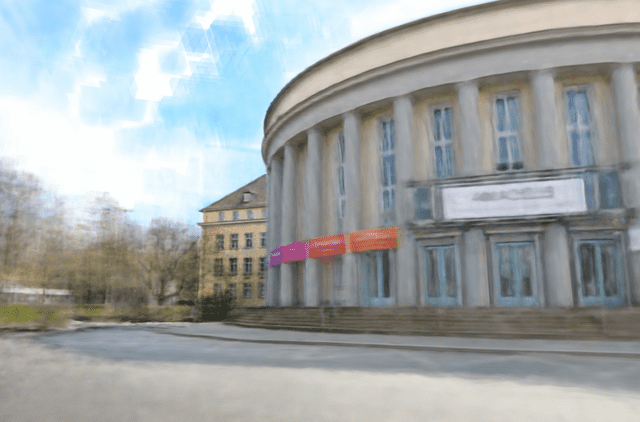} & 
    \includegraphics[width=1.0\linewidth]{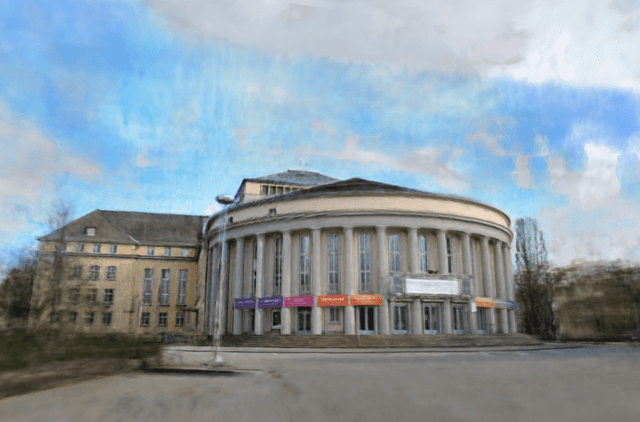} & 
    \includegraphics[width=1.0\linewidth]{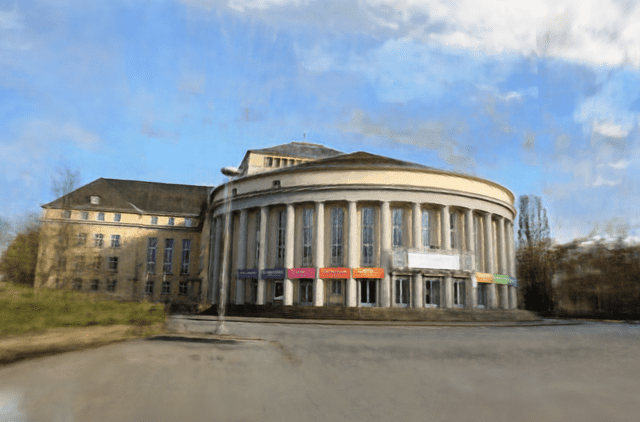} \\ 
    GT &
    \includegraphics[width=1.0\linewidth]{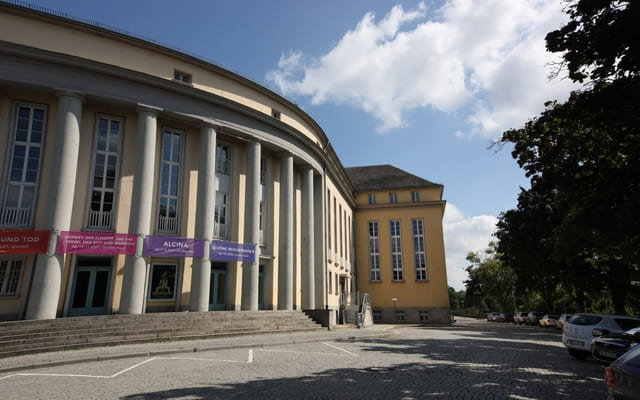} & 
    \includegraphics[width=1.0\linewidth]{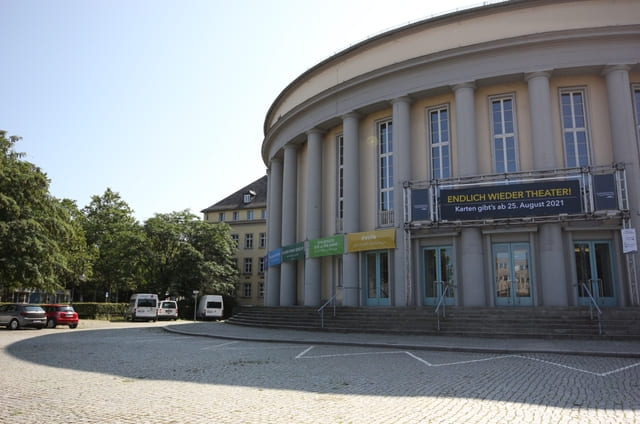} & 
    \includegraphics[width=1.0\linewidth]{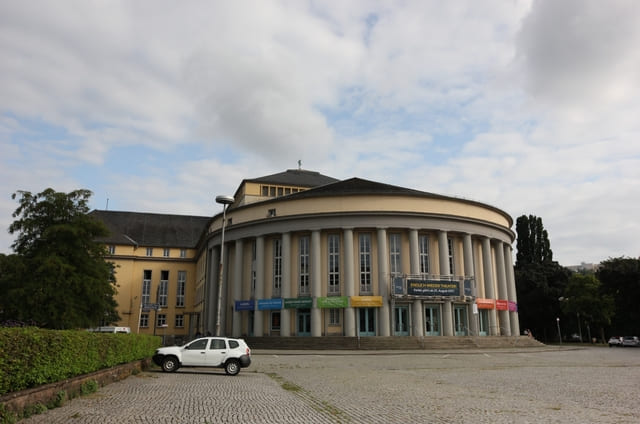} & 
    \includegraphics[width=1.0\linewidth]{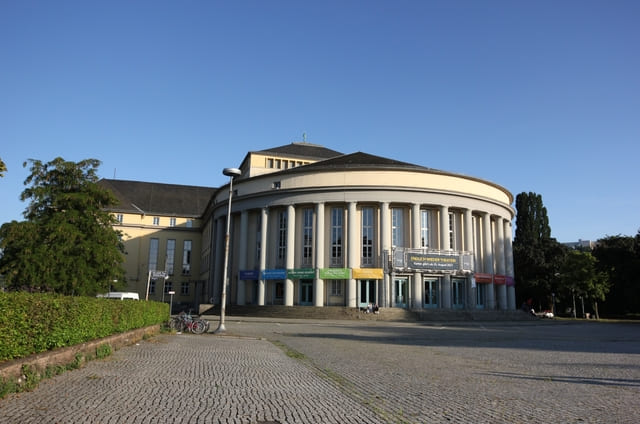} \\
    \end{tabular}
  }
    \caption{Site 2}
    \label{fig:qualitative-b}
  \end{subfigure}
  \begin{subfigure}{1.0\linewidth}
    \centering
    \resizebox{0.9\linewidth}{!}{
    \begin{tabular}{{>{\centering\arraybackslash}m{0.14\linewidth}m{0.22\linewidth}m{0.22\linewidth}m{0.22\linewidth}m{0.22\linewidth}}}
    NeRF-OSR~\cite{rudnev2022nerf} &
    \includegraphics[width=1.0\linewidth]{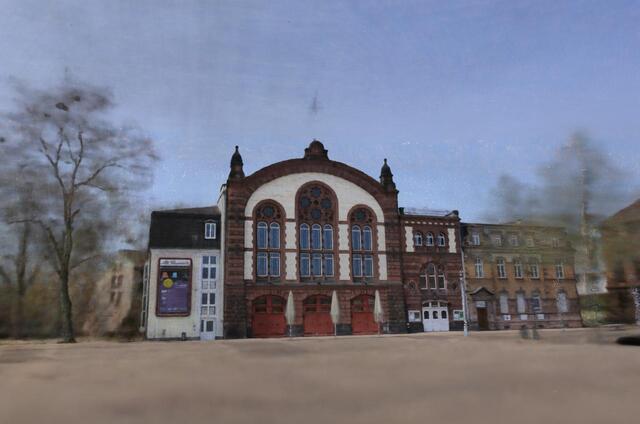} & 
    \includegraphics[width=1.0\linewidth]{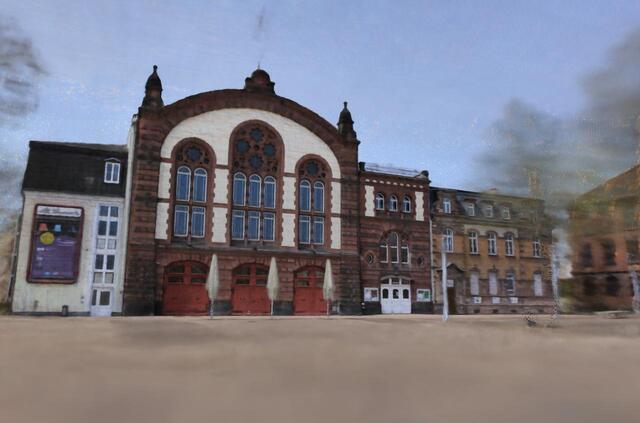} & 
    \includegraphics[width=1.0\linewidth]{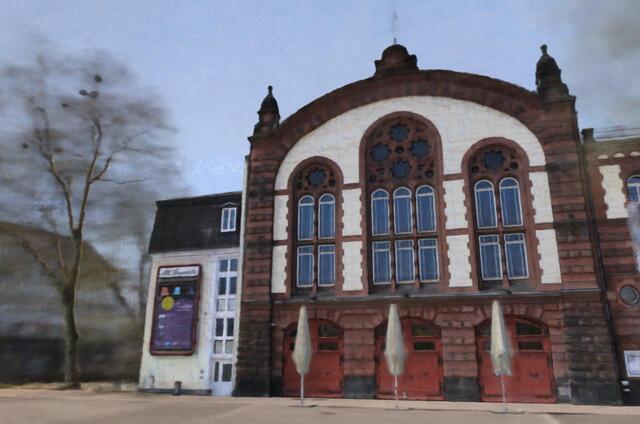} & 
    \includegraphics[width=1.0\linewidth]{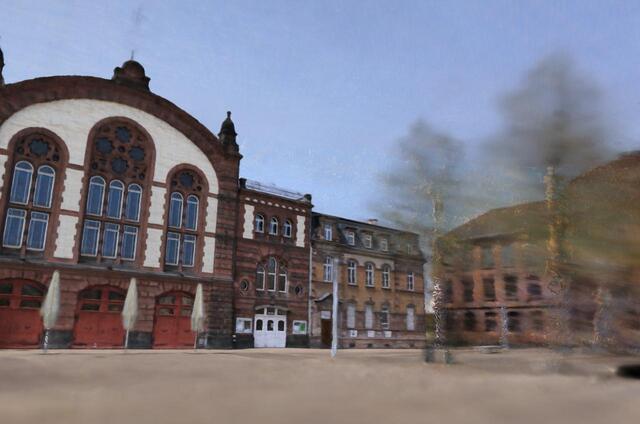} \\
    Ours &
    \includegraphics[width=1.0\linewidth]{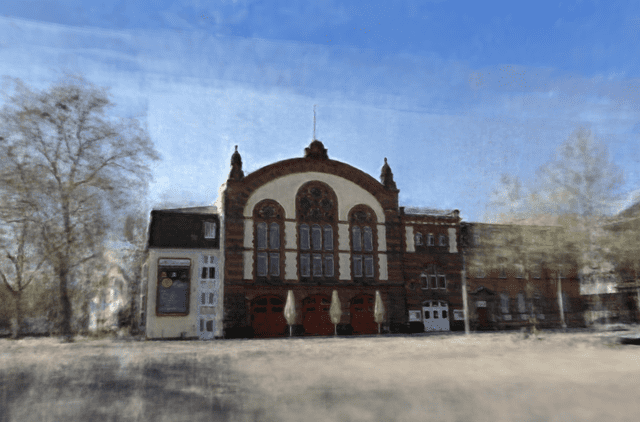} & 
    \includegraphics[width=1.0\linewidth]{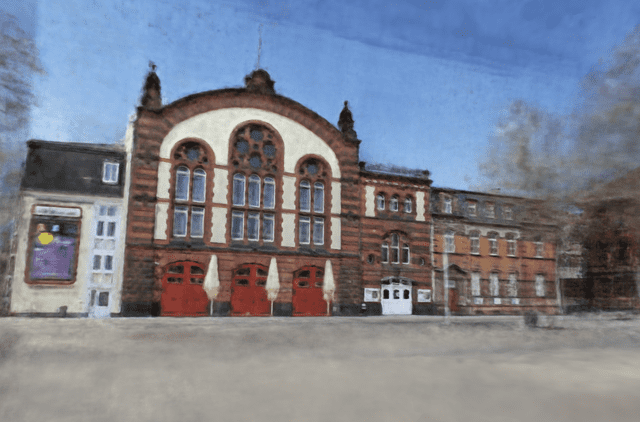} & 
    \includegraphics[width=1.0\linewidth]{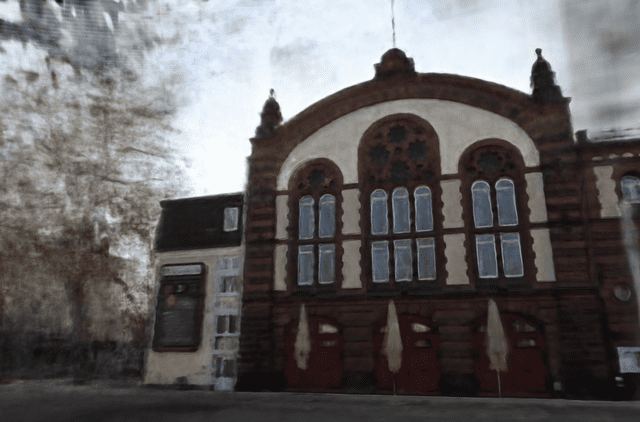} & 
    \includegraphics[width=1.0\linewidth]{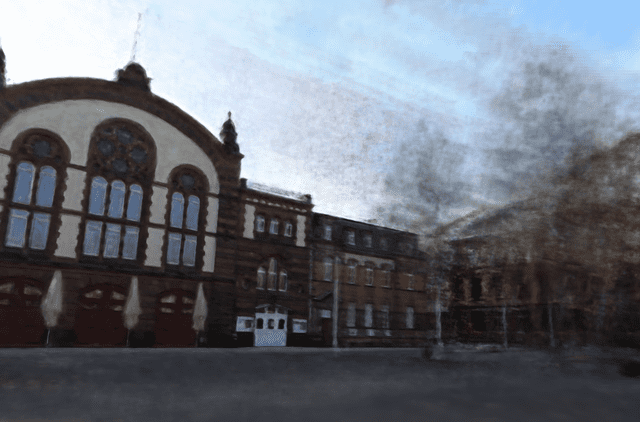} \\
    GT &
    \includegraphics[width=1.0\linewidth]{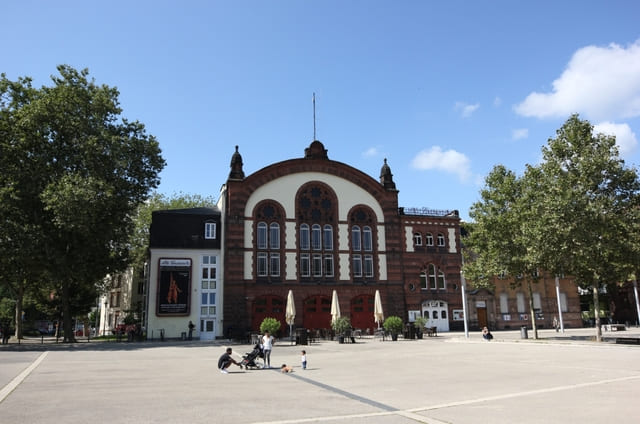} & 
    \includegraphics[width=1.0\linewidth]{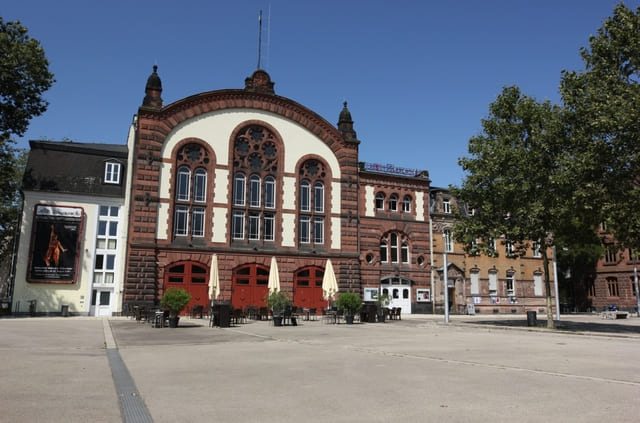} & 
    \includegraphics[width=1.0\linewidth]{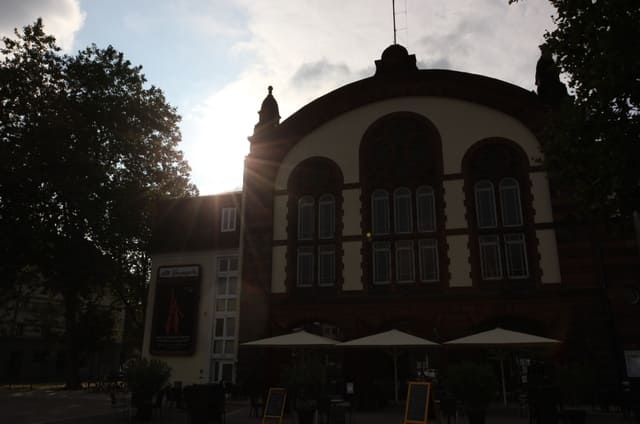} & 
    \includegraphics[width=1.0\linewidth]{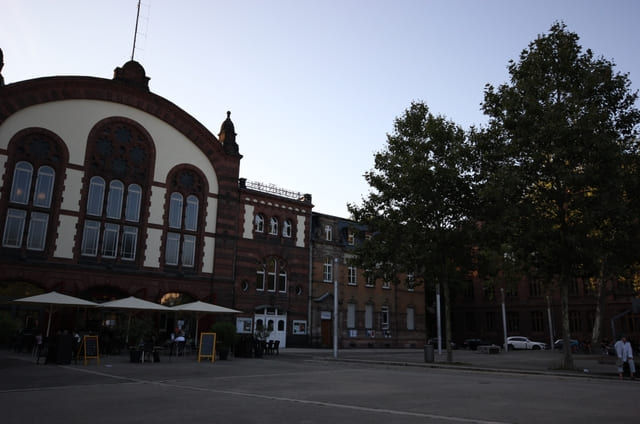} \\
    \end{tabular}
  }   
  \caption{Site 3}
    \label{fig:qualitative-c}
  \end{subfigure}
  \caption{\textbf{Qualitative Results.} We visualize novel view synthesis results from four different views. Regarding the lighting of the test set images, our method aligns the time and utilizes a latent vector extracted from the left half of the images. In contrast, NeRF-OSR utilizes default lighting and only conducts novel view synthesis without applying relighting to the test set images. Hence, in these results, the color of the sky remains consistent across all views. Additionally, the shadows cast by buildings on the ground or themselves are much more distinct in our results, highlighting the enhanced clarity in our rendering.
}
  \label{fig:qualitative}
\end{figure*}

\paragraph{Baselines.}
We compare our approach with NeRF-OSR~\cite{rudnev2022nerf}, the state-of-the-art model and the only existing model to handle novel view synthesis and relighting simultaneously.
\yr{We also compare with the relighting models~\cite{yu2020self, philip2019multi} that do not perform novel view synthesis.} 
These values are from the NeRF-OSR~\cite{rudnev2022nerf} paper.
Note that, in the case of Site 1, only 104 out of 160 training images have information on capture time, thus we trained our model with those 104 images. 
For a fair comparison, we also trained NeRF-OSR with this image set.

\begin{table}
  \centering
  \resizebox{0.95\linewidth}{!}{
    \begin{tabular}{{l|cccc}}
    Method & PSNR $\uparrow$ & SSIM $\uparrow$ & MSE $\downarrow$ & MAE $\downarrow$\\
    \midrule
     & \multicolumn{4}{c}{Site 1} \\
    \midrule
    Yu~\etal~\cite{yu2020self} & 17.87 & 0.378 & \underline{0.017} & 0.097 \\
    Philip~\etal~\cite{philip2019multi} & 16.63 & 0.367 & 0.023 & 0.113  \\
    NeRF-OSR~\cite{rudnev2022nerf} & \textbf{18.72} & 0.468 & \textbf{0.014} & \textbf{0.090} \\
    NeRF-OSR$^\dagger$~\cite{rudnev2022nerf} & \underline{18.32} & \textbf{0.623} & 0.019 & \underline{0.091} \\
    NeRF-OSR$^\ddagger$~\cite{rudnev2022nerf} & 13.33 & 0.524 & 0.059 & 0.177 \\
    Ours & 17.30 & \underline{0.542} & 0.021 & 0.096 \\
    \midrule
     & \multicolumn{4}{c}{Site 2} \\
    \midrule
    Yu~\etal~\cite{yu2020self} & 15.28 & 0.385 & 0.032 & 0.138 \\
    Philip~\etal~\cite{philip2019multi} & 12.34 & 0.272 & 0.065 & 0.2  \\
    NeRF-OSR~\cite{rudnev2022nerf} & \underline{15.43} & \underline{0.517} & \textbf{0.029} & \underline{0.133} \\
    Ours & \textbf{15.63} & \textbf{0.632} & \underline{0.030} & \textbf{0.111} \\
    \midrule
     & \multicolumn{4}{c}{Site 3} \\
    \midrule
    Yu~\etal~\cite{yu2020self} & 15.17 & 0.376 & 0.033 & 0.133 \\
    Philip~\etal~\cite{philip2019multi} & 12.28 & 0.319 & 0.062 & 0.179  \\
    NeRF-OSR~\cite{rudnev2022nerf} & \underline{16.65} & \underline{0.501} & \textbf{0.024} & \underline{0.114} \\
    Ours & \textbf{16.74} & \textbf{0.653} & \textbf{0.024} & \textbf{0.093}
    \end{tabular}
  }
  \caption{\textbf{Quantitative Results.} Our SR-TensoRF is competitive with other methods.
  $\dagger$ use the same 104 training images as ours for training, and $\ddagger$ use the test set environment map for evaluation.
  The results of NeRF-OSR are taken from the original paper, while those of $\dagger$ and $\ddagger$ are computed following the methodology in the NeRF-OSR paper. Note that the reported metrics may exhibit slight variations due to the absence of publicly available masks.
  }
  \label{tab:quantitative}
\end{table}

\begin{table}
  \centering
  \resizebox{0.9\linewidth}{!}{
  \begin{tabular}{{l|ccccc}}
    Method & PSNR $\uparrow$ & SSIM $\uparrow$ & MSE $\downarrow$ & MAE $\downarrow$ & Time \\
    \midrule
     & \multicolumn{4}{c}{Site 1} \\
    \midrule
    NeRF-OSR~\cite{rudnev2022nerf} & 18.30 & 0.627 & 0.018 & 0.089 & 32h \\
    Ours & 17.60 & 0.557 & 0.020 & 0.092 & 1h 11m \\
    \midrule
     & \multicolumn{4}{c}{Site 2} \\
    \midrule
    NeRF-OSR~\cite{rudnev2022nerf} & 16.16 & 0.691 & 0.029 & 0.107 & 61h\\
    Ours & 15.83 & 0.641 & 0.032 & 0.114 & 3h 41m \\
    \midrule
     & \multicolumn{4}{c}{Site 3} \\
    \midrule
    NeRF-OSR~\cite{rudnev2022nerf} & 16.49 & 0.707 & 0.025 & 0.095 & 58h\\
    Ours & 17.27 & 0.667 & 0.022 & 0.087 & 2h 46m
    \end{tabular}
  }
  \caption{\textbf{Results on Half Images.} Evaluation on the right half, which is not used for the latent vector optimization, yields similar quantitative results.}
  \label{tab:half}
\end{table}

\subsection{Analysis of SR-TensoRF}
\label{sec:analysis}

\paragraph{Benefit of sun direction input.}
Unlike methods that utilize environment maps for relighting, our approach eliminates the need for preprocessing steps like environment map creation and, if necessary, its conversion to spherical harmonics (as shown in Fig.~\ref{fig:teaser}).
Additionally, even the captured coordinates, time, and date are readily available metadata in mobile environments, incurring minimal associated costs.

\paragraph{Benefit of TensoRF.}
As shown in Tab.~\ref{tab:half}, SR-TensoRF achieves notably faster training than NeRF-OSR~\cite{rudnev2022nerf} thanks to our foundation on the TensoRF~\cite{chen2022tensorf} framework.
This extends to the inference process as well.
While this relative time efficiency is not directly transferrable to unbounded scenes, we successfully leverage the benefits of TensoRF with the assistance of cubemaps.

\begin{figure}[t!]
  \centering
  \begin{subfigure}{0.32\linewidth}
    \centering
    \includegraphics[width=1.0\linewidth]{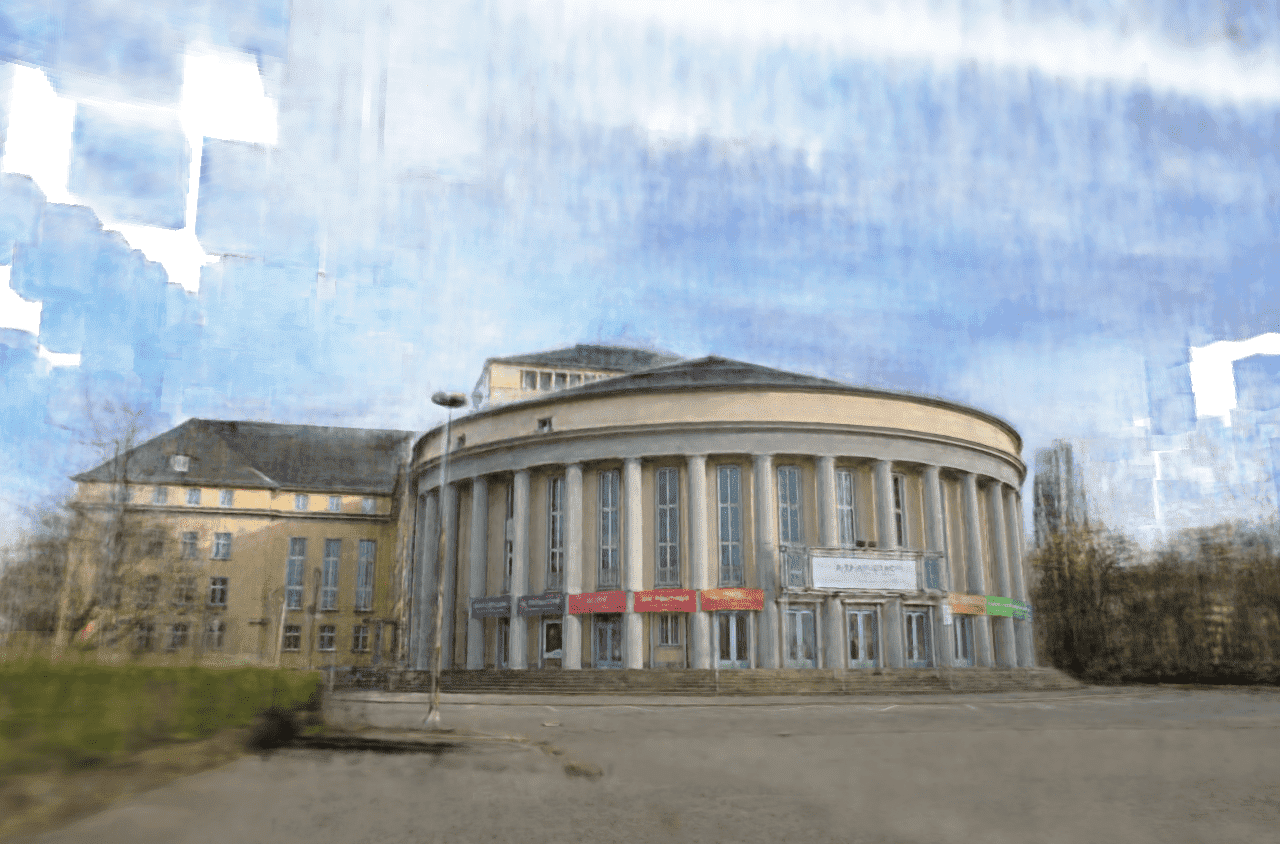}
    \caption{w/o cubemap}
    \label{fig:cubemap-a}
  \end{subfigure}
  \begin{subfigure}{0.32\linewidth}
    \centering
    \includegraphics[width=1.0\linewidth]{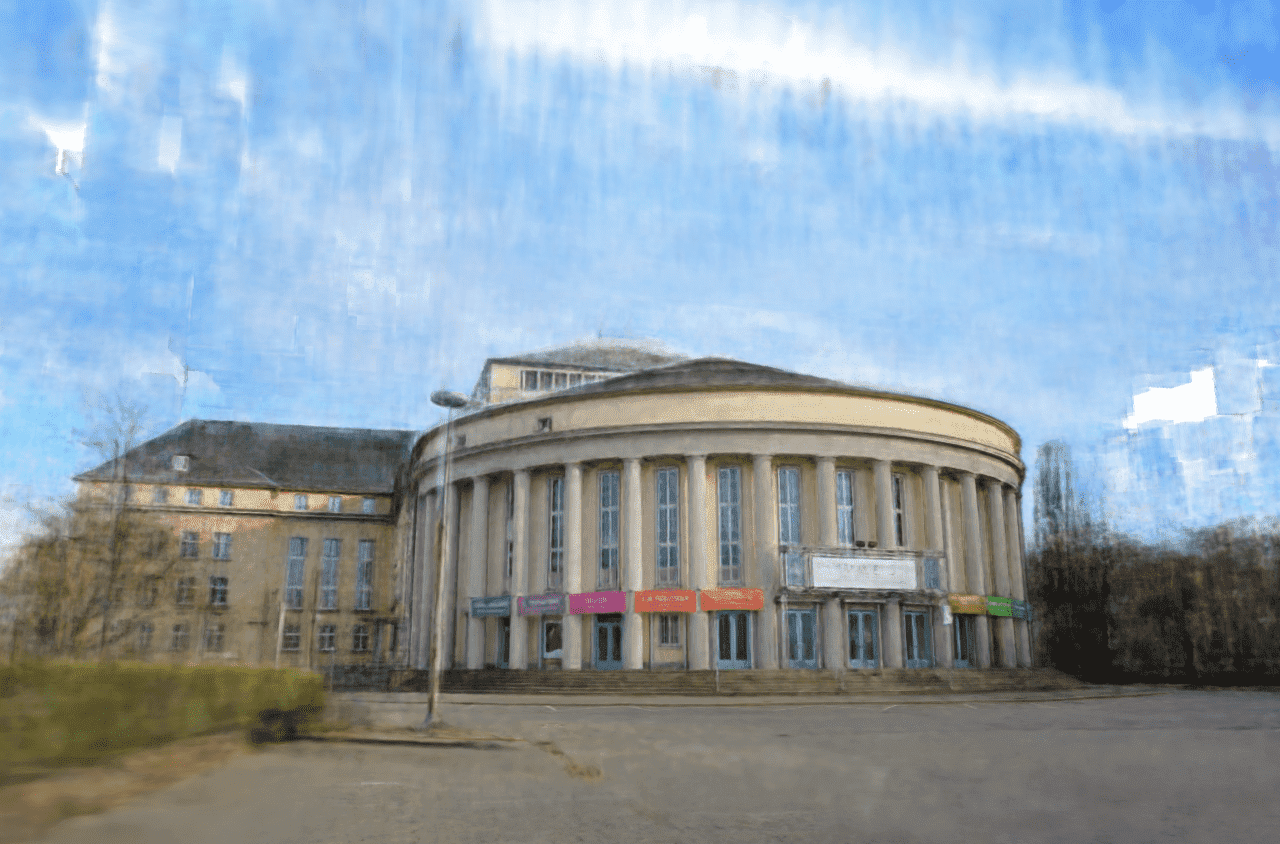}
    \caption{ray contraction}
    \label{fig:cubemap-b}
  \end{subfigure}
  \begin{subfigure}{0.32\linewidth}
    \centering
    \includegraphics[width=1.0\linewidth]{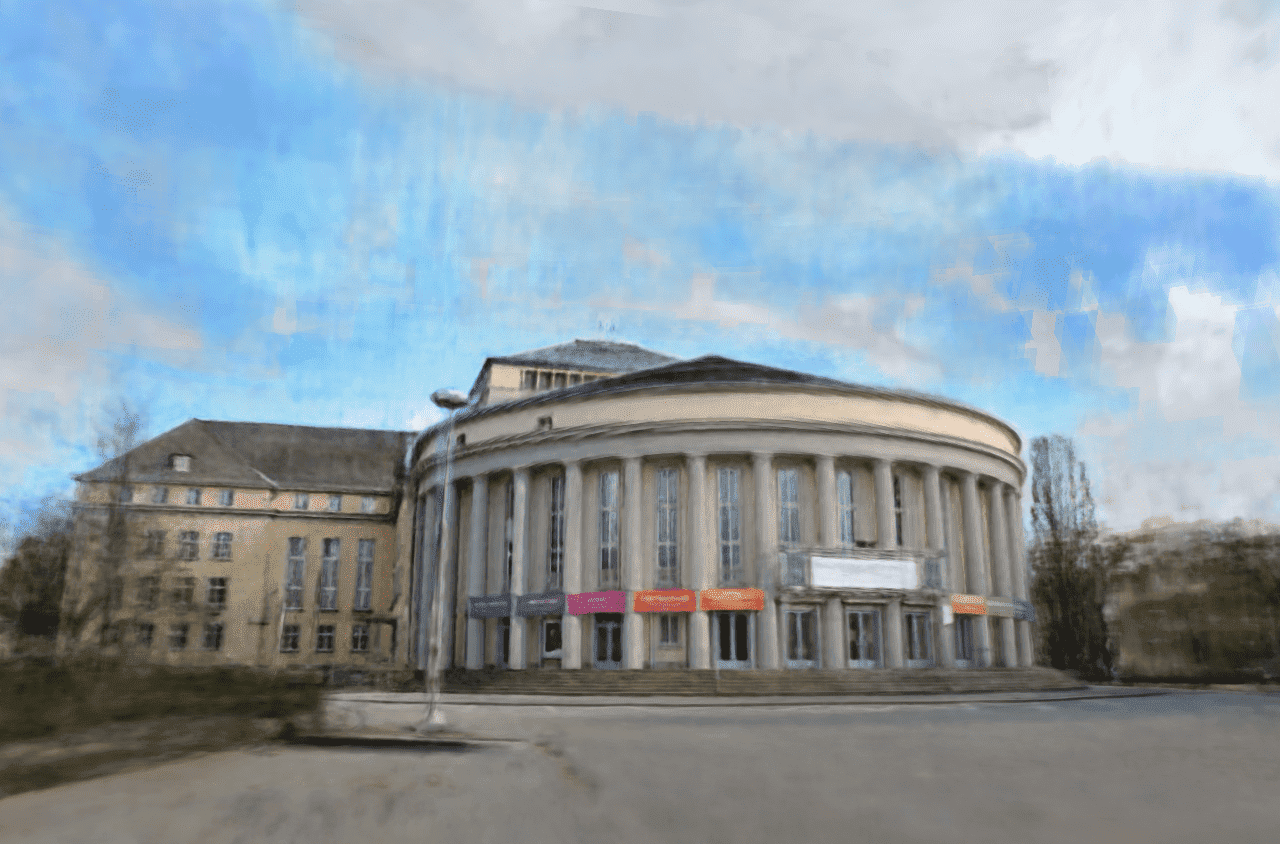}
    \caption{ours}
    \label{fig:cubemap-c}
  \end{subfigure}
  \caption{\textbf{Cubemap Ablation.} SR-TensoRF extends its rendering capabilities to the sky. As we progress from (a) a model without cubemap and ray contraction to (b) a model with only ray contraction, and finally to (c) a model with both, the training efficiency also increases, masking out unnecessary voxels.}
  \label{fig:cubemap}
\end{figure}

\paragraph{Benefit of cubemap.}
As shown in Fig.~\ref{fig:cubemap-a}, without considering unbounded scenes, there are problems: artifacts appear in a place where they should not be (sky, void areas, etc.), or holes are formed.
While ray contraction in Fig.~\ref{fig:cubemap-b} mitigates this to some extent, the alpha mask of TensoRF still leads to hole formation. 
However, upon the incorporation of cubemaps as in Fig.~\ref{fig:cubemap-c}, the holes disappear, resulting in a more natural depiction of the sky.
Moreover, this enables better utilization of the alpha mask, consequently reducing training time.

\begin{figure*}[t!]
  \centering
  \begin{subfigure}{0.9\linewidth}
    \centering
    \resizebox{1.0\linewidth}{!}{
    \begin{tabular}{{>{\centering\arraybackslash}m{0.14\linewidth}m{0.22\linewidth}m{0.22\linewidth}m{0.22\linewidth}m{0.22\linewidth}}}
    NeRF-OSR~\cite{rudnev2022nerf} &
    \includegraphics[width=1.0\linewidth]{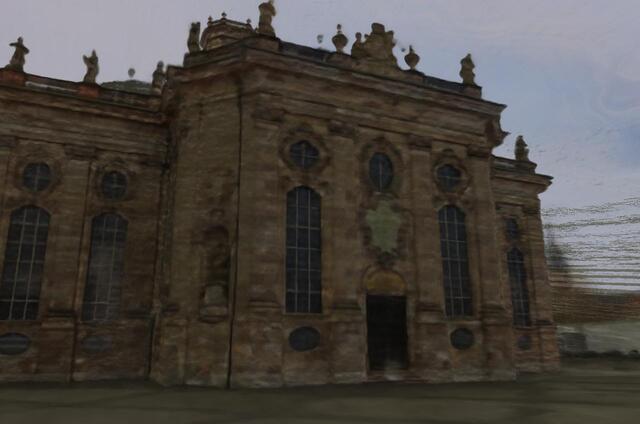} & 
    \includegraphics[width=1.0\linewidth]{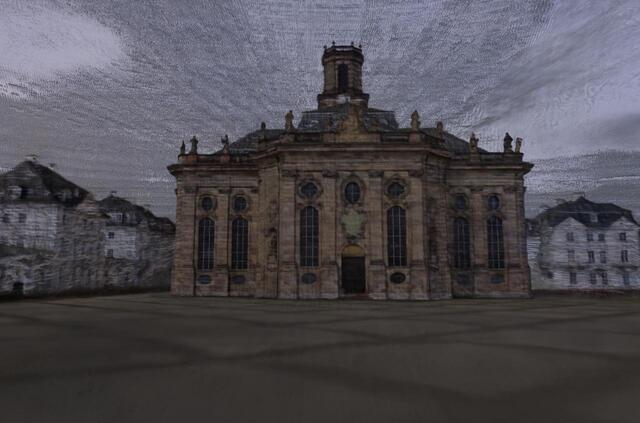} & 
    \includegraphics[width=1.0\linewidth]{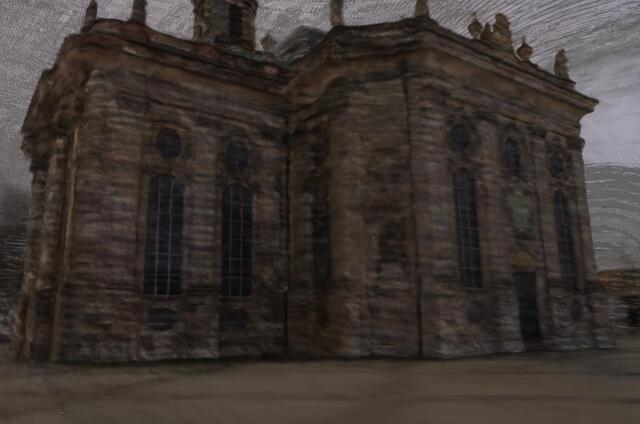} & 
    \includegraphics[width=1.0\linewidth]{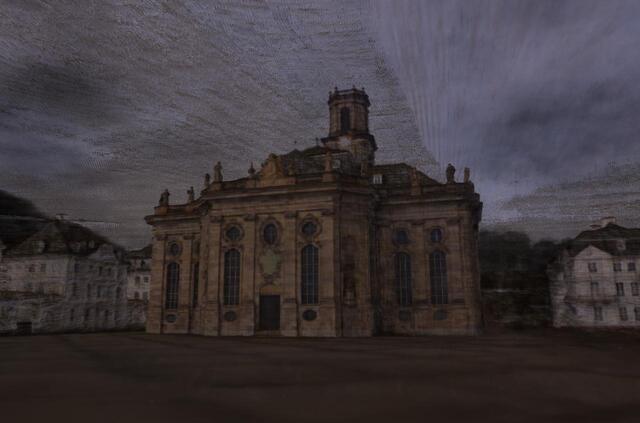} \\
    Ours &
    \includegraphics[width=1.0\linewidth]{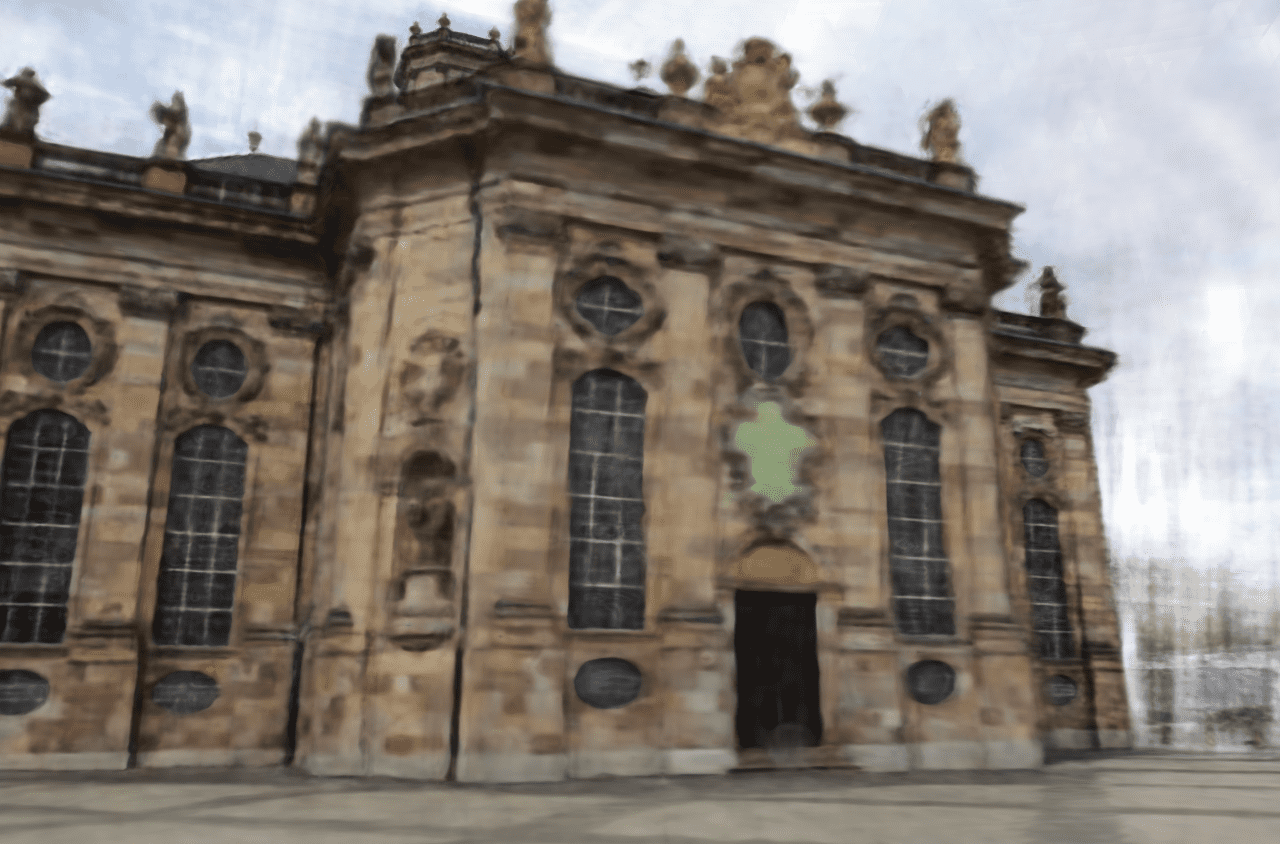} & 
    \includegraphics[width=1.0\linewidth]{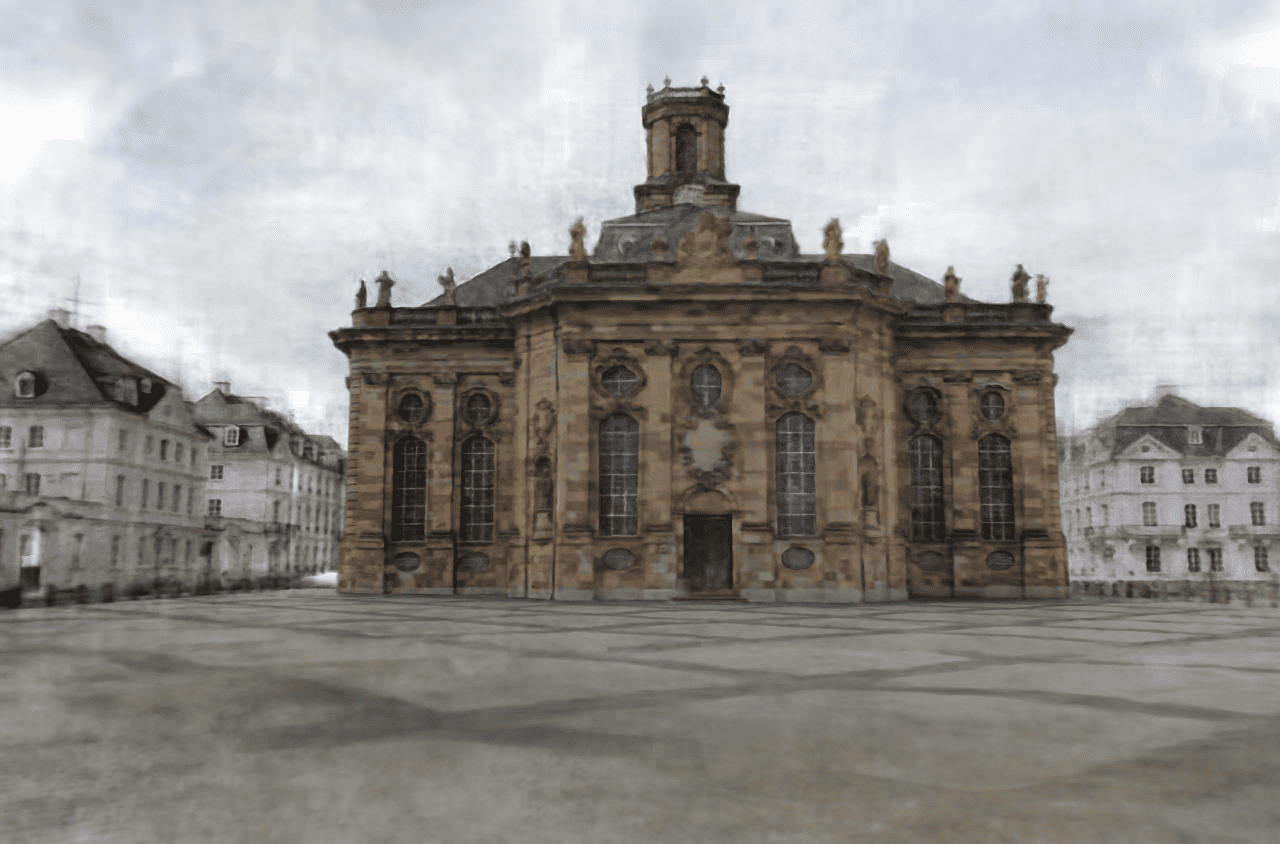} & 
    \includegraphics[width=1.0\linewidth]{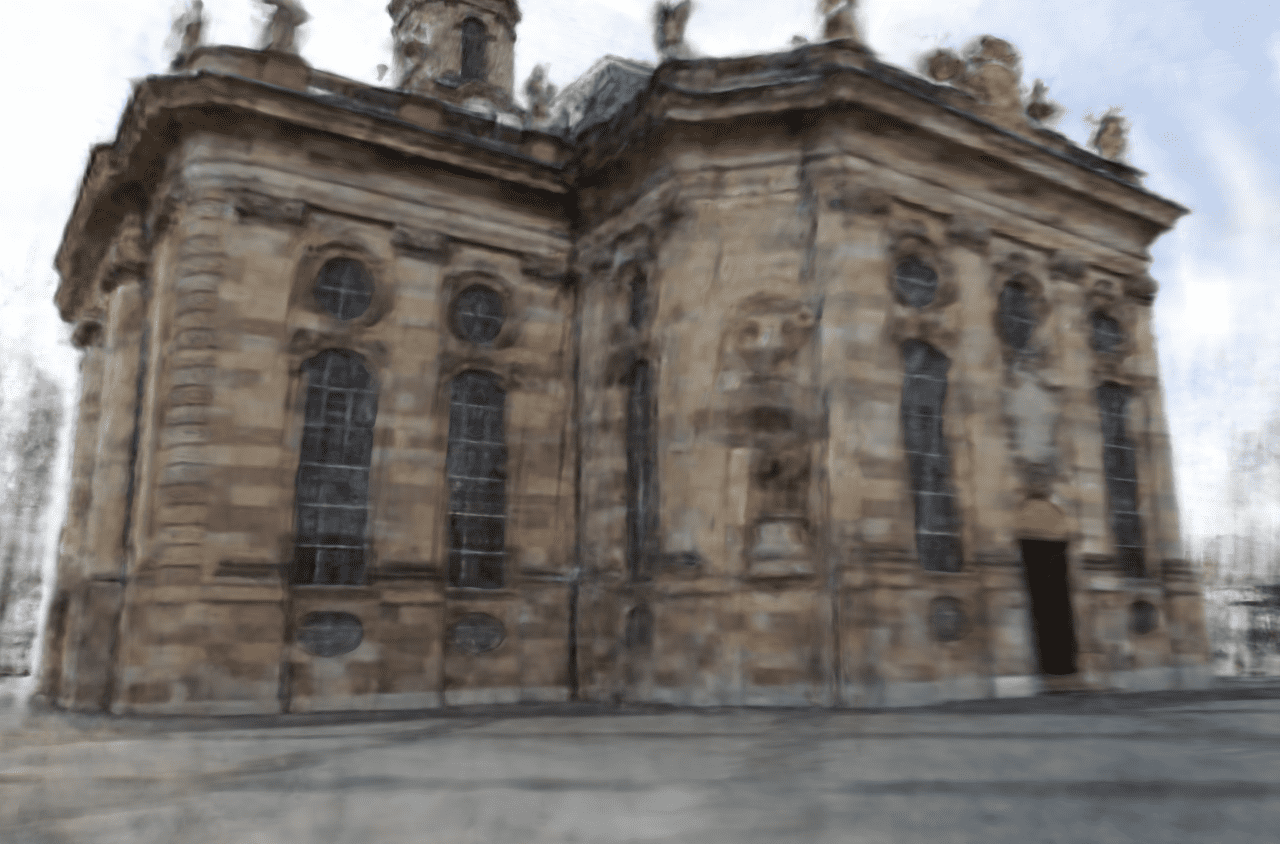} & 
    \includegraphics[width=1.0\linewidth]{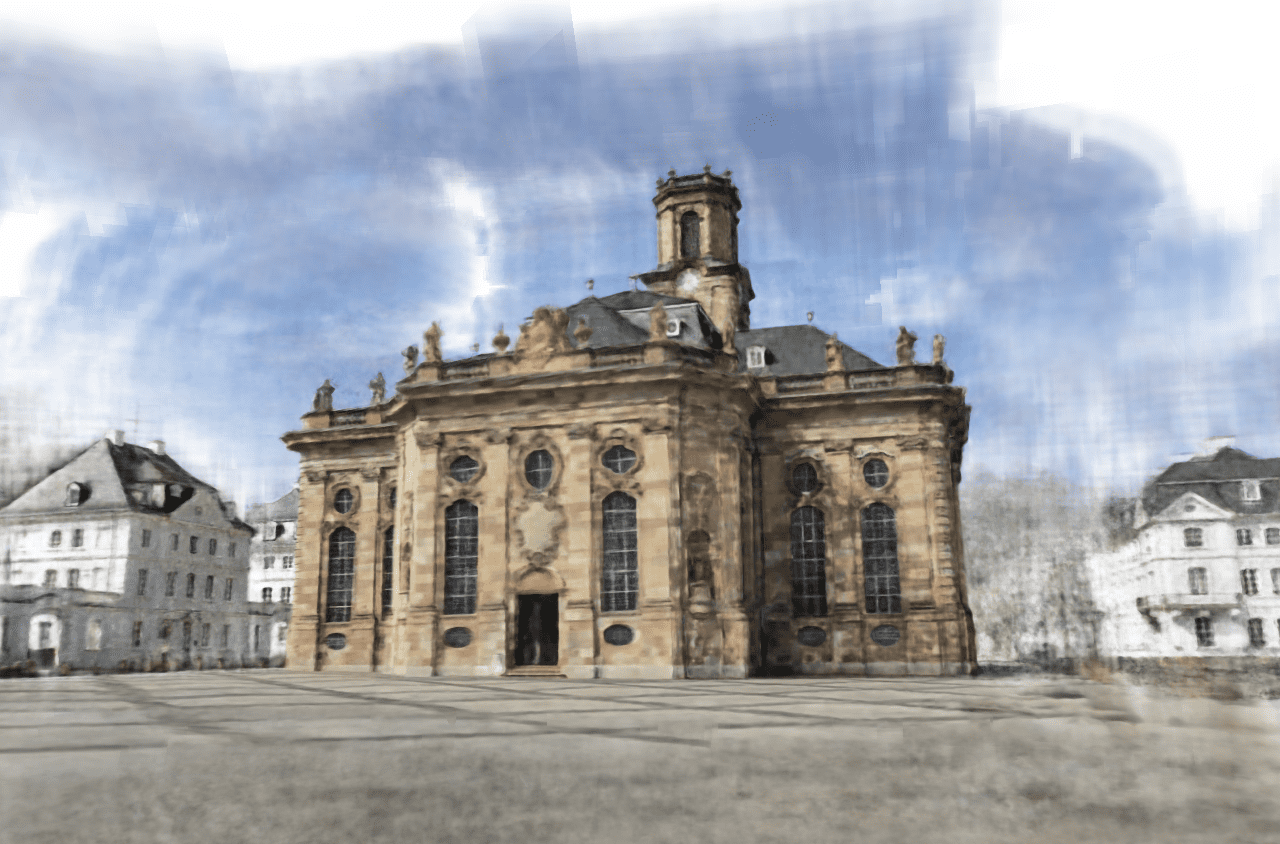} \\
    Shadow &
    \includegraphics[width=1.0\linewidth]{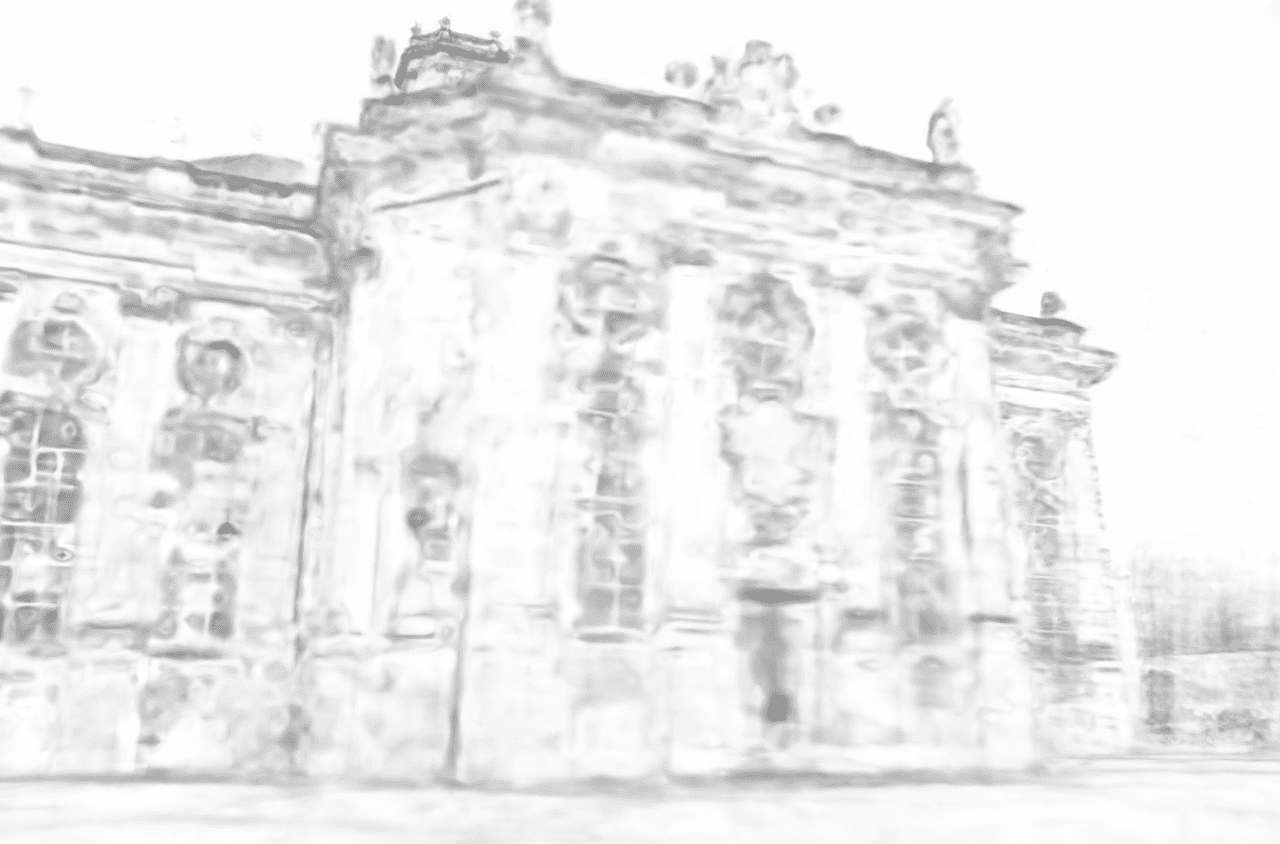} & 
    \includegraphics[width=1.0\linewidth]{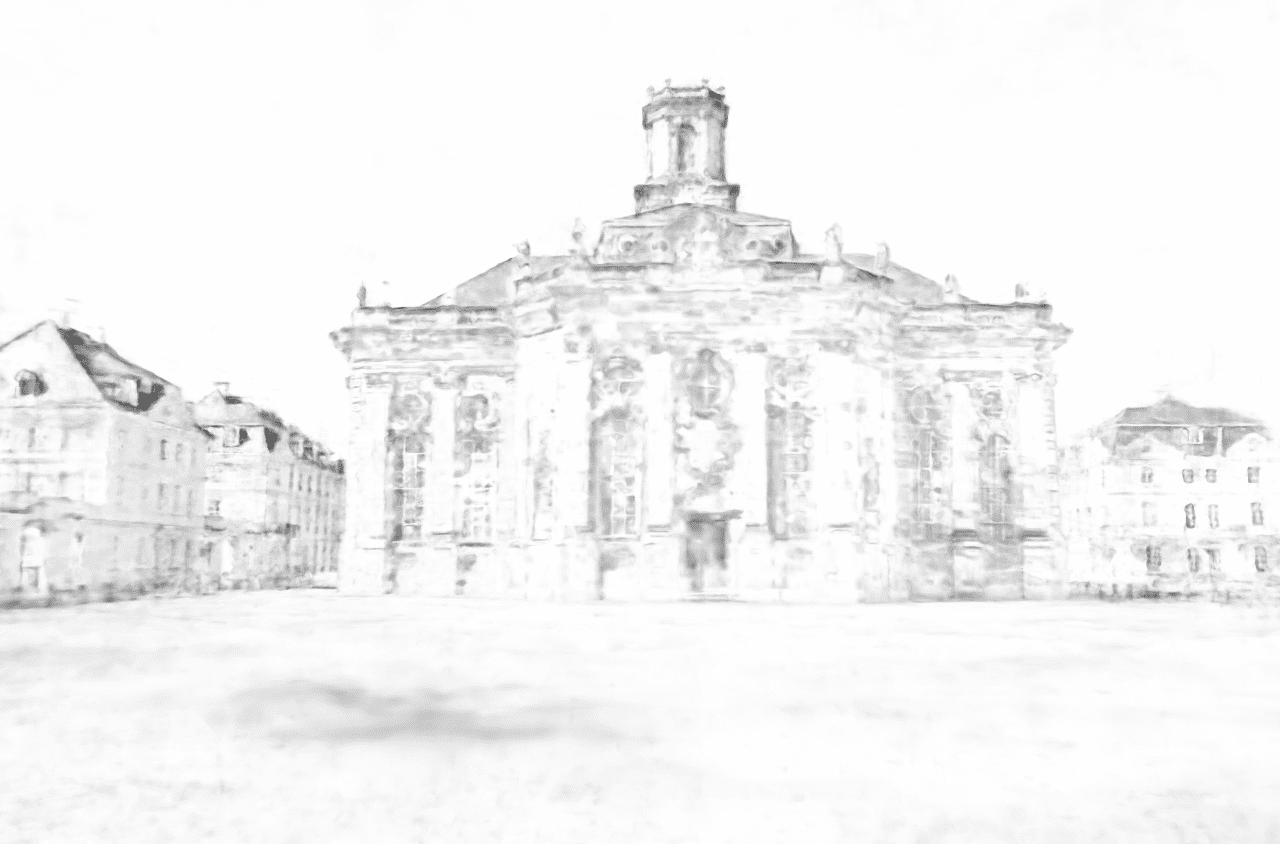} & 
    \includegraphics[width=1.0\linewidth]{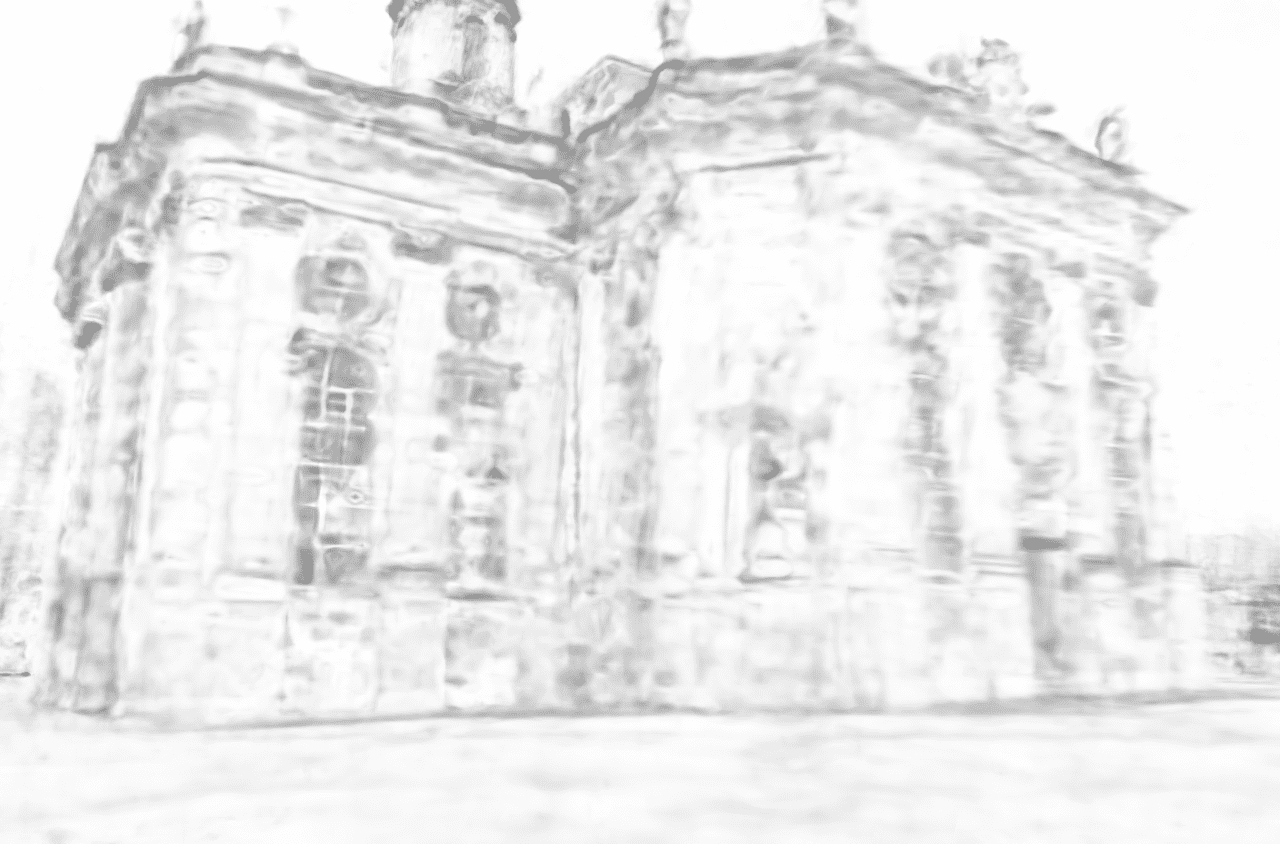} & 
    \includegraphics[width=1.0\linewidth]{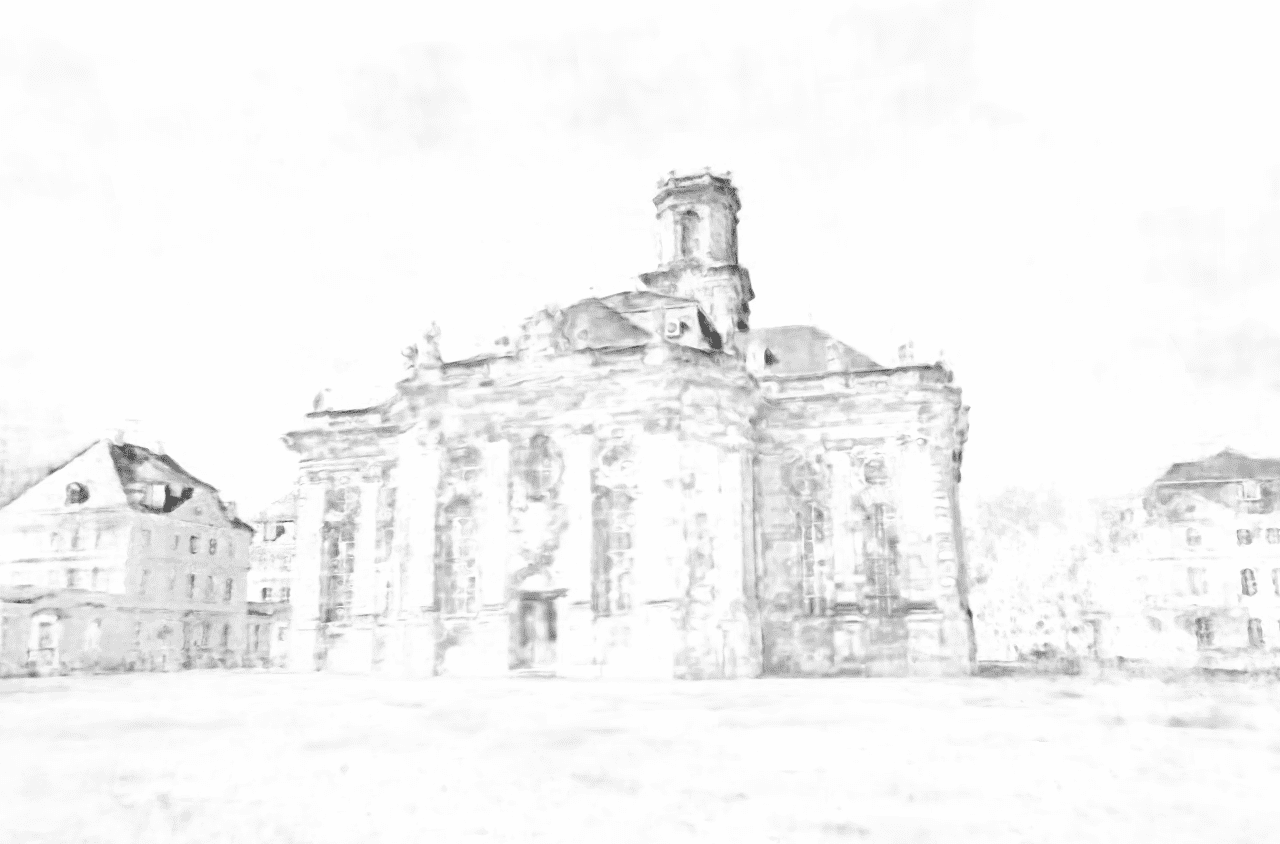} 
    \end{tabular}
  }
    \label{fig:relight-c}
  \end{subfigure}
  \vspace{-2mm}
  \caption{\textbf{Relight Rendering.} SR-TensoRF is capable of producing plausible relit images without using environment maps. Even under new lighting conditions, our approach maintains the reconstruction quality without degradation.}
  \label{fig:relight}
\end{figure*}

\begin{table*}[t]
  \centering
  \resizebox{0.9\linewidth}{!}{
    \begin{tabular}{{l|cccc|cccc|cccc}}
    Method & PSNR $\uparrow$ & SSIM $\uparrow$ & MSE $\downarrow$ & MAE $\downarrow$ & PSNR $\uparrow$ & SSIM $\uparrow$ & MSE $\downarrow$ & MAE $\downarrow$ & PSNR $\uparrow$ & SSIM $\uparrow$ & MSE $\downarrow$ & MAE $\downarrow$\\
    \midrule
     & \multicolumn{4}{c|}{Site 1} & \multicolumn{4}{c|}{Site 2} & \multicolumn{4}{c}{Site 3} \\
    \midrule
    TensoRF~\cite{chen2022tensorf} & 15.32 & 0.511 & 0.034 & 0.123 & 13.43 & 0.583 & 0.050 & 0.145  & 14.35 & 0.626 & 0.039 & 0.122 \\
    \midrule
    \midrule
    Ours & \textbf{17.30} & \underline{0.542} & \textbf{0.021} & \textbf{0.096} & \textbf{15.63} & \textbf{0.632} & \textbf{0.030} & \textbf{0.111} & \textbf{16.74} & \textbf{0.653} & \textbf{0.024} & \textbf{0.093} \\
    (-) shadow & \underline{17.21} & \textbf{0.543} & \textbf{0.021} & \underline{0.097} & \underline{15.46} & \underline{0.631} & \underline{0.031} & \underline{0.114}  & 16.48 & 0.651 & 0.026 & 0.097 \\	
    (-) cubemap & 16.66 & 0.541 & 0.027 & 0.105 & \underline{15.46} & 0.623 & 0.032 & \underline{0.114}  & \underline{16.66} & \textbf{0.653} & \textbf{0.024} & \textbf{0.093}\\	
    (-) latent vector & 15.34 & 0.525 & 0.035 & 0.126 & 14.92 & 0.623 & 0.041 & 0.126 & 15.83 & 0.643 & 0.030 & 0.106 \\
  \end{tabular}
  }
  \vspace{-2mm}
  \caption{\textbf{Ablation.} SR-TensoRF demonstrates a performance drop when individually removing each component.}
  \vspace{-1mm}
  \label{tab:ablation}
\end{table*}

\subsection{Comparion with Other Methods}
Our quantitative results, as shown in Tab.~\ref{tab:quantitative}, demonstrate that our model generally outperforms previous studies on Site 2 and 3 across various metrics. 
However, our model's performance is relatively lower on Site 1, which can be attributed to the prevalence of cloudy training images in this particular scene, hindering proper learning of the sun effect. 
The results of NeRF-OSR~\cite{rudnev2022nerf} are obtained by applying a default environment map to all scenes during evaluation. 
Since NeRF-OSR dataset includes spherical harmonics extracted from environment map images of the test set only for Site 1, we applied this environment map lighting to Site 1 as well.
Despite the higher accuracy of the test set environment map compared to the default one for test set images, the results show a decline, highlighting NeRF-OSR's reliance on lighting for scene reconstruction.
Our model, on the other hand, achieves competitive results with NeRF-OSR, even without depending on test set environment maps, and effectively applies lighting to the test set.

In Fig.~\ref{fig:qualitative}, we show our qualitative results. 
The rendered images of NeRF-OSR~\cite{rudnev2022nerf} display uniform colors across all scenes and views in the test set due to the use of consistent lighting. 
The outcomes obtained by individually applying environment maps corresponding to each ground truth can be seen in Fig.~\ref{fig:relight} (only for Site 1).
Fig.~\ref{fig:qualitative-b} demonstrates that building shadows, which are not well-reconstructed in NeRF-OSR, are successfully generated in our model.
Furthermore, in Fig.~\ref{fig:qualitative-c}, it is evident that the buildings are rendered with varying brightness based on the direction of sunlight, whether it is facing the front or rear of the structures.

\begin{figure}[t!]
  \centering
  \begin{subfigure}{0.3\linewidth}
    \centering
    \includegraphics[width=1.0\linewidth]{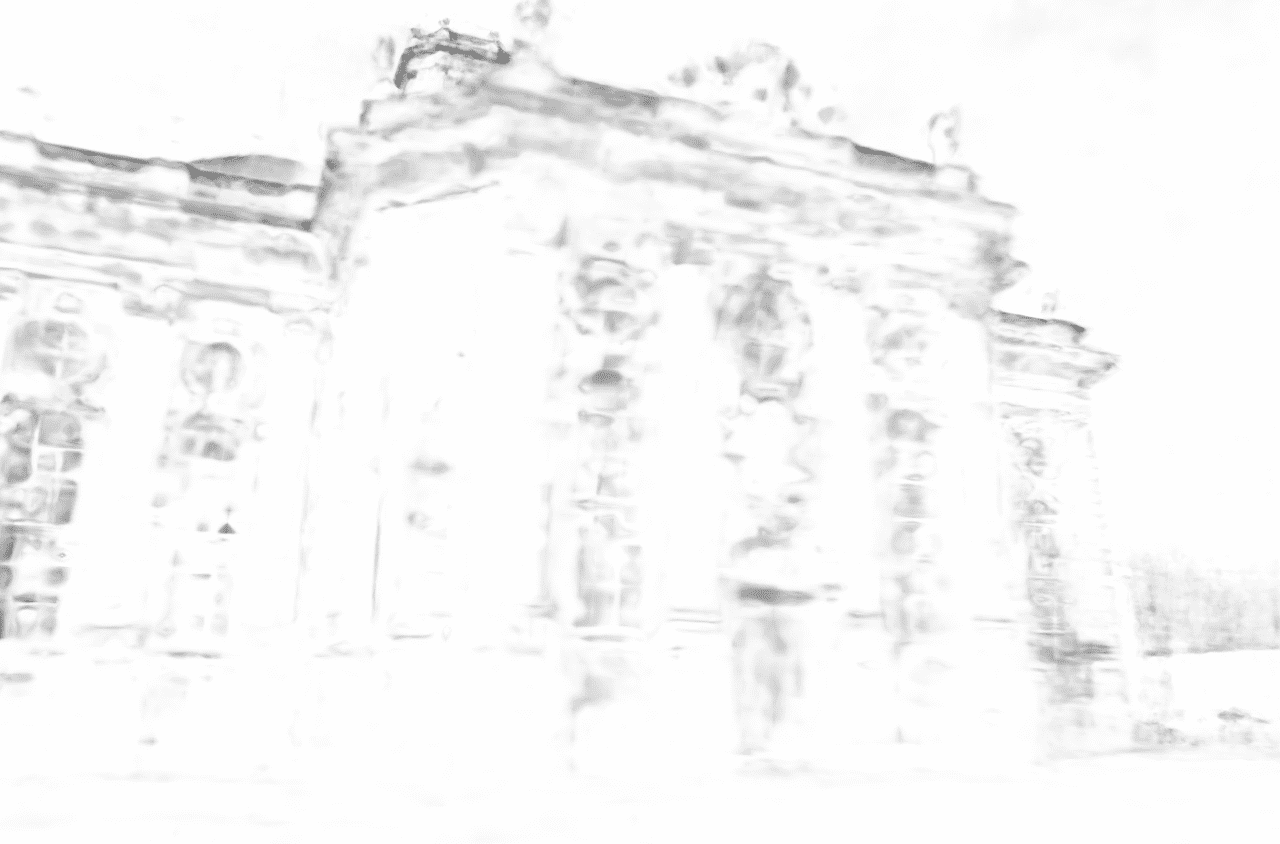}
    \caption{AM 11:40}
    \label{fig:time-a}
  \end{subfigure}
  \begin{subfigure}{0.3\linewidth}
    \centering
    \includegraphics[width=1.0\linewidth]{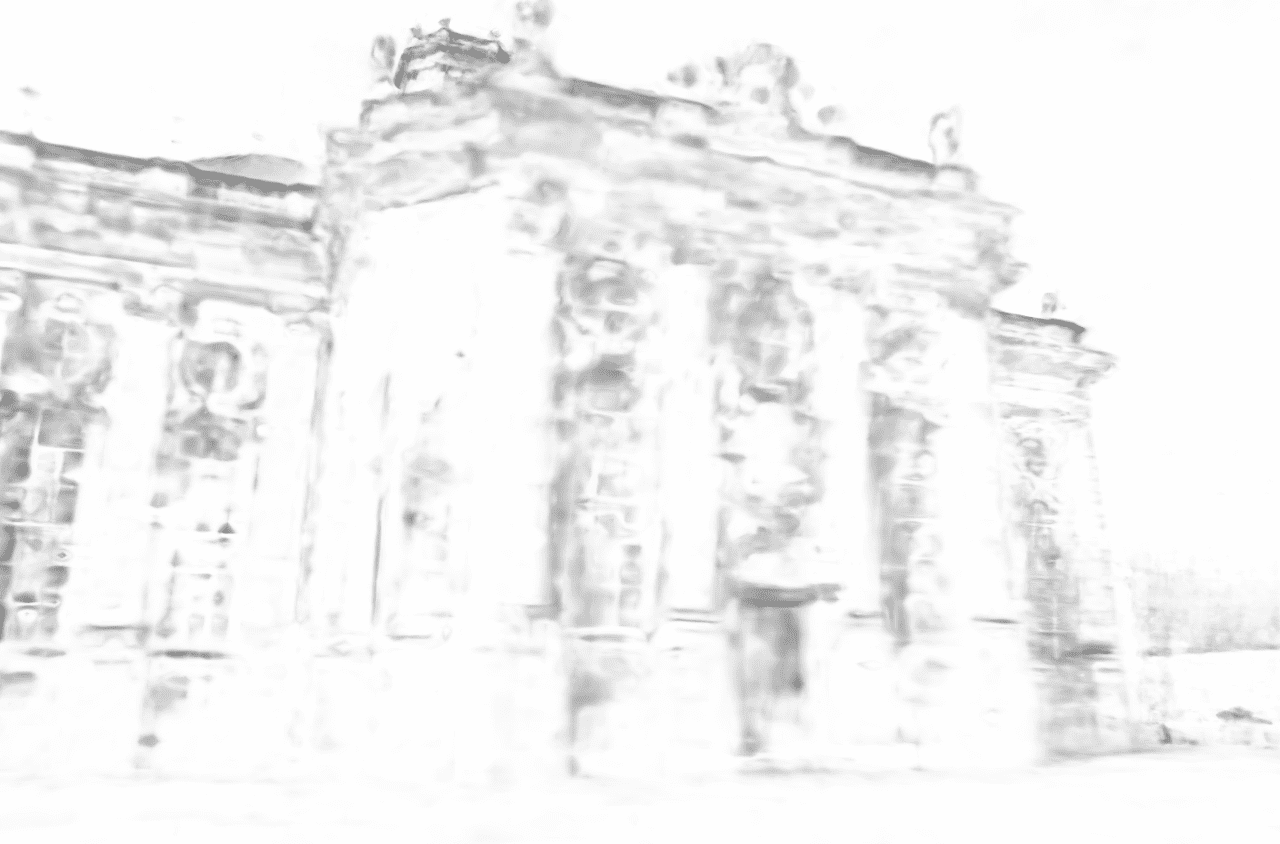}
    \caption{PM 12:40}
    \label{fig:time-b}
  \end{subfigure}
  \begin{subfigure}{0.3\linewidth}
    \centering
    \includegraphics[width=1.0\linewidth]{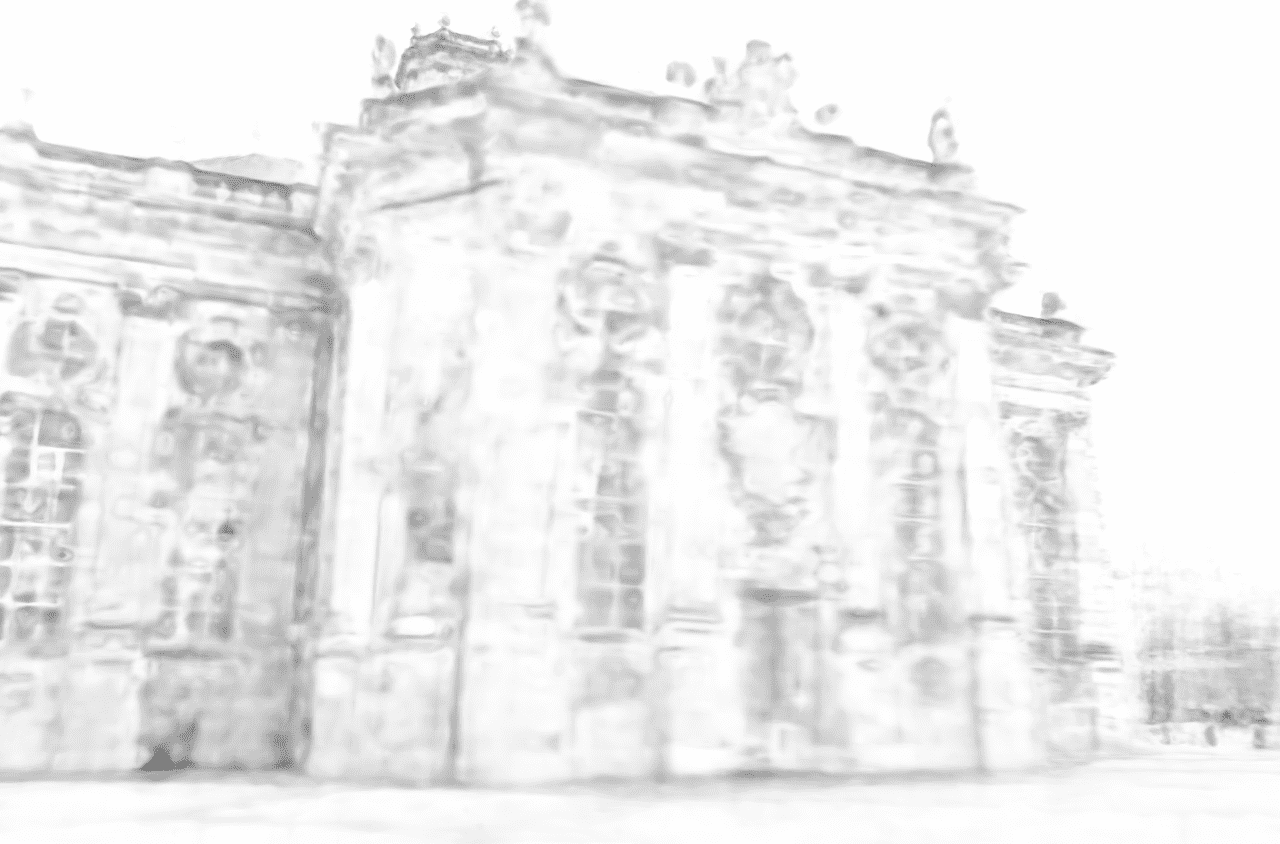}
    \caption{PM 14:40}
    \label{fig:time-c}
  \end{subfigure}
  \vspace{-2mm}
  \caption{\textbf{Relight by time.} Results when keeping other factors constant and only changing the sun direction.
  SR-TensoRF can generate varying forms of shadows that change over time.}
  \vspace{-3mm}
  \label{fig:time}
\end{figure}

\subsection{Relighting Results}
Fig.~\ref{fig:relight} shows the qualitative results of relighting. 
Every first row showcases the results of applying the test set environment map on NeRF-OSR~\cite{rudnev2022nerf} (NeRF-OSR$^\ddagger$ of Tab.~\ref{tab:quantitative}). 
It is evident that the quality is notably compromised compared to the results where the same default environment map was applied to all scenes, as observed in Fig.~\ref{fig:qualitative-a}. 
The second row presents the results of our SR-TensoRF, with different sun directions and latent vectors applied to the same viewing direction as in the first row. 
In this case, the latent vectors were randomly selected from the left-half images of the test set.
Noticeable variations in lighting, distinct from the reconstructions in Fig.~\ref{fig:qualitative-a}, can be observed. 
The third row depicts the shadows utilized in the rendering process of the second row.

Fig.~\ref{fig:time} shows the shadows rendered for relighting. 
Given the same viewing direction, it is evident that SR-TensoRF successfully generates shadows that vary over time. 
For instance, in the case of the buildings in Site 1, which face east, the shadows are appropriately generated as the sun transitions from the southern direction in Fig.~\ref{fig:time-a}, moving towards the southwest in Fig.~\ref{fig:time-b}, and further tilting westward in Fig.~\ref{fig:time-c}.
Such results are achievable by directly utilizing the sun direction.
Further relighting results for various scenes can be found in the supplementary material.

\subsection{Ablation Study}

To validate the effectiveness of our SR-TensoRF, we conduct an ablation study as in Tab.~\ref{tab:ablation}. 
When comparing the results of experiments where shadow, cubemap, and latent vector are individually omitted, our full SR-TensoRF, which integrates all the components, exhibits the highest performance.
The most substantial decrease in performance occurs when the latent vector is removed, primarily due to the significant impact of tone on quantitative metrics.
This is particularly evident for Site 1, where many training images exhibit cloudy weather, accordingly the effects of sunlight are minimal and only the tones tend to vary. 
Latent vectors, however, alone cannot enable our time-varying relighting capabilities. 
In addition, small holes in the sky may not have a significant impact on quantitative metrics, yet they can be visually prominent.
Addressing this issue with a cubemap solution results in more visually appealing results.

\section{Conclusion}
\label{sec:conclusion}
We propose a sun-aligned outdoor scene relighting model SR-TensoRF, which consists of appearance, density, and cubemap tensors. 
Our model utilizes the sun direction for relighting, bypassing the need for environment maps. 
Additionally, it employs latent vectors to disentangle various tones of images.
Our proposed cubemap tensor efficiently combines the fast training speed of TensoRF while effectively handling unbounded scenes.
With our SR-TensoRF, we demonstrate that the sun direction can be a viable alternative to environment maps, particularly for outdoor scene relighting. 
SR-TensoRF not only simplifies the relighting pipeline but also enables the generation of dynamic shadows over time.

\vspace{-3mm}
\paragraph{Future work.}
Our research can be extended in various directions. 
As we tackle unbounded scenes, there is potential to model shadows cast by surrounding objects such as trees that might not have been captured in the training images.
Furthermore, relighting methods that account for cloudy weather conditions could be explored. 
Additionally, our approach, initially designed for daytime scenes, could be extended to nighttime scenarios by adapting the lighting direction, similar to sun direction utilization.

\vspace{-3mm}
\paragraph{Acknowledgements.}
This work was supported by NRF grant (2021R1A2C3006659) and IITP grant (2022-0-00320) funded by Korean Government. 

\clearpage

{\small
\bibliographystyle{ieee_fullname}
\bibliography{egbib}
}

\end{document}


\title{Fast Sun-aligned Outdoor Scene Relighting based on TensoRF \\ ---Supplementary Material---} 

\maketitle

\setcounter{section}{0}
\setcounter{table}{0}
\setcounter{figure}{0}
\renewcommand\thesection{\Alph{section}}
\renewcommand\thetable{\Alph{table}}
\renewcommand\thefigure{\Alph{figure}}

\thispagestyle{empty}

\section{Implementation Details}

\begin{table}[!t]
\centering
\resizebox{1.0\linewidth}{!}{
\begin{tabular}{l|c|c|c}
\toprule
  & Site 1 & Site 2 & Site 3\\
  \midrule
  Learning Rate (for Tensors) & \multicolumn{3}{c}{$[0.01, 0.002]$} \\
  \midrule
  Learning Rate (for MLPs) & \multicolumn{3}{c}{$[0.001, 0.0002]$} \\
  \midrule
  Iteration & 54860 & 158777 & 136095 \\
  \midrule
  $\lambda_{1} (\text{for } \mathcal{L}_1)$ & \multicolumn{3}{c}{$[0.01, 0.001]$} \\
  \midrule
  $\lambda_{2} (\text{for }\mathcal{L}_\text{TV of }\mathcal{G}_{\sigma})$ & \multicolumn{3}{c}{$0.1$} \\
  \midrule
  $\lambda_{3} (\text{for }\mathcal{L}_\text{TV of }\mathcal{G}_a)$ & \multicolumn{3}{c}{$0.01$} \\
  \midrule
  $\lambda_{3} (\text{for }\mathcal{L}_\text{TV of }\mathcal{G}_{cubemap})$ & \multicolumn{3}{c}{$0.01$} \\
  \midrule
  $\lambda_{2} (\text{for }\mathcal{L}_\text{shadow} \text{ regularization})$ & \multicolumn{3}{c}{$0.001$} \\
\bottomrule
\end{tabular}}
\captionof{table}{
    \textbf{Details of hyperparameters and loss weight terms.}
    $[a, b]$ indicates the annealing from $a$ to $b$.
    }
    \label{supp_tab:training_detail}
\end{table}

\begin{figure}[t!]
  \centering
  \begin{subfigure}{0.3\linewidth}
    \centering
    \includegraphics[width=1.0\linewidth]{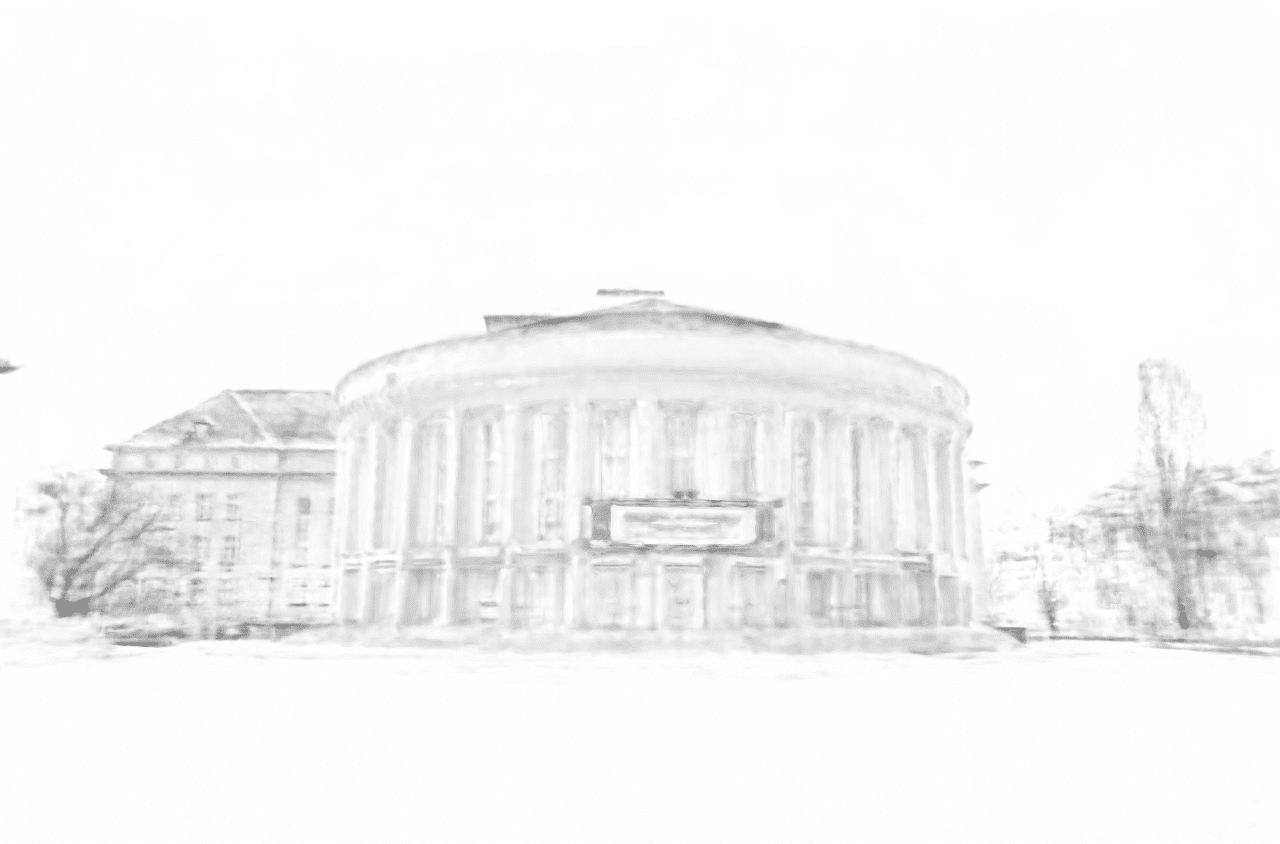}
    \caption{AM 08:00}
    \label{supp_fig:time-a}
  \end{subfigure}
  \begin{subfigure}{0.3\linewidth}
    \centering
    \includegraphics[width=1.0\linewidth]{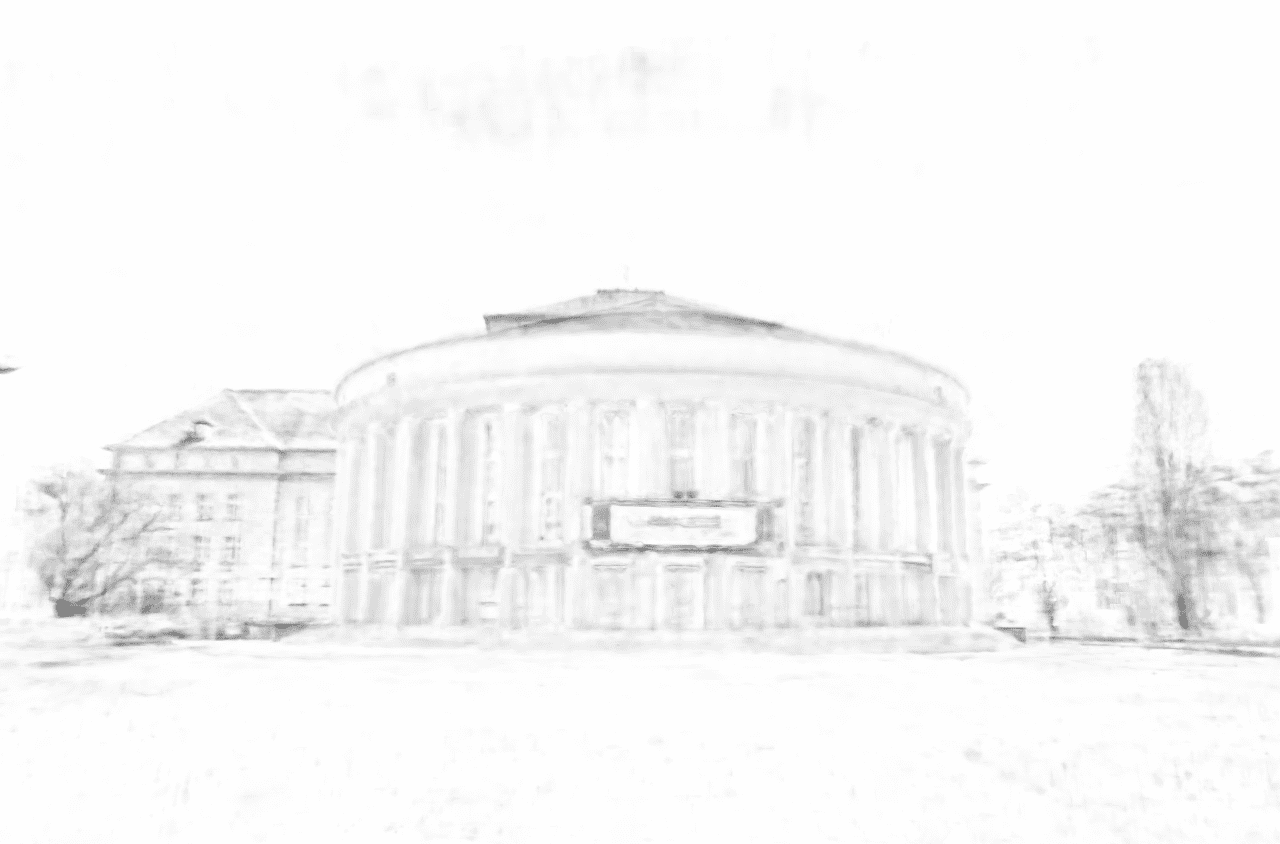}
    \caption{PM 14:00}
    \label{supp_fig:time-b}
  \end{subfigure}
  \begin{subfigure}{0.3\linewidth}
    \centering
    \includegraphics[width=1.0\linewidth]{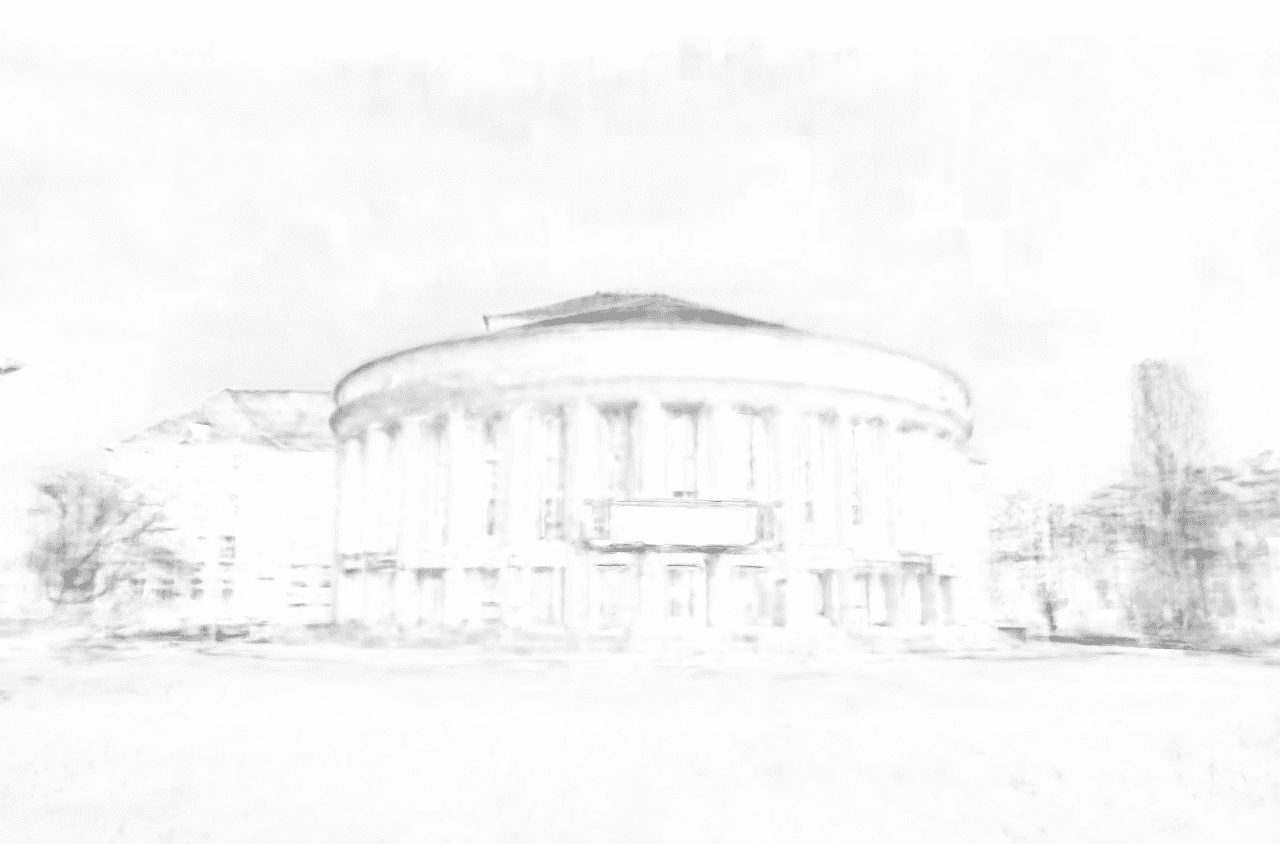}
    \caption{PM 19:00}
    \label{supp_fig:time-c}
  \end{subfigure}
  \caption{\textbf{Relight by time.} Results when keeping other factors constant and only changing the sun direction for Site 2. The building is facing northwest.}
  \label{supp_fig:time2}
\end{figure}

\begin{figure}[t!]
  \centering
  \begin{subfigure}{0.3\linewidth}
    \centering
    \includegraphics[width=1.0\linewidth]{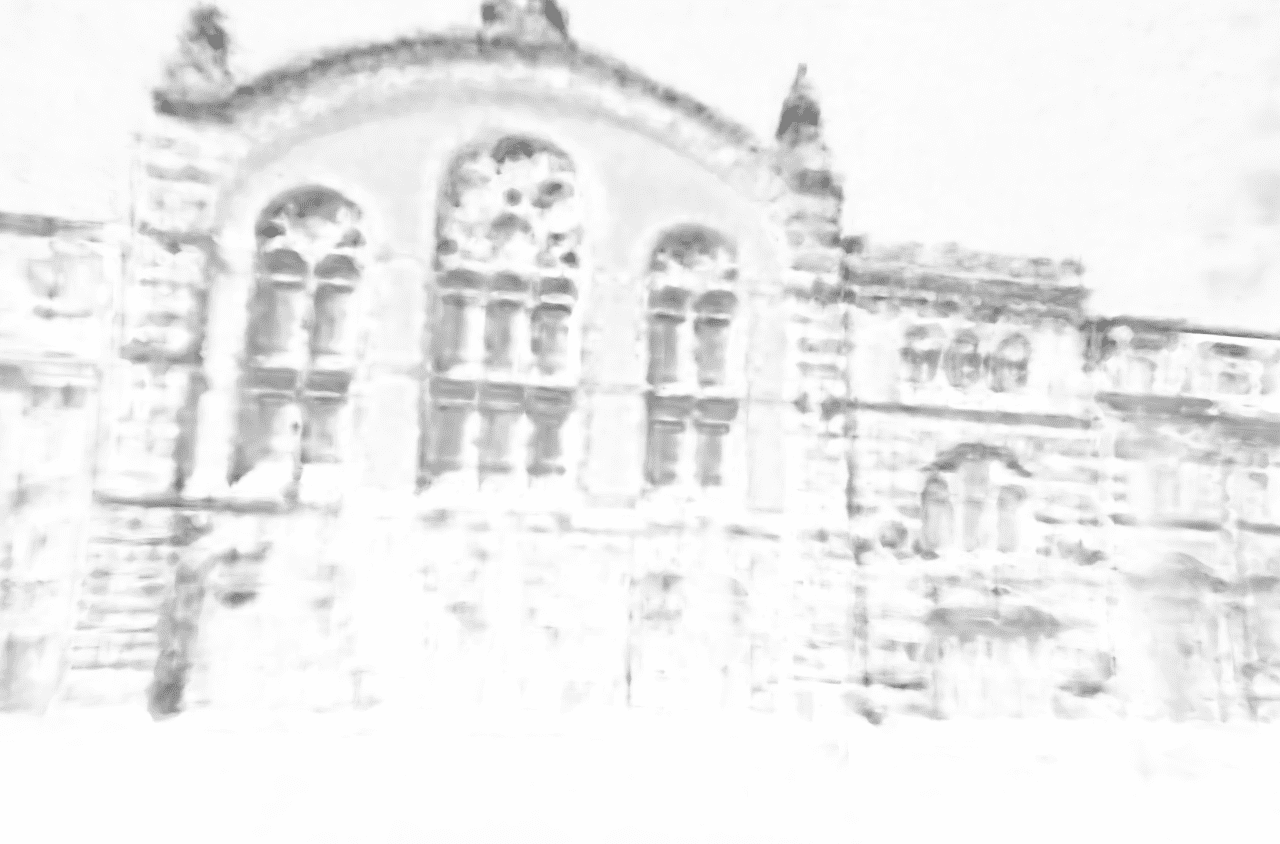}
    \caption{AM 11:00}
    \label{supp_fig:time-a}
  \end{subfigure}
  \begin{subfigure}{0.3\linewidth}
    \centering
    \includegraphics[width=1.0\linewidth]{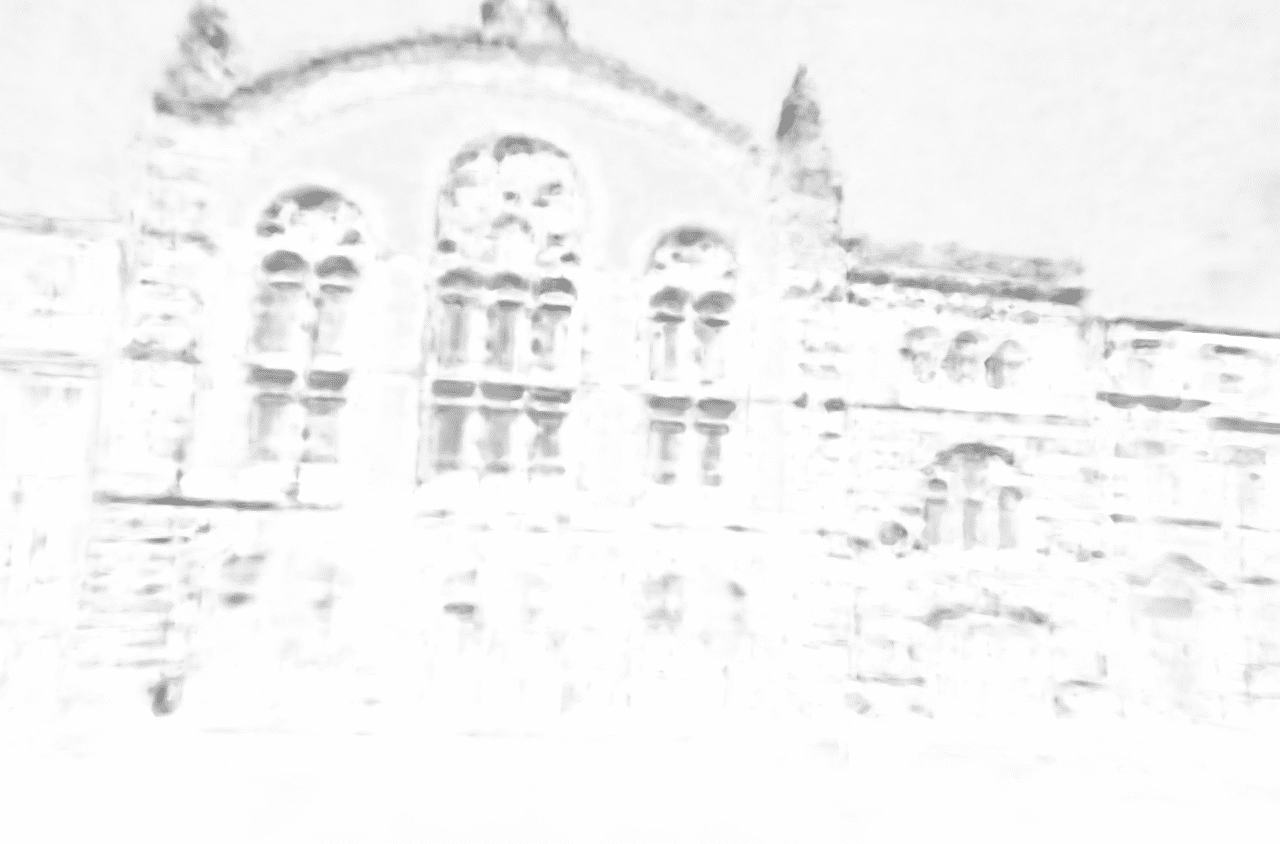}
    \caption{PM 12:00}
    \label{supp_fig:time-b}
  \end{subfigure}
  \begin{subfigure}{0.3\linewidth}
    \centering
    \includegraphics[width=1.0\linewidth]{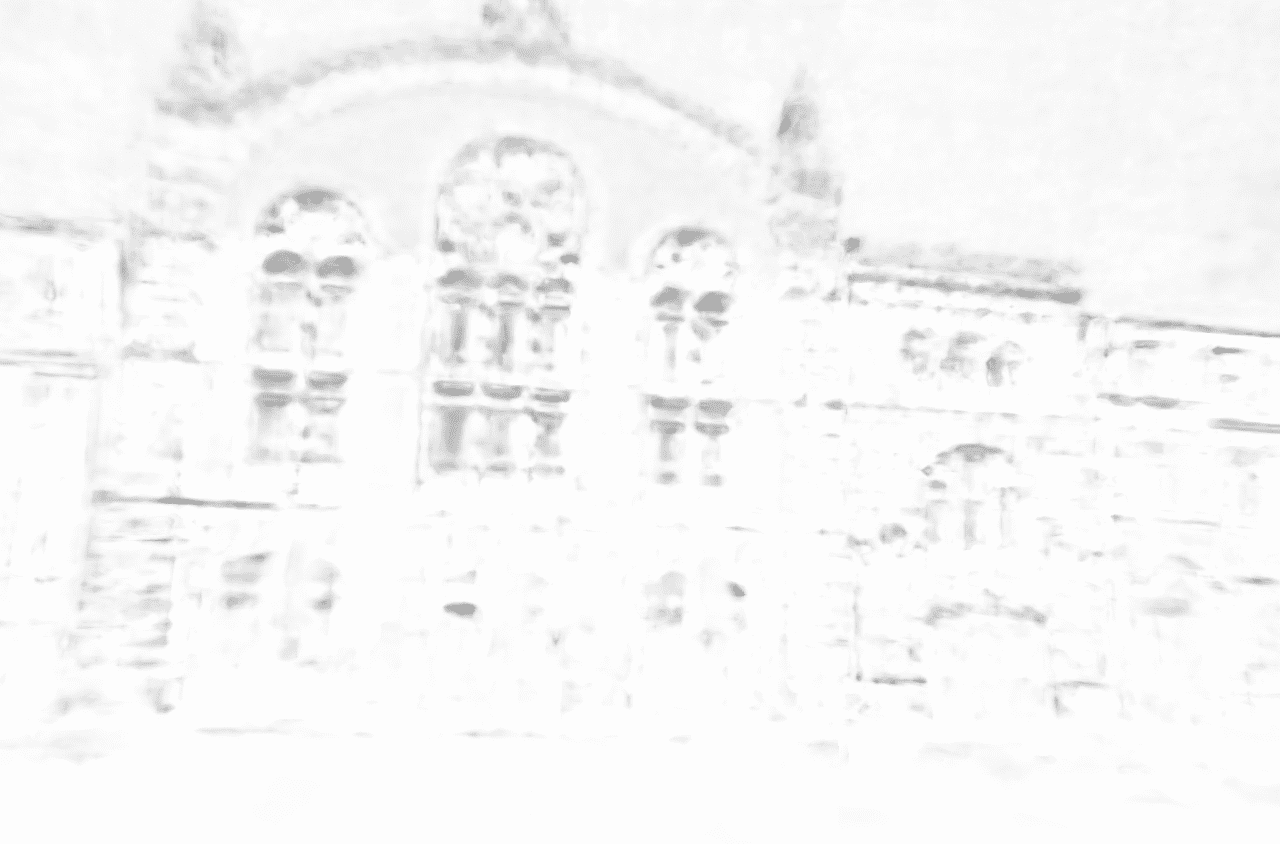}
    \caption{PM 13:00}
    \label{supp_fig:time-c}
  \end{subfigure}
  \caption{\textbf{Relight by time.} Results when keeping other factors constant and only changing the sun direction for Site 3. The building is facing southeast.}
  \label{supp_fig:time3}
\end{figure}

Tab.~\ref{supp_tab:training_detail} presents the details of hyperparameters and our weights of the loss terms, which are tailored to the NeRF-OSR dataset~\cite{rudnev2022nerf}.

\section{Relighting Results}

\begin{figure*}[t!]
  \centering
    \begin{subfigure}{0.9\linewidth}
    \centering
    \resizebox{1.0\linewidth}{!}{
    \begin{tabular}{{>{\centering\arraybackslash}m{0.14\linewidth}m{0.22\linewidth}m{0.22\linewidth}m{0.22\linewidth}m{0.22\linewidth}}}
    Ours &
    \includegraphics[width=1.0\linewidth]{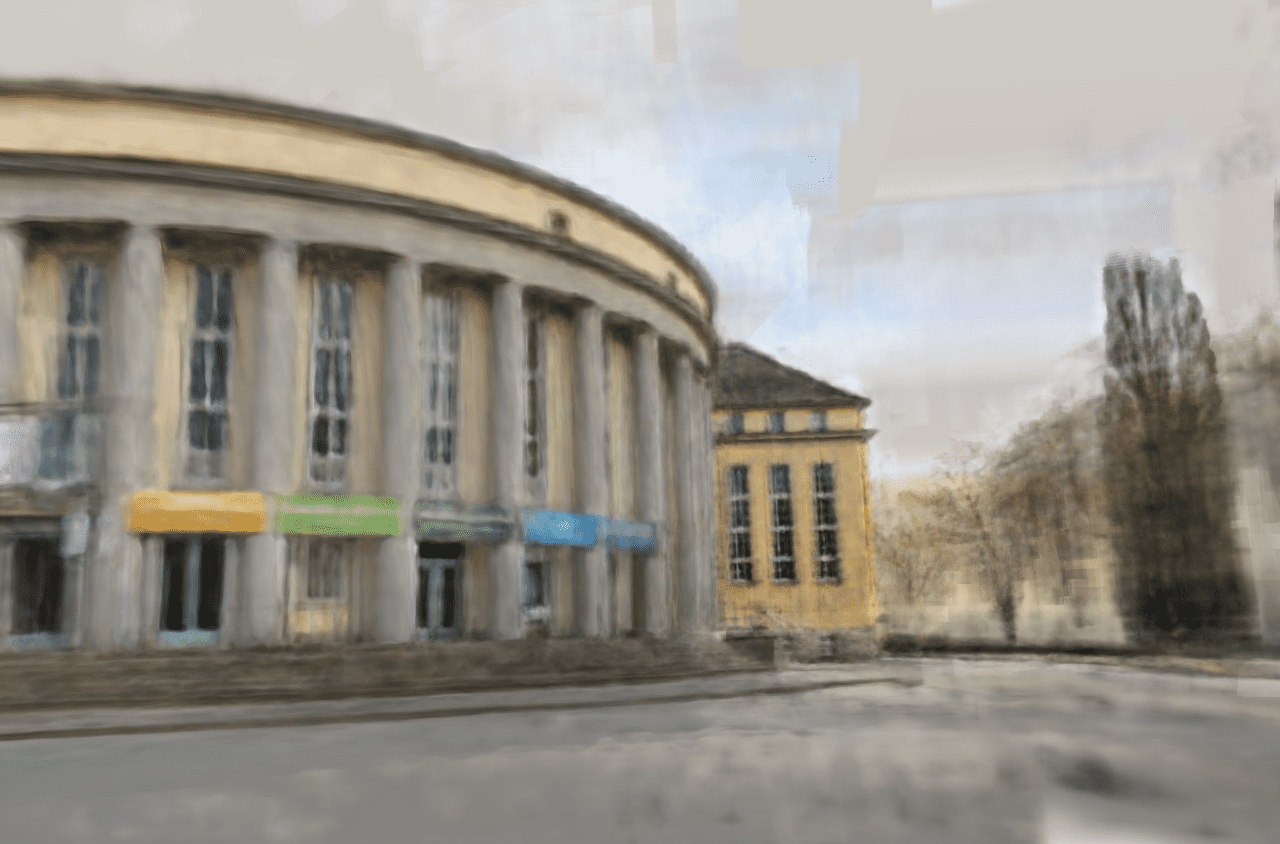} & 
    \includegraphics[width=1.0\linewidth]{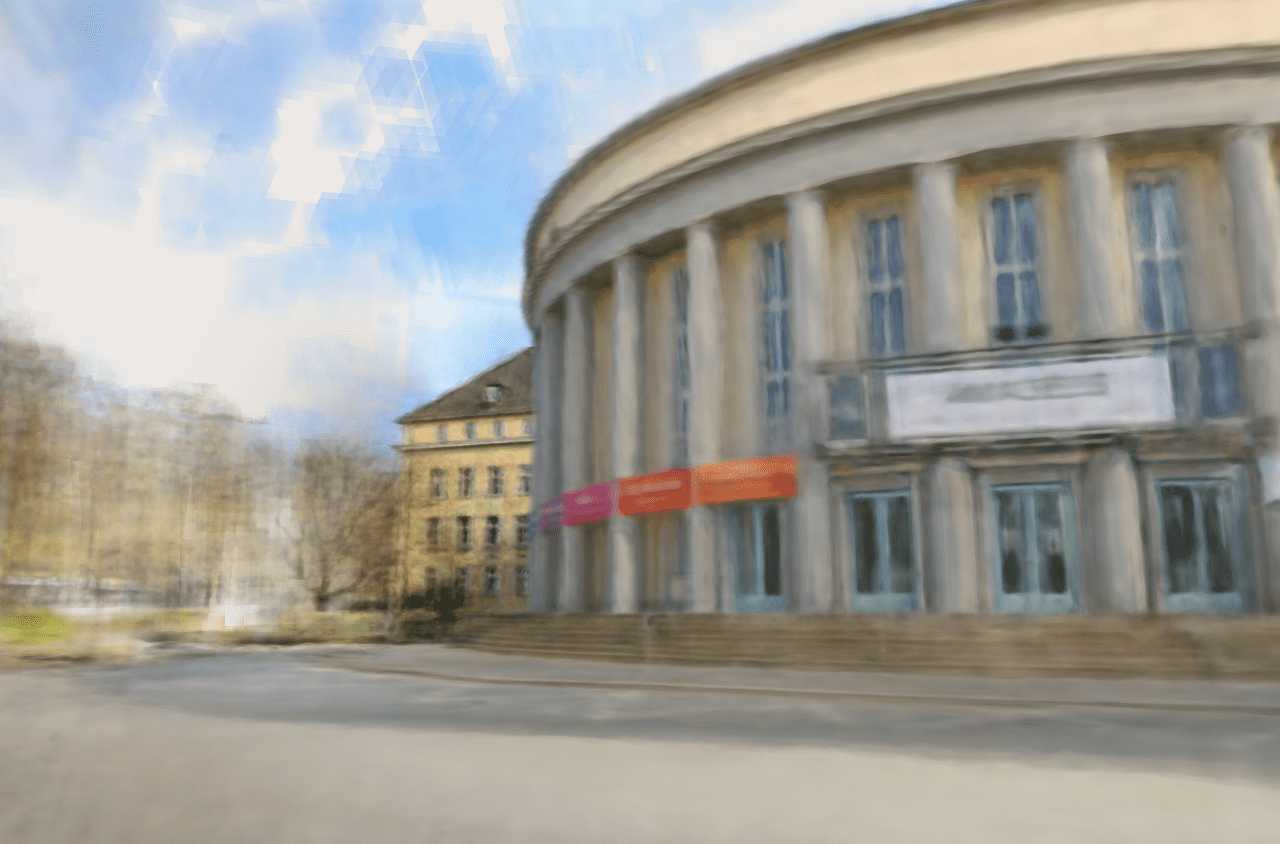} & 
    \includegraphics[width=1.0\linewidth]{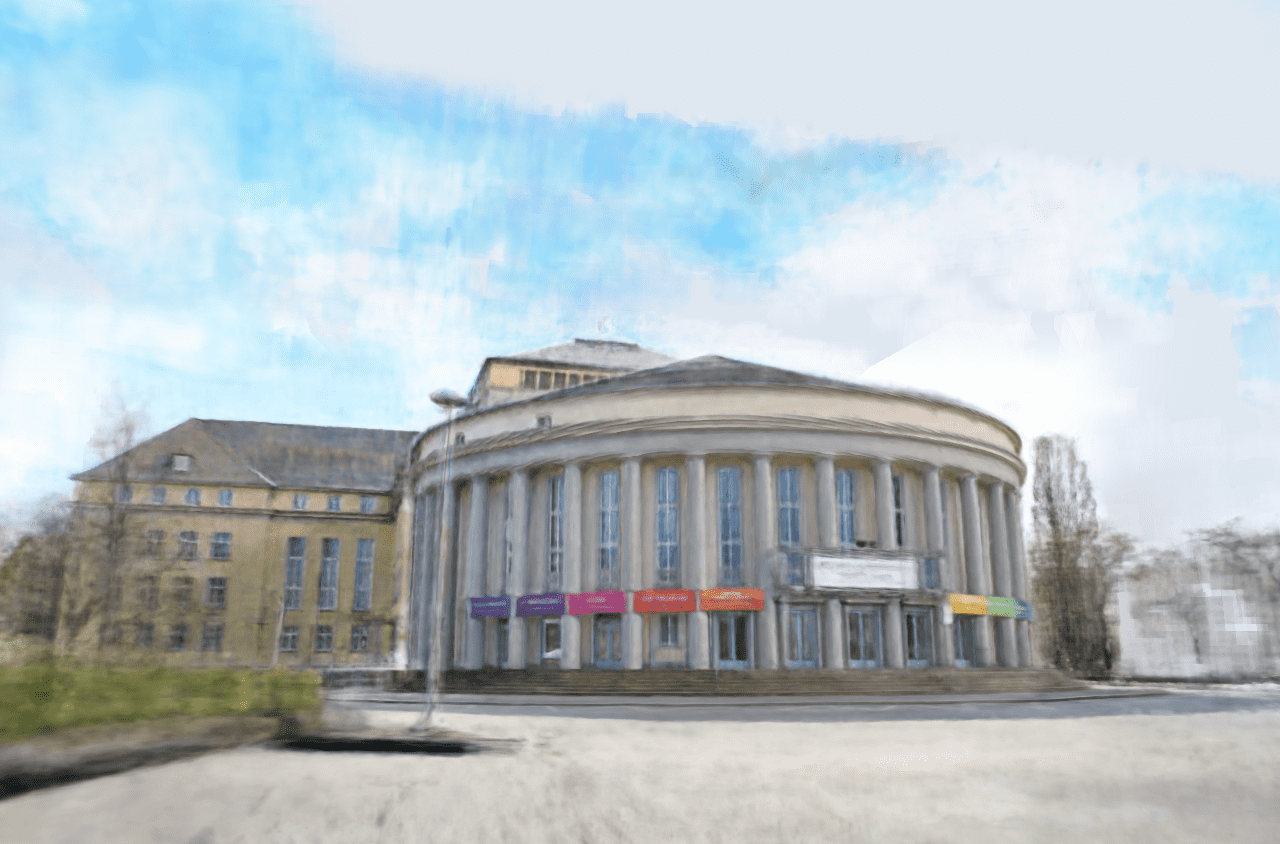} & 
    \includegraphics[width=1.0\linewidth]{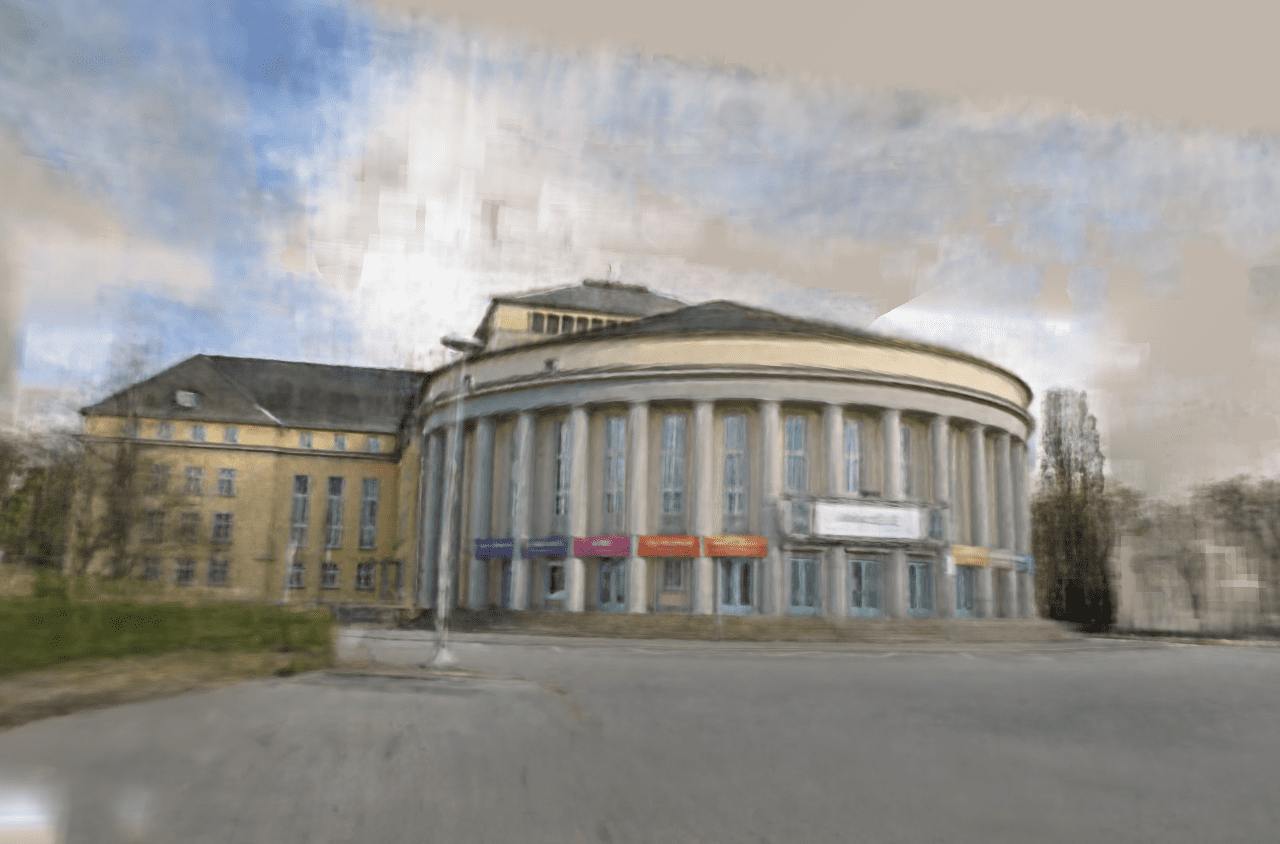} \\ 
    Shadow &
    \includegraphics[width=1.0\linewidth]{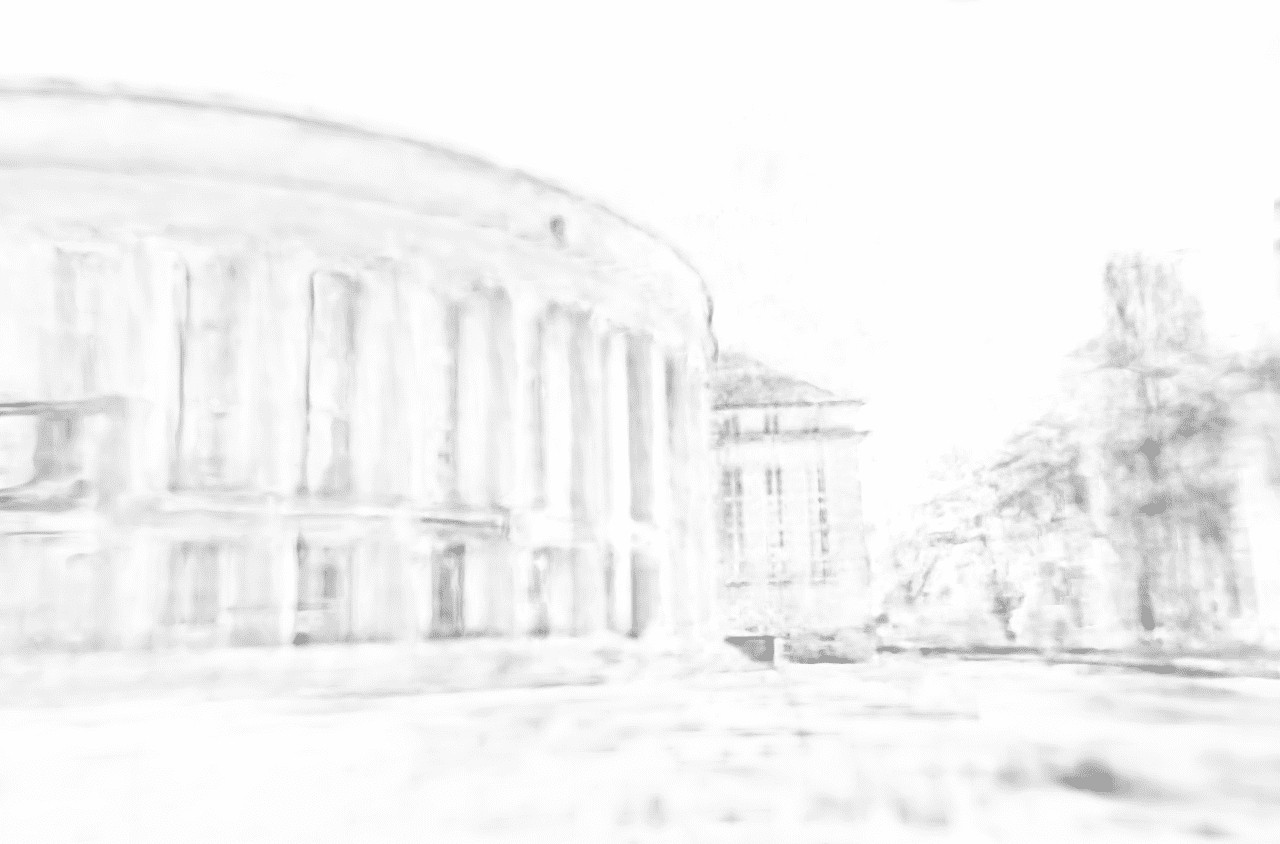} & 
    \includegraphics[width=1.0\linewidth]{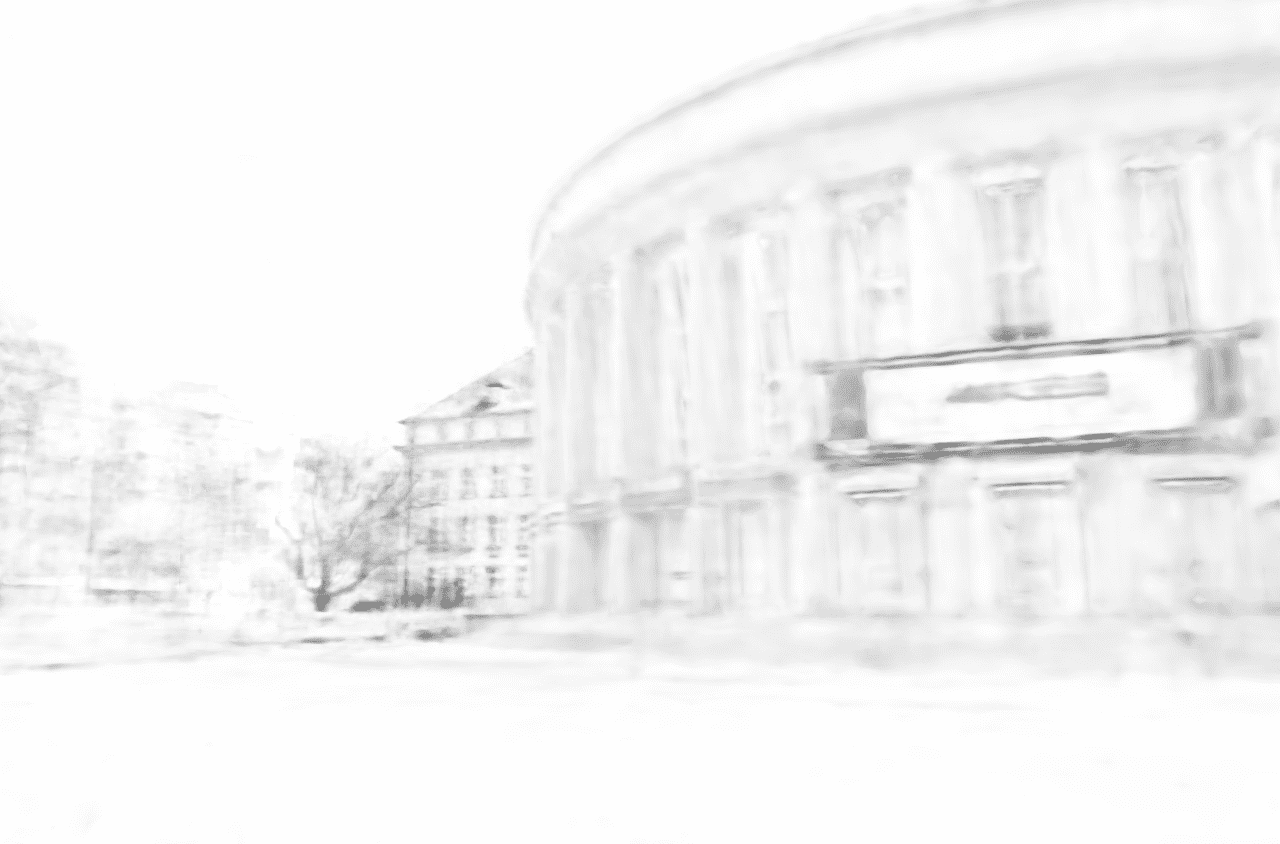} & 
    \includegraphics[width=1.0\linewidth]{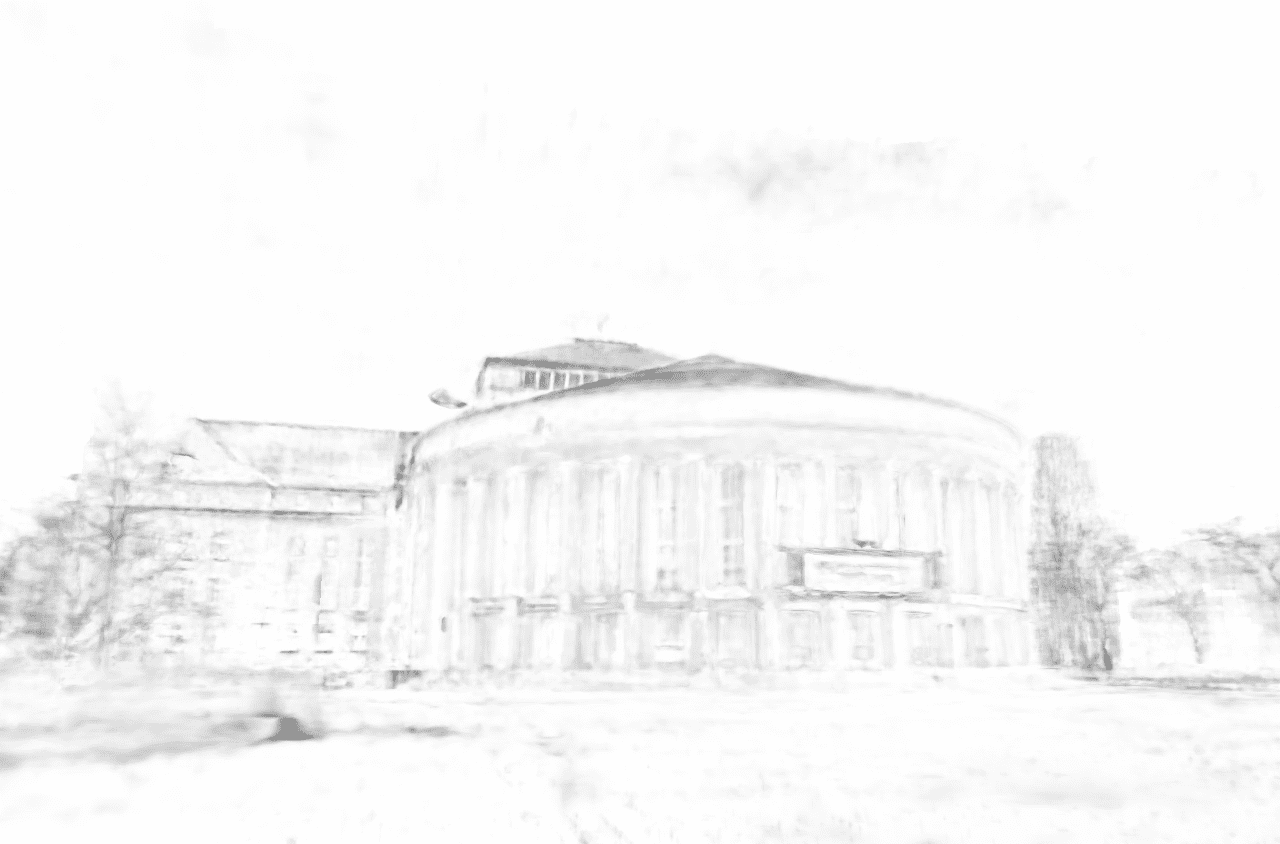} & 
    \includegraphics[width=1.0\linewidth]{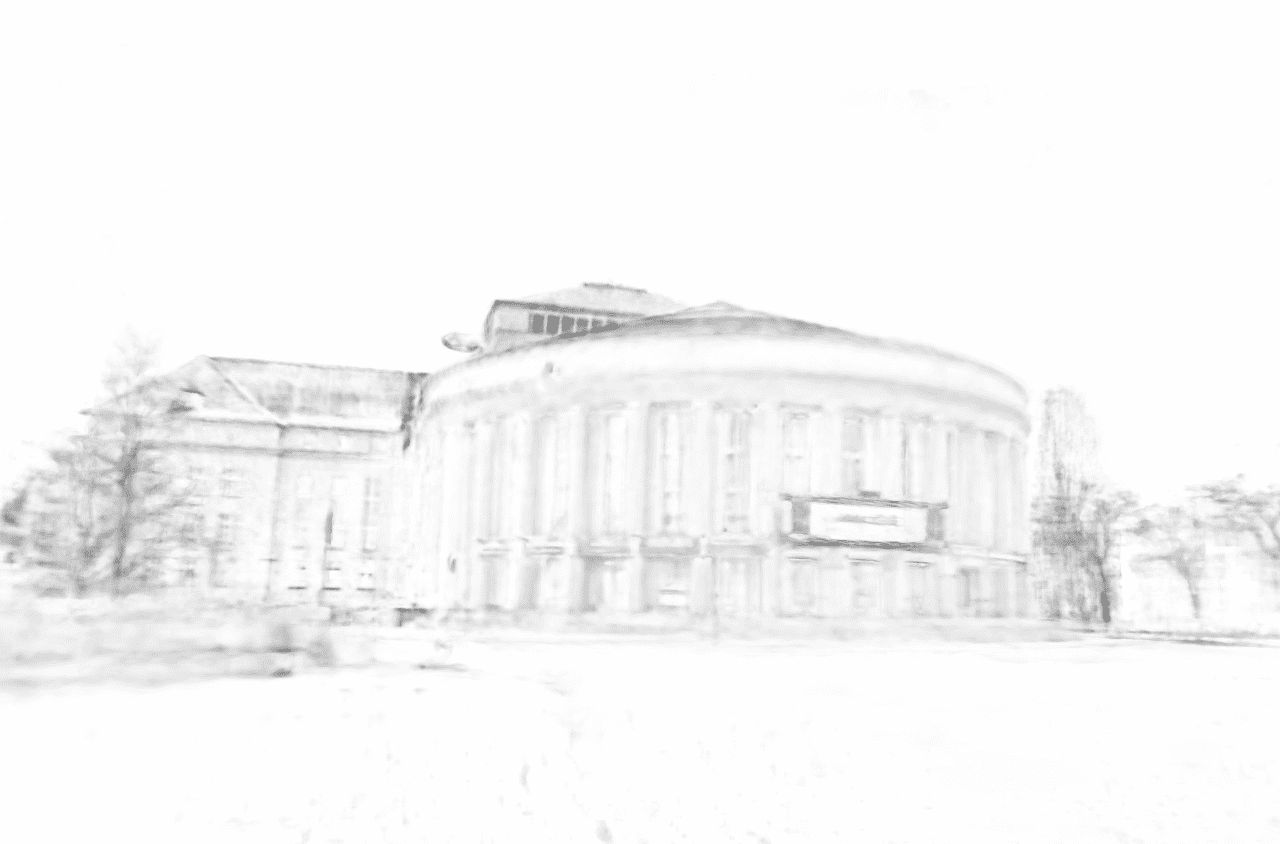} \\
    \end{tabular}
  }
  \caption{Site 2}
    \label{supp_fig:relight-a}
  \end{subfigure}
  \begin{subfigure}{0.9\linewidth}
    \centering
    \resizebox{1.0\linewidth}{!}{
    \begin{tabular}{{>{\centering\arraybackslash}m{0.14\linewidth}m{0.22\linewidth}m{0.22\linewidth}m{0.22\linewidth}m{0.22\linewidth}}}
    Ours &
    \includegraphics[width=1.0\linewidth]{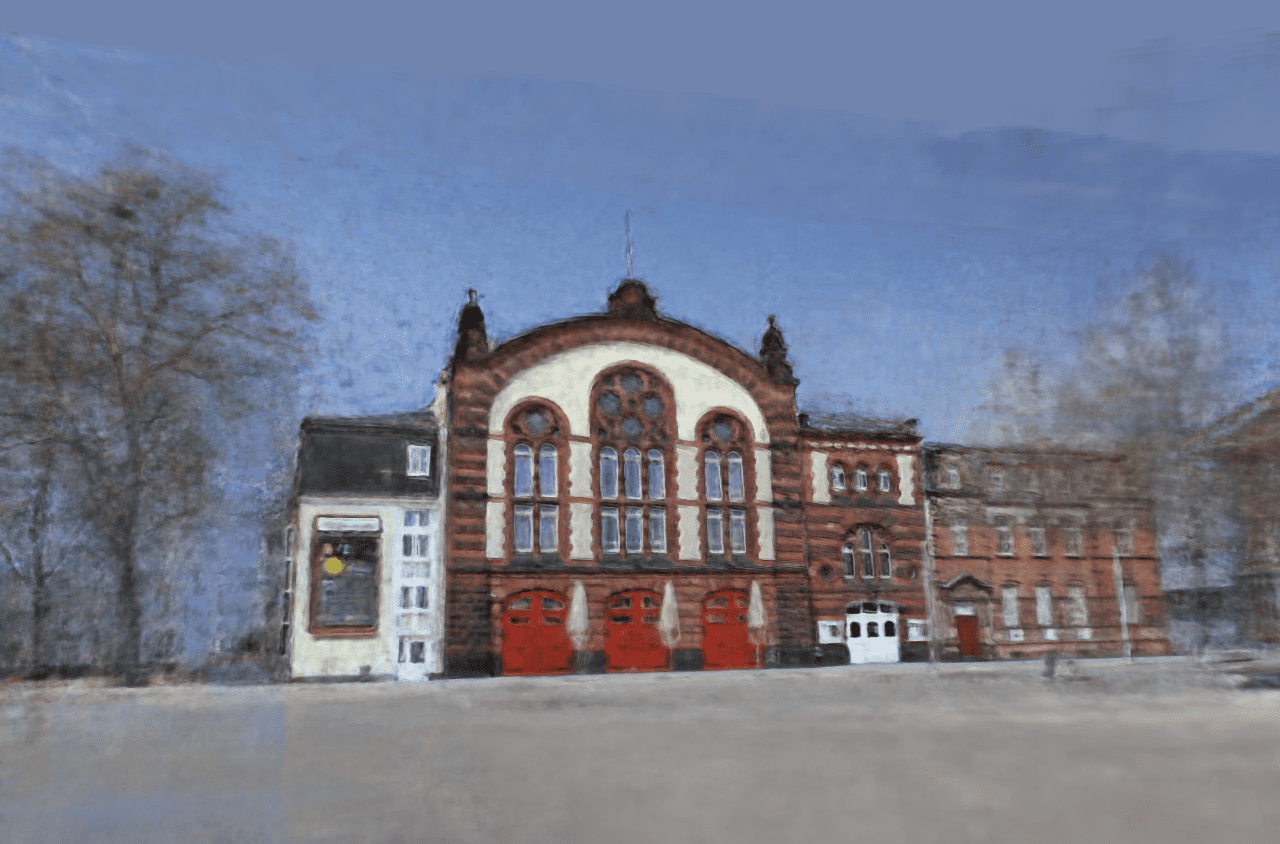} & 
    \includegraphics[width=1.0\linewidth]{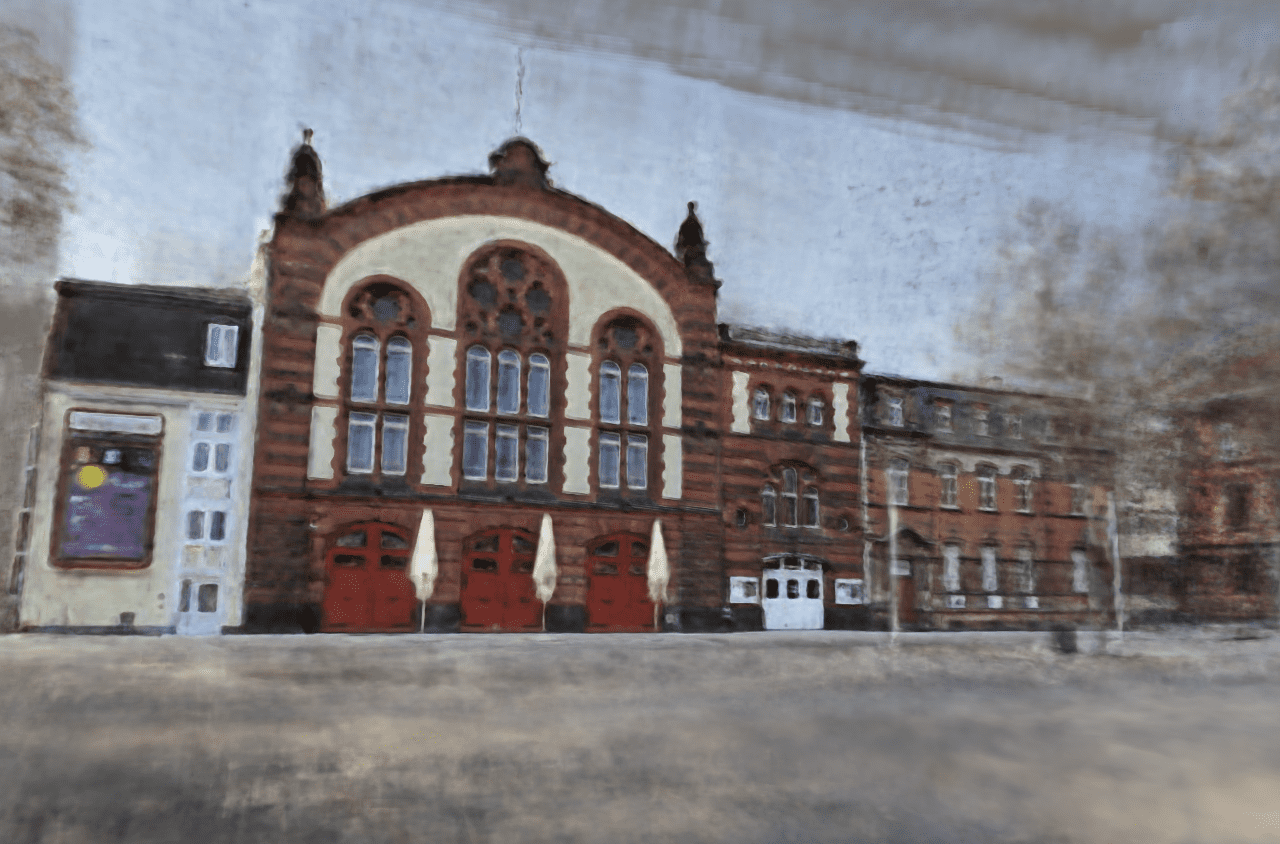} & 
    \includegraphics[width=1.0\linewidth]{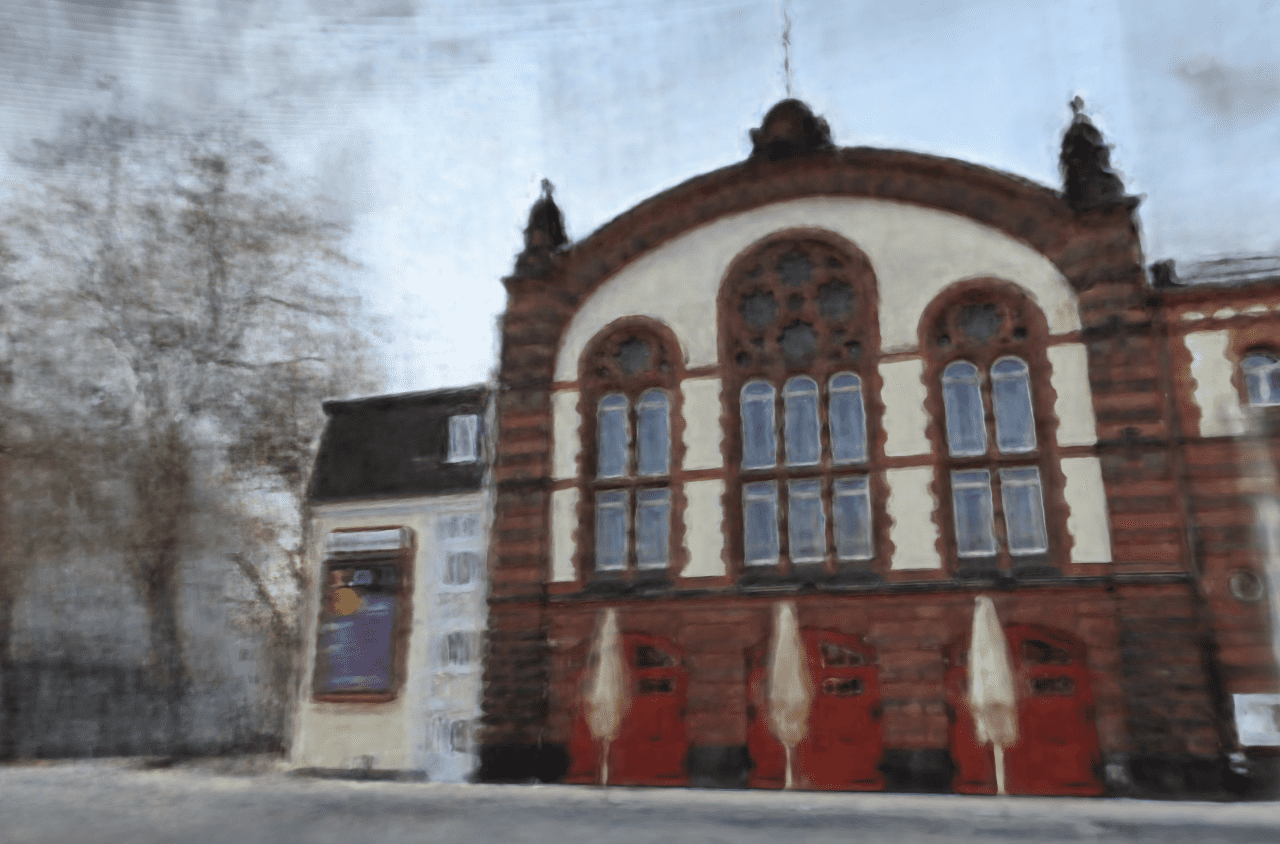} & 
    \includegraphics[width=1.0\linewidth]{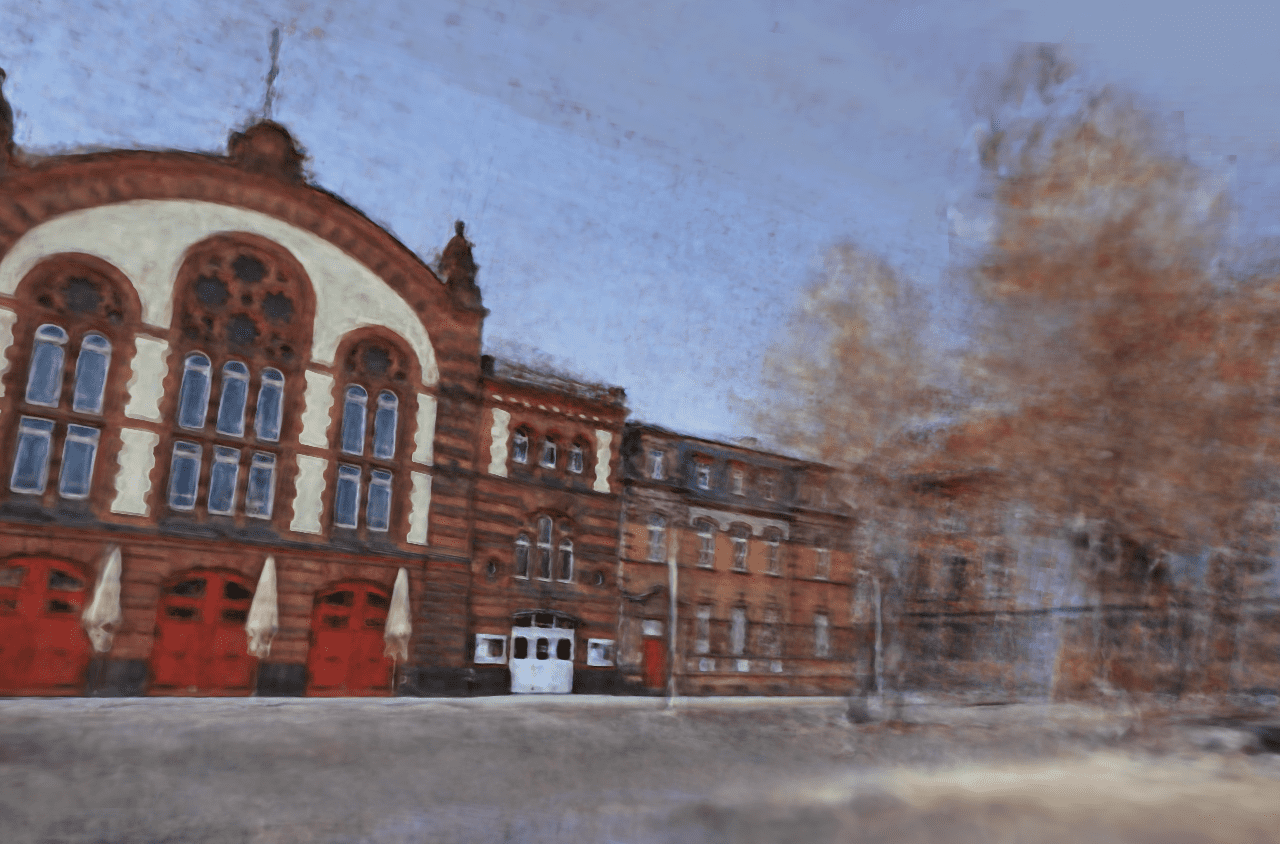} \\
    Shadow &
    \includegraphics[width=1.0\linewidth]{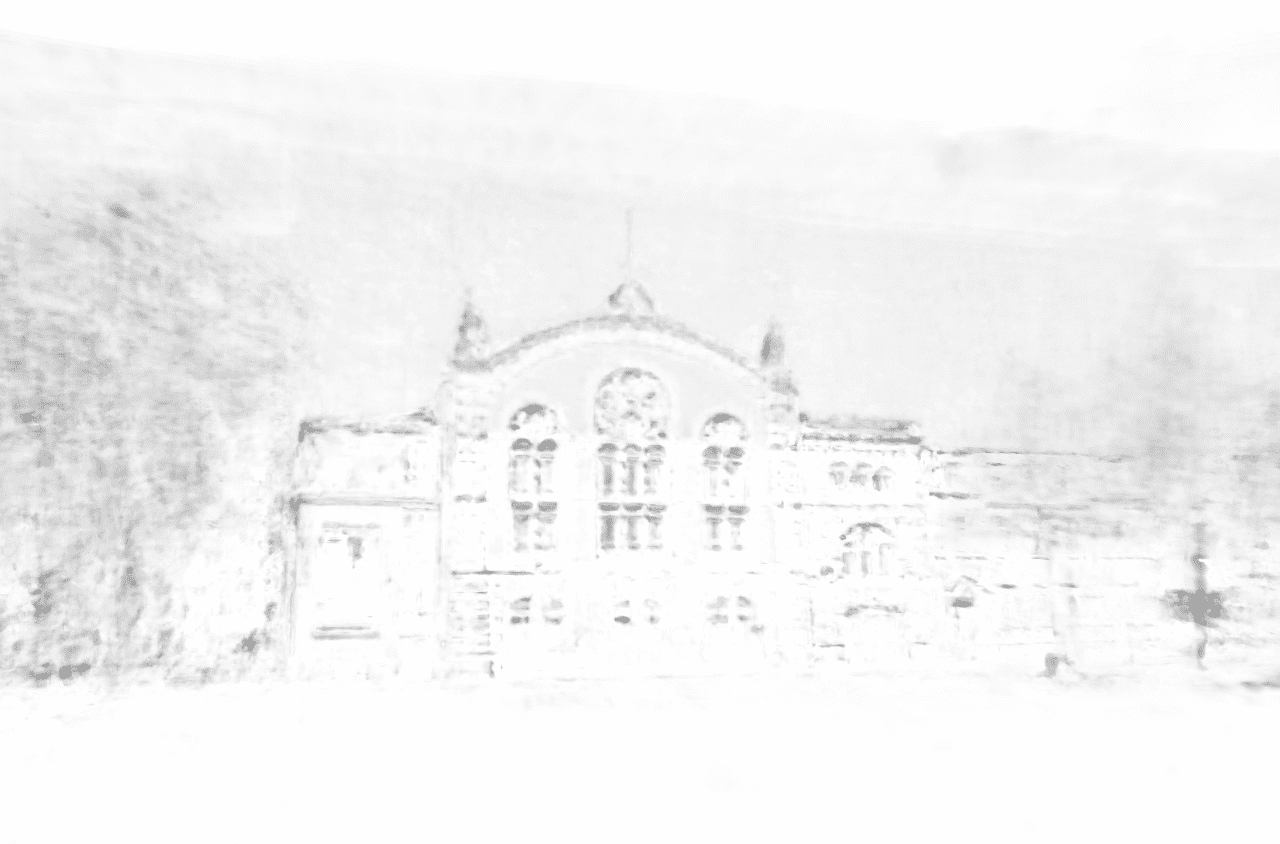} & 
    \includegraphics[width=1.0\linewidth]{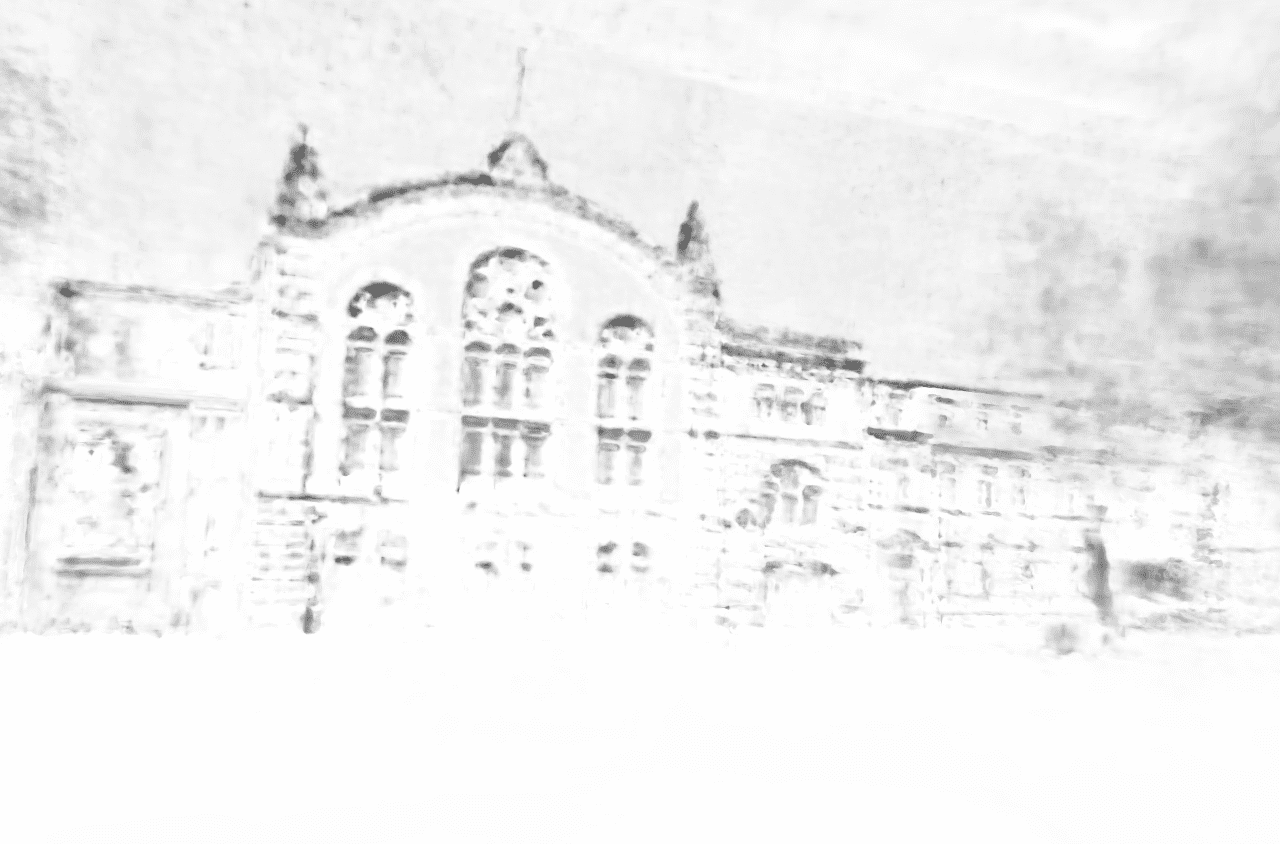} & 
    \includegraphics[width=1.0\linewidth]{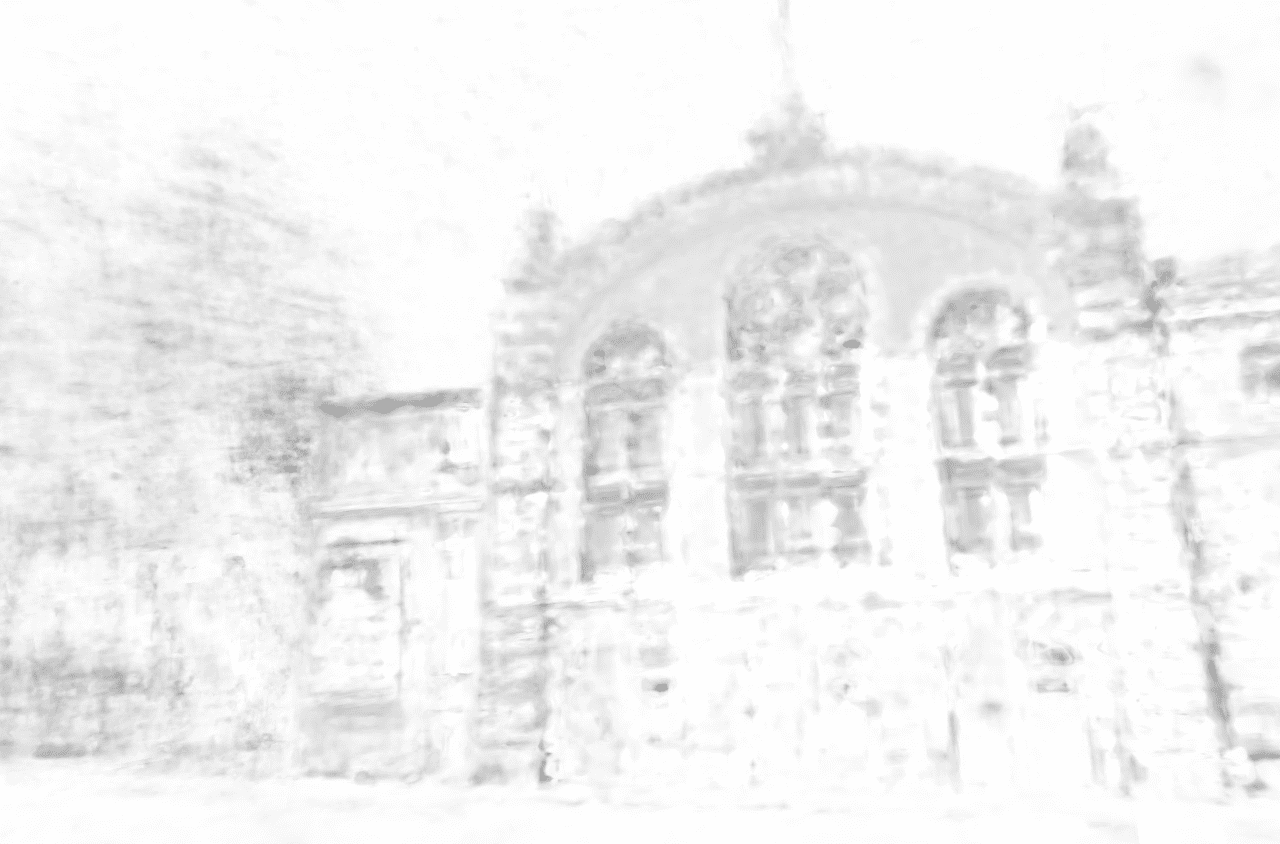} & 
    \includegraphics[width=1.0\linewidth]{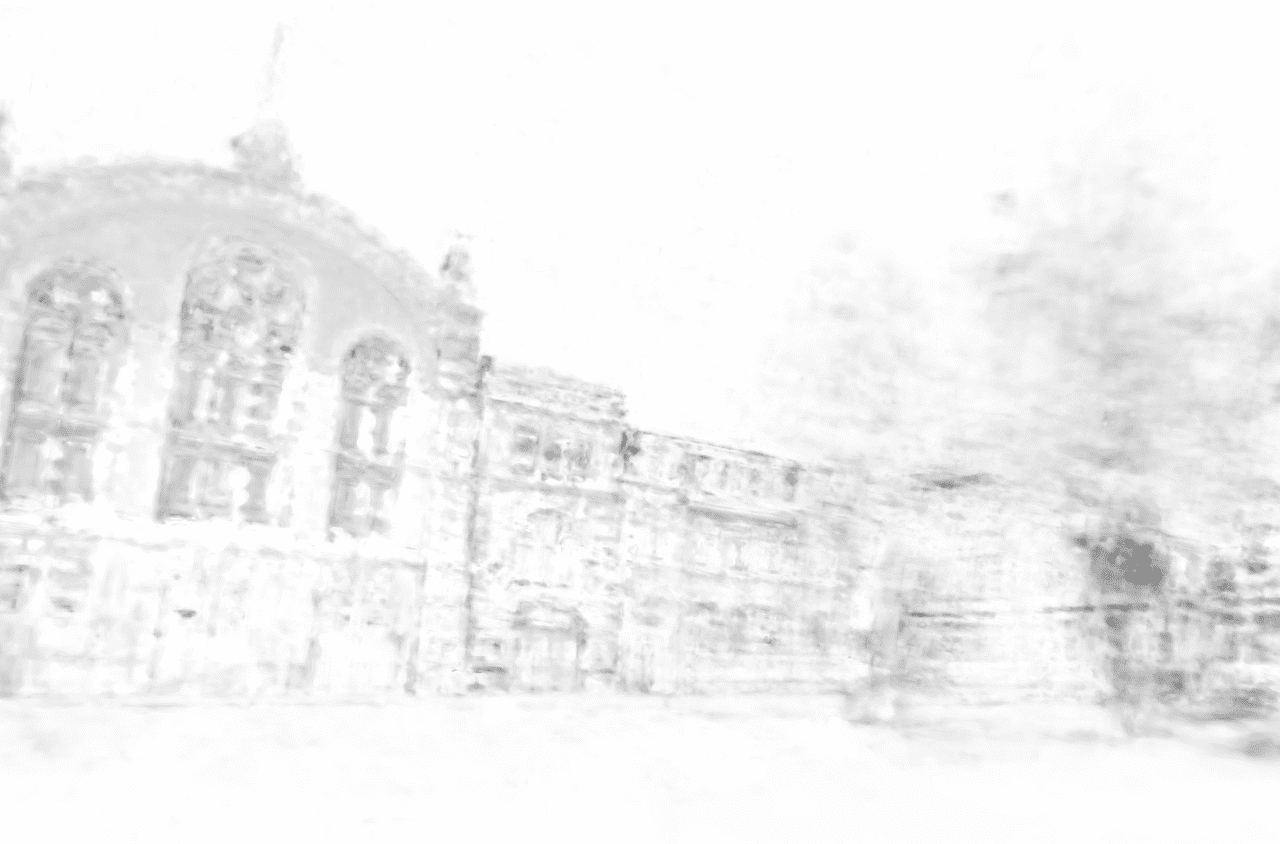} \\
    \end{tabular}
  }
  \caption{Site 3}
    \label{supp_fig:relight-b}
  \end{subfigure}
  \caption{\textbf{Additional Relight Rendering.}}
  \label{supp_fig:relight}
\end{figure*}

Fig.~\ref{supp_fig:time2} shows the shadows rendered for relighting for Site 2, and Fig.~\ref{supp_fig:time3} for Site 3.
Fig.~\ref{supp_fig:relight} shows the additional qualitative results of relighting.

\clearpage

{\small
\bibliographystyle{ieee_fullname}
\bibliography{egbib}
}